\documentclass[journal]{IEEEtran}
\usepackage{cuted}%%\stripsep-3pt
\usepackage{amsmath,lipsum}
\usepackage{mathrsfs}
\usepackage{mathtools, nccmath}
\usepackage{amsthm}
\usepackage{graphicx}
\usepackage{subfigure}
\usepackage[justification=RaggedRight]{caption}
\usepackage{booktabs}
\usepackage[colorlinks,linkcolor=blue]{hyperref}
\usepackage{cite}
\usepackage[numbers,sort]{natbib}
\usepackage{newtxmath}
\usepackage{multirow}
\usepackage{marvosym}
\usepackage{bm}
\usepackage{float}
\usepackage[noend]{algpseudocode}
\usepackage{algorithmicx,algorithm}
\usepackage{tabularx}
\usepackage{booktabs}
\usepackage{multicol}
\usepackage{array} 
\usepackage{dcolumn}
\usepackage{multicol}
\DeclareUnicodeCharacter{FB01}{fi}

\ifCLASSINFOpdf
\else
\fi

\begin{document}
\title{Wavelet Regularization Benefits Adversarial Training}

\author{Jun Yan\thanks{Jun Yan, Huilin Yin (corresponding author), Xiaoyang Deng, Ziming Zhao, Wancheng Ge, and Hao Zhang is with the College of Electronic and Information Engineering at Tongji University, Shanghai, China (E-mail: yanjun@tongji.edu.cn, yinhuilin@tongji.edu.cn, dxytju@tongji.edu.cn, zzm\_0623@163.com, gwc828@tongji.edu.cn, zhang\_hao@tongji.edu.cn). Gerhard Rigoll (rigoll@tum.de) is with the Institute for Human-Machine Communication in Technical University of Munich.},
        Huilin Yin\textsuperscript{\Letter},
        Xiaoyang Deng,
        Ziming Zhao,
        Wancheng Ge,
        Hao Zhang, and Gerhard Rigoll~\IEEEmembership{IEEE Fellow} %.}
       % <-this % stops a space
       
}


\markboth{Preprint}%
{Shell \MakeLowercase{\textit{et al.}}: Wavelet Decomposition Benefit Adversarial Training}



\maketitle

\begin{abstract}
 \textcolor{black}{Adversarial training methods are state-of-the-art (SOTA) empirical defense methods against adversarial examples. Many regularization methods have been proven to be effective with the combination of adversarial training. Nevertheless, such regularization methods are implemented in the time domain. Since adversarial vulnerability can be regarded as a high-frequency phenomenon, it is essential to regulate the adversarially-trained neural network models in the frequency domain. Faced with these challenges, we make a theoretical analysis on the regularization property of wavelets which can enhance adversarial training. We propose a wavelet regularization method based on the Haar wavelet decomposition which is named Wavelet Average Pooling.  This wavelet regularization module is integrated into the wide residual neural network so that a new WideWaveletResNet model is formed. On the datasets of CIFAR-10 and CIFAR-100, our proposed Adversarial Wavelet Training method realizes considerable robustness under different types of attacks. It verifies the assumption that our wavelet regularization method can enhance adversarial robustness especially in the deep wide neural networks. The visualization experiments of the Frequency Principle (F-Principle) and interpretability are implemented to show the effectiveness of our method. A detailed comparison based on different wavelet base functions is presented. The code is available at the repository: \url{https://github.com/momo1986/AdversarialWaveletTraining}.}
\end{abstract}


% \fi
\IEEEpeerreviewmaketitle

%\IEEEPARstart{T}{he} existence of adversarial examples can be seen as the crisis of the neural networks. The researchers are devoted to the proposal of effective defense methods to reply to the challenges of the attack methods~\cite{FGSM, PGD, MIM, C&W}.
\section{Introduction}
\textcolor{black}{\IEEEPARstart{T}{he} existence of adversarial examples is the obstruction of the realization of trustworthy deep learning. The neural networks with the high recognition performance on the clean testing datasets would be vulnerable to the white-box adversarial examples~\cite{L-BFGS, FGSM, PGD, MIM, C&W} which the adversaries have the knowledge of the target models and the training data distribution and black-box adversarial examples~\cite{ZOO, AutoZOOM, TIAttack, NATTACK, BanditPriors} which the adversaries have no access to the internal structures of the neural networks.}
% wish you the best of success.

\textcolor{black}{Many empirical defense methods have been proposed to mitigate the adversarial effects including distillation~\cite{DistillationDefense}, randomization~\cite{XieRandomization}, guided denoising~\cite{GuidedDenoiser}, pixel defending~\cite{PixelDefend}, pixel masking~\cite{PixelMasking}, Bayesian deep learning~\cite{AdvBNN}, and so on. Among these methods, adversarial training can be regarded as the SOTA defense method to promote robustness in the optimization process of neural networks~\cite{PGD, TRADES, RobustOverfitting, FAT, BagTricksAT, AdversarialDistributionalTraining}. The core idea is to improve the robust generalization ability in the min-max optimization process, in which the perturbations should be as large as possible while the expected loss of the model should be as small as possible. The ``gradient penalty" effect~\citep{advGradientPenalty} can be achieved via the Nash equilibrium so that the neural network models are satisfied with the small Lipschitz constraint. The proposal and development of adversarial training methods help improve the performance of the neural network model under adversarial attacks.}

\textcolor{black}{To realize the minimization of adversarial empirical risk, the regularization tricks would be combined with adversarial training methods to boost robustness, including logit pairing~\cite{ALP}, parametric noise injection~\cite{ParametricNoiseInjection}, early stopping~\cite{RobustOverfitting, BagTricksAT}, weight decay~\cite{BagTricksAT}, Jacobian regularization~\cite{JacobianRegularization}, and the utilization of CutMix method~\cite{CutMixRobustness}. All these methods are implemented in the time domain, which may ignore the phenomenon that the adversarial example is a high-frequency phenomenon~\cite{FourierRobustness, UAPFourierRobustness}. Therefore, it is worthwhile to regulate adversarial training in the frequency domain to boost adversarial robustness~\cite{FPrincipleTheory1, FPrincipleTheory2, FourierRobustness, HighFreqGeneralization}. The wavelet has properties of smoothness, approximation, and reproduction of polynomials~\cite{MallatBook,DaubechiesBook} which is suitable to handle nonstationary signals. It is a potential regularization method to handle high-frequency and nonstationary adversarial perturbations.}
%However, there are several open problems to be resolved. Firstly, the minimization of adversarial empirical risk would lead to the allocation of zero weight on the non-robust features~\cite{RobustnessOddsAcc}. Therefore, the robustness may be at odds with the clean accuracy. Similar viewpoints would be held in the previous researches~\cite{TRADES, MitgatingRobustnessAcc, LookRobustnessAcc}. Secondly, there is still room for the improvement of the robustness based on current technologies~\cite{TRADES, SelfAdaptiveTraining}. Thirdly, only few researches~\cite{BagTricksAT} give a roadmap on the real application of the adversarial training methods. Last but not least, the theoretical research on the robustness of neural networks and adversarial training is still not saturated.

  \begin{figure*}[!t]
        \centering
\includegraphics[scale=0.45]{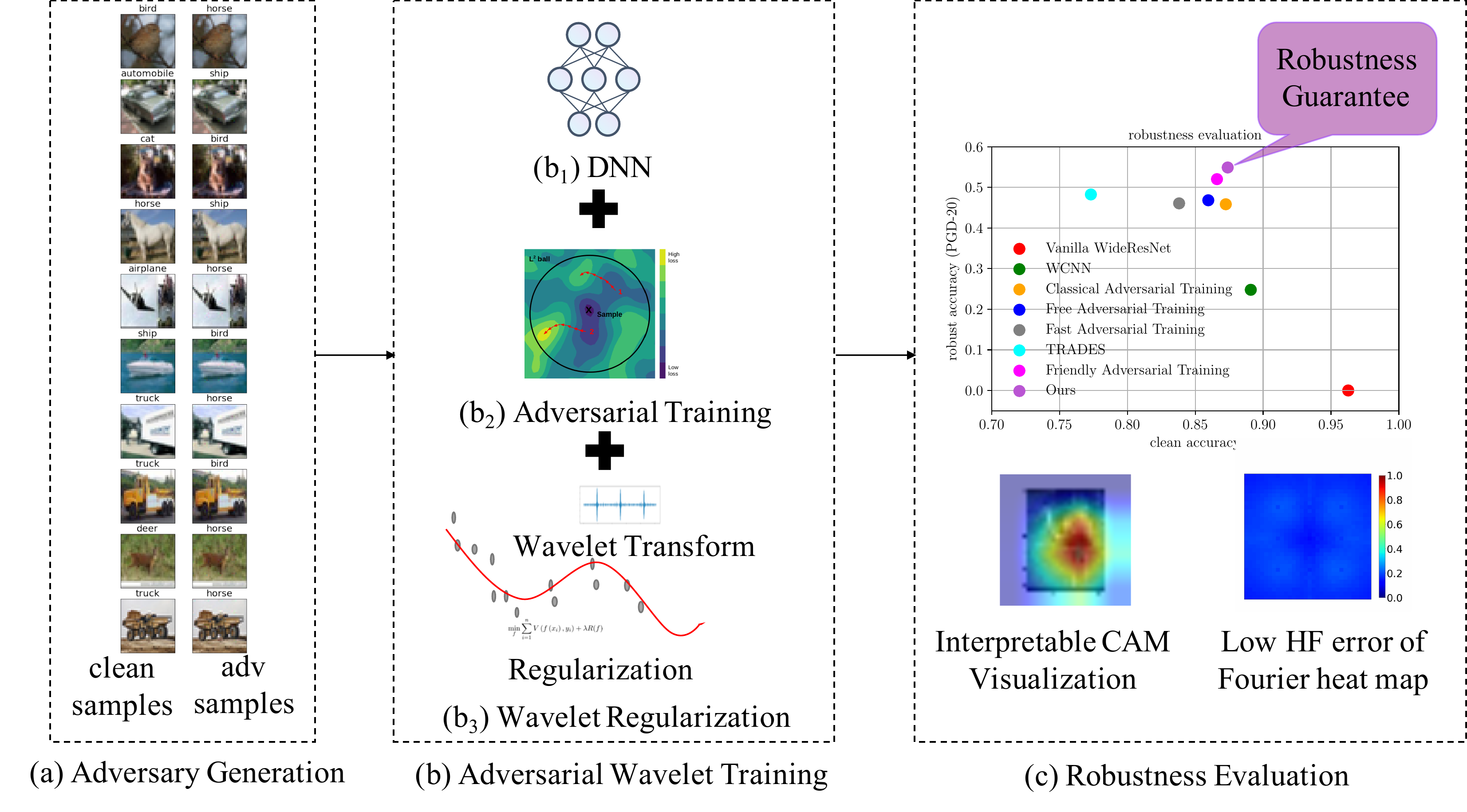}
\caption{Our Adversarial Wavelet Training framework. Fig. (a) is the data pipeline with the adversary generation on the clean dataset. Fig. (b) illustrates the Adversarial Wavelet Training method which is incorporated adversarial training and wavelet regularization can boost the robustness of DNNs (Fig.(b2) is cited from the blog article~\cite{robustnessBlogFig} @Towards Data Science). Fig. (c) demonstrates that our proposed method can realize the robustness guarantee. Moreover, the learning feature of our proposed method is interpretable with the visualization of the class activation mapping (CAM) method. From the perspective of F-Principle, our proposed method realizes a low high-frequency error in the Fourier heat map.}
\label{fig:AdvWaveletTrainingFrameWork}
\end{figure*}
\textcolor{black}{Since the high-frequency perturbation may lead to the adversarial vulnerabilities~\cite{FourierRobustness, UAPFourierRobustness}, the utilization of the wavelet would be a feasible regularization method. In this paper, we propose a framework combined with wavelet transform and adversarial training. As Fig.\ref{fig:AdvWaveletTrainingFrameWork} illustrates, the \textbf{Adversarial Wavelet Training} method incorporates an adversarial training framework with the regularization method of wavelet function implemented in the frequency domain. For our proposed method, we evaluate the clean accuracy and the robust accuracy on the CIFAR-10 and CIFAR-100 datasets~\cite{CIFAR-10}. We also give an elaborate theoretical analysis of our adversarial wavelet training method that such a framework satisfies the Lipschitz constraint.}

Our contributions of this paper are listed as follows:
\begin{itemize}%项目符号开始
\item We make the theoretical analyses about the robustness under the framework combined with the wavelet analysis and adversarial training for the proposal of adversarial wavelet training.
\item Our proposed wavelet regularization method based on adversarial training realizes the considerable robustness under the different types of attacks including white-box attacks, black-box attacks, adaptive attacks, and auto attacks on the CIFAR-10 dataset and CIFAR-100 dataset~\cite{CIFAR-10}.
\item We make an empirical comparison between different wavelet base functions to prove that the Haar base function is the most suitable.
\end{itemize}

This paper is organized as follows. The related work is introduced in Section II. In Section III, we make a theoretical analysis to show the effectiveness of wavelet transform as a regularization method for adversarial training. Our proposed method is illustrated in Section IV. The experiment detail and result would be described in Section V. Finally, Section VI concludes our research.
 
\section{Related Work}
\textcolor{black}{In this section, we would have a literature review on the work related to adversarial robustness. The adversarial defense methods against the adversarial attacks would be introduced. Moreover, we would focus on the utilization of wavelet methods in deep learning.}
\subsection{Adversarial Robustness: Attack And Defense}
\textcolor{black}{In the field of computer vision, the adversarial examples that the original clean images are combined with the imperceptible noises can fool the neural visual systems. In the category of white-box attacks, the adversaries could query the information of model structures and model parameters. The search for minimal adversarial perturbations can be modeled as an optimization problem~\cite{L-BFGS, C&W}, a gradient perturbation problem~\cite{FGSM,  MIM, PGD}, or a classification hyperplane problem~\cite{DeepFool}. However, in the real applications of artificial intelligence (AI), the knowledge of the models cannot be obtained. The research on the black-box attacks with the manipulation of the model input/output (I/O) is more universal and realistic in such a scenario. A lot of methods have been proposed on such a research topic including the attacks based on the transferability~\cite{CurlsWhey, IlyasTransferAttack}, gradient estimation~\cite{ZOO, BanditPriors, AutoZOOM}, decision~\cite{BoundaryAttack, RandomSignFlipAttack}, and sampling~\cite{NATTACK}. On the evaluation of adversarial robustness, only the empirical study on the static attack methods~\cite{FGSM, PGD} is insufficient. Then, the adaptive attack methods~\cite{C&W, BPDA, OnAdaptiveAttack} are proposed to break the defense methods with counter-countermeasures. To fix the pitfalls of improper tuning of hyperparameters of the attacks, gradient obfuscation, or masking in the robustness evaluation, the auto-attack method~\cite{AutoAttack} with the extension of the multi-step iterative attack method~\cite{PGD} can provide a fair evaluation of the adversarial robustness. The existence of adversarial examples is a huge challenge but also a great promise for trustworthy deep learning.}

\textcolor{black}{To handle the security threat of deep learning, many empirical defense methods have been proposed. The adversarial attack detection methods can reject the adversarial examples to avoid deceit. The detection methods can be built on the support vector machines (SVMs)~\cite{SVMDetection}, SVMs with Discrete Wavelet Transform and Discrete Sine Transform~\cite{WaveletSVMDetection}, deep Bayesian classifier~\cite{BayesianClassifierDetection}, variational autoencoder (VAE)~\cite{VAEDetector}, perturbation rectifying  network~\cite{PerturbRectifyNetwork}, local intrinsic dimensionality~\cite{LIDDetector}, K-density detector with Reverse Cross-Entropy~\cite{PangAdvDetector}, Fisher information matrix~\cite{FisherInformationMetricDetector}, nonlinear autoencoder (AE)~\cite{DAMAD}, and so on. The attack mitigation methods are inclined to the inference on the adversarial examples rather than the direct rejection. The input-transformation methods~\cite{JPEGDefense, TVM, FeatureSqueezing} can handle the black-box noise attacks but shrivel under the white-box adversarial attacks~\cite{BPDA}. The recent work of pixel masking~\cite{PixelMasking} explores cognitive data augmentation to improve robustness under white-box adversarial attacks. The ensemble methods~\cite{HuanruiEnsemble, PangEnsemble}, the randomization methods~\cite{XieRandomization, AdvBNN}, and the combination of the ensemble method and the randomization method~\cite{RSE} are proposed to realize the security by obscurity. The distillation method~\cite{DistillationDefense} and the pruning method~\cite{SparseDNN} attempt to realize the adversarial robustness through the model compression. }

\textcolor{black}{Among the empirical defense methods, adversarial training~\cite{PGD, TRADES, EnsembleAdversarialTraining, SelfAdaptiveTraining, FastAdversarialTraining, FreeAdversarialTraining, BagTricksAT, FAT, AdversarialDistributionalTraining} is considered the feasible scheme against adversarial attacks. The restraint on the gradients is validated to be beneficial to the adversarial robustness~\cite{PGD}. Many pieces of research related to adversarial training have been delivered including the realization of a better trade-off between clean accuracy and robust accuracy~\cite{TRADES, RobustnessOddsAcc, MitgatingRobustnessAcc, LookRobustnessAcc}, its combination with semi-supervised learning~\cite{SelfAdaptiveTraining}, curriculum learning~\cite{FAT}, or ensemble method~\cite{EnsembleAdversarialTraining}, and reduction on the training cost~\cite{FreeAdversarialTraining, FastAdversarialTraining}.}
\par{\textcolor{black}{At the model level, most adversarial training methods are implemented on the WideResNet model~\cite{WideResNet}. Diverse pieces of theoretical literature show that the wide model can find the approximately-global-optimal minima during the optimization process~\cite{GDSearch, WideResNetLossSurface, WuOverparameterization} with the learning dynamics of a linear model~\cite{GaussianProcessNN, WideNetworkAsLinearModel}. The overparameterization phenomenon of WideResNet would make the loss surface of deep learning more smooth and could transform the non-convex optimization problem into the ``analogous" convex problem. The theoretical foundation can be referred to the work on learning theory~\cite{ConvergenceOperParameterization, CaoGeneralizationBoundWideNetwork,AllenZhuOverParameterization,ZouOverparameterization}. The recent research via the tool of tensor interpolation~\cite{LawRobustness} and isoperimetry~\cite{UniversalLawRobustness} also supports the conjecture that the overparameterization is the robustness guarantee of many deep learning models including WideResNet. It is counterintuitive in the context of structural risk minimization (SRM). Moreover, the ``overfitting" problem is a spiny problem for the realization of adversarial empirical risk minimization (ERM) during the defense process. Thus, it incurs a need to utilize regularization techniques in adversarial training to restrain the solution space. The recent research from the perspective of neural tangent kernel (NTK)~\cite{WuWRNRobustness} suggests that the regularization method is still essential to tune the perturbation stability of deep wide neural networks under the attacks despite the intrinsic robust property of the deep wide neural model structures. It breeds the necessity of the research of regularization methods in the adversarial training of wide deep neural networks.}}
\par{\textcolor{black}{Currently, early stopping and weight decay are two effective regularization methods in adversarial training~\cite{RobustOverfitting, BagTricksAT}. Other regularization methods like logit pairing~\cite{ALP}, parametric noise injection~\cite{ParametricNoiseInjection}, curvature regularization~\cite{CurvatureRegularization}, and Jacobian regularization~\cite{JacobianRegularization} can be combined with adversarial training. Not long ago, it has been proven that the CutMix method~\cite{CutMix} can also help boost adversarial robustness~\cite{CutMixRobustness}. However, all these methods are implemented in the time domain. The Fourier perspective~\cite{FourierRobustness, UAPFourierRobustness} provides an illustration of the adversarial vulnerability. Therefore, it inspires us to provide a regularization method for adversarial training in the frequency domain.}}
\subsection{Wavelet On Deep Learning}
Wavelet analysis is an important research field of modern mathematics~\cite{DaubechiesBook, MeyerBook, MallatBook} that inspires wireless communication, image science, and quantum physics. In the field of statistical learning, researchers were devoted to analyzing the multi-scale property of the wavelet on trees, graphs, and other high-dimensional data for a further combination of semi-supervised learning or supervised learning~\cite{MultiScaleWavelet2010, FunctionSpaceWavelet}. The wavelet transform is a feasible method to achieve the regularization~\cite{DaubechiesBook, MallatBook}. 
\par{After the renaissance of deep learning, some research achievements are at the intersection of wavelet analysis and deep learning, including the method of group invariant scattering based on non-orthogonal Morlet wavelet~\cite{ScatteringNetwork, DeepHybridNetwork} or orthogonal Haar wavelet~\cite{HaarScatteringNetwork}, the CNN models based on the multi-resolution character of wavelet~\cite{WCNN, DeepAdaptiveWaveletNetwork}, wavelet pooling~\cite{WaveletPooling}, and deep learning denoising method based on the wavelet~\cite{WaveletDenoisingRobustnes}. The robustness of wavelet against noise~\cite{WaveletDenoisingRobustnes} and its down-sampling property in the frequency domain~\cite{WaveletPooling} inspire us to explore the application of wavelet in the research of adversarial training. }
%After the renaissance of deep learning, some research achievements are at the intersection of wavelet analysis and deep learning. Bruna et al.~\cite{ScatteringNetwork} proposed the method of group invariant scattering based on non-orthogonal Morlet wavelet~\cite{ScatteringNetwork}. Oyallon et al.~\cite{DeepHybridNetwork} extended the application of the scattering network on the ImageNet dataset. Cheng et al.~\cite{HaarScatteringNetwork} improved the structure of the scattering network to replace Morlet wavelet with orthogonal Haar wavelet. Fujieda et al.~\cite{WCNN} proposed a wavelet CNN model, which combines a multi-resolution analysis and convolution layers into one model with the utilization of the spectral information. Rodr{\'{\i}}guez et al.~\cite{DeepAdaptiveWaveletNetwork} proposed a CNN structure combined with the adaptive lifting mechanism in the wavelet analysis. Our research is inspired by the previous work~\cite{WaveletDenoisingRobustnes, WaveletPooling}. Li et al.~\cite{WaveletDenoisingRobustnes} proposed a wavelet deep learning method which is proved to be robust against noise. Williams et al.~\cite{WaveletPooling} explored the method to do down-sampling on the frequency domain.

However, all these methods are used in the field of natural training and never extended to robust learning. \textcolor{black}{The work of wave transform attack~\cite{WaveTransformAttack} is an exception that combines the wavelet method with the research of adversarial examples. Such a study demonstrates the feasibility of adversarial robustness on wavelet.} Therefore, we are devoted to combining the adversarial training with the wavelet method to utilize the character of wavelet: multi-resolution analysis, and regularization. Besides these motivations, wavelet transform has also connections with the Frequency Principle (F-Principle)~\cite{FourierRobustness} in deep learning. The traditional Fourier transform is coarse-grained and global. In contrast, the multi-resolution property of the wavelet transform can aggregate the low-frequency elements and the high-frequency elements in a more rational way to promote robustness under the perturbations. 
\section{Theoretical Preliminary}
\subsection{Wavelet Transform As Regularization}
Our analysis is inspired by the pure mathematics research~\cite{DaubechiesBook, MallatBook}, and we extended it to adversarial deep learning.
The robustness of neural networks can be described in the concept of ``Lipschitz constraint".
\newtheorem{definition}{Definition}[section]\label{LipschitzConstraint}
\begin{definition}(Lipschitz continuity)
If $\left|x_{1}-x_{2}\right| \leqslant \varepsilon$, and $\left|f\left(x_{1}\right)-f\left(x_{2}\right)\right| \leqslant L\left|x_{1}-x_{2}\right|$, then the function $f(x)$ satisfies  with the Lipschitz constant $L$, $0<L<\infty$.
\end{definition}
The robust optimization of neural network is based on a locally considerable Lipschitz constraint ($L$ should be as small as possible). Such a constraint could keep the stability of the system under perturbation.
\newtheorem{definition2}{Definition}[section]\label{ContinuousWaveletTransform}
\begin{definition}(Continuous wavelet transform)
There exist a dilation factor ``$a$'' and a translation factor ``$b$" which vary continuously over the real number space $\mathbb{R}$, the continuous wavelet transform is given by ``resolution identity" formula defined in Eq. (\ref{ResolutionIdentity}):
\begin{equation}\label{ResolutionIdentity}
f=C_{\psi}^{-1} \int_{-\infty}^{\infty} \int_{-\infty}^{\infty} \frac{d a d b}{a^{2}}\left\langle f, \psi^{a, b}\right\rangle \psi^{a, b},
\end{equation}	
where the wavelet base is $\psi^{a, b}(x)=|a|^{-1 / 2} \psi\left(\frac{x-b}{a}\right)$, and $(, )$ denotes the $L^{2}$-inner product defined in the Hilbert space. For the constant $C_{\psi}$, it is determined by the base $\psi$ and defined in Eq. (\ref{ConstantWithBase}):
\begin{equation}\label{ConstantWithBase}
C_{\psi}=2 \pi \int_{-\infty}^{\infty} d \xi|\hat{\psi}(\xi)|^{2}|\xi|^{-1}.
\end{equation}
A doubly-indexed family of wavelets from $\psi$ by dilating and translating is generated by Eq. (\ref{WaveletBase}):
\begin{equation}\label{WaveletBase}
\psi^{a, b}(x)=|a|^{-1 / 2} \psi\left(\frac{x-b}{a}\right),
\end{equation}
where $a, b \in \mathbb{R},\ a \neq 0$. The normalization of base is $\left\|\psi^{a, b}\right\|=\|\psi\|\ for\ a,\ b $ with the assumption of $\|\psi\|=1$. The continuous wavelet form can be written as the form of Eq. (\ref{NewContinuousWaveletForm}):
\begin{equation}\label{NewContinuousWaveletForm}
\begin{aligned}
\left(T^{\text {wav }} f\right)(a, b) &=\left\langle f, \psi^{a, b}\right\rangle \\
&=\int d x f(x)|a|^{-1 / 2} \psi\left(\frac{x-b}{a}\right).
\end{aligned}
\end{equation}
Note that $\left(T^{\operatorname{wav}} f\right)(a, b) \mid \leq\|f\|$.
\end{definition}
The study of computer vision is based on discrete space. The introduction of the continuous wavelet transform could be a preliminary for the further description of the discrete wavelet transform.

The theorems below were previously depicted in the book of Daubechies~\cite{DaubechiesBook}, some theorems are ameliorated to describe the scenario of adversarial robustness clearly while others are directly derived from the research achievement of pure mathematics.
\newtheorem{theorem}{Theorem}[section]
\begin{theorem} (Restated Lipschitz constraint of wavelet base in the H{\"{o}}lder space)\label{LipschitzWaveletHolederSpace}:
There exists an input feature variable $x$ with the perturbation $\delta$, suppose that $\int d x(1+|x|)|\psi(x)|<\infty$ and the Fourier transform $\hat{\psi}(x)$ has the property $\hat{\psi}(0)=0$. The bounded function $f$ in the H{\"{o}}lder space is continuous with the exponent $\alpha$ ($0<\alpha \leq 1$) as defined in Eq. (\ref{LipschitzExponet}):
\begin{equation}\label{LipschitzExponet}
|f(x + \delta)-f(x)| \leq C|\delta|^{\alpha}.
\end{equation}
Then, its wavelet transform is satisfied with Eq. (\ref{WaveletTransformHolder}):
\begin{equation}\label{WaveletTransformHolder}
\left|T^{\text {wav }}(a, b)\right|=\left|\left\langle f, \psi^{a, b}\right\rangle\right| \leq C^{\prime}|a|^{\alpha+1 / 2}.
\end{equation}
\label{WaveletLipschitzHolder}
\end{theorem} 
The Theorem \ref{LipschitzWaveletHolederSpace} derives a theoretical framework that wavelet base is Lipschitz-continuous in the H{\"{o}}lder space. It supports the property of the local stability of the wavelet base. The related proof would be provided in the appendix.
\newtheorem{theorem2}{Theorem}[section]
\begin{theorem} (Compact support of wavelet base)\label{CompactSupportWavelet}~\cite{DaubechiesBook}
Suppose that the wavelet base $\psi$ is compactly supported, the universal function $f \in L^{2}(\mathbb{R})$ is bounded and continuous with the exponent $\alpha,\  0<\alpha \leq 1$. Then, the wavelet transform of $f$ is satisfied with Eq. (\ref{HoelderWaveletTransform}):
\begin{equation}\label{HoelderWaveletTransform}
\left|\left\langle f, \psi^{a, b}\right\rangle\right| \leq C|a|^{\alpha+1 / 2},
\end{equation}
 where $f$ is H{\"{o}}lder continuous with the exponent $\alpha$. 
\end{theorem} 
The Theorem \ref{CompactSupportWavelet} provides an upper bound of the transform of a compactly supported wavelet base. Generally speaking, such an upper bound determines the mutation point of the signal. The proof would be provided in the appendix.

Both the Theorem \ref{LipschitzWaveletHolederSpace} and the Theorem \ref{CompactSupportWavelet} indicate that the H{\"{o}}lder continuity of a function can be characterized by the decay value $\alpha$ of the wavelet transform. Moreover, the wavelet transform can also be used to characterize
the local regularity of the signal as a mathematical zoom, which surpasses the windowed Fourier transform. It helps wavelet transform avoid the aliasing effect of high-frequency signals and singular signals.
\newtheorem{theorem3}{Theorem}[section]
\begin{theorem} (Restated local regularity of wavelet transform)\label{LocalRegularityWavletForm}
Suppose that $\int d x(1+|x|)|\psi(x)|<\infty$ and $\int d x \psi(x)=0$. If a bounded function $f$ is H{\"{o}}lder continuous in $x_{0}$, with the exponent $\alpha \in[ 0,1]$. On the robustness problem with the perturbation $\delta$, there exists an inequation $\left|f\left(x_{0}+\delta\right)-f\left(x_{0}\right)\right| \leq C|\delta|^{\alpha}$. Furthermore, it can be deduced that $\left|\left\langle f, \psi^{a, x_{0}+b}\right\rangle\right| \leq C|a|^{1 / 2}\left(|a|^{\alpha}+|b|^{\alpha}\right)$.
\end{theorem} 
The Theorem \ref{LocalRegularityWavletForm} illustrates the local regularity of the wavelet transform. It is a further derivation of Theorem \ref{LipschitzWaveletHolederSpace} with a different restricted condition which supports the property of local regularity of wavelet base. The corresponding proof would be provided in the appendix.
\newtheorem{theorem4}{Theorem}[section]
\begin{theorem} (Upper bound of wavelet transform inside the H{\"{o}}lder space)\label{UpperBoundWavelet}~\cite{DaubechiesBook}
Suppose that the wavelet base $\psi$ is compactly supported, the function $f \in L^{2}(\mathbb{R})$ is bounded and continuous with the variable of $\gamma>0$ and $\alpha \in[ 0, 1]$, there exists a formula defined in Eq. (\ref{ContinuousWaveletHoelderSapce}):
\begin{equation}\label{ContinuousWaveletHoelderSapce}
\begin{array}{c}
\left|\left\langle f, \psi^{a, b}\right\rangle\right| \leq C|a|^{\gamma+1 / 2} \quad \text { uniformly in } b, \\
\left|\left\langle f, \psi^{a, b+x_{0}}\right\rangle\right| \leq C|a|^{1 / 2}\left(|a|^{\alpha}+\frac{|b|^{\alpha}}{|\log | b||}\right),
\end{array}
\end{equation}
where $f$ is H{\"{o}}lder continuous with the exponent $\alpha$. 
\end{theorem} 
The Theorem \ref{UpperBoundWavelet} derives an upper bound of the wavelet transform inside the H{\"{o}}lder space. Both Theorem \ref{CompactSupportWavelet} and Theorem \ref{UpperBoundWavelet} show that compactly-supported wavelet bases have strong regularization constraints. The related proof would be provided in the appendix.
\newtheorem{theorem5}{Theorem}[section]
\begin{theorem} (Restated Lipschitz constraint of wavelet in the multi-resolution space)\label{HoelderContinuityInSmallPerturbation}
Suppose that the perturbation is $2^{-(j+1)} \leq \delta \leq 2^{-j}$, there exists a variable $l \in N$ which satisfies one of the two followings $(\ell-1) 2^{-j} \leq x \leq x+\delta \leq \ell 2^{-j}$ or $(\ell-1) 2^{-j} \leq x \leq \ell 2^{-j} \leq x+\delta \leq(\ell+1) 2^{-j}$. Further, it can leads to the H{\"{o}}lder continuity with the exponent $\alpha$ defined in Eq. (\ref{HoelderRepresentation}):
\begin{equation}\label{HoelderRepresentation}
|f(x)-f(x+\delta)| \leq C^{\prime} 2^{-\alpha j} \leq C^{\prime \prime}|\delta|^{\alpha}.
\end{equation}
\end{theorem}
The Theorem \ref{HoelderContinuityInSmallPerturbation} gives an illustration of the Lipschitz continuity in the ``multi-resolution" space. The proof would be provided in the appendix. Such a theorem can be the theoretical support of the ``mathematical microscope" character of the wavelet. This property is beneficial to extract the detailed information of object during the recognition. Also, the regularity properties of wavelet base functions based on the Lipschitz constraint would assist the adversarial training method to boost the robustness of models. The related proof would be provided in the appendix.
\subsection{Multi-resolution Analysis Of Wavelet}
For the wavelet transform function $f$, there exist different resolutions. We know that the resolution $2^{j-1}$ is more fine-grained than the resolution $2^{j}$. The approximation with different resolutions can be the linear combination of the wavelet base $\psi_{j, k}$. A ladder of the space $\left(V_{j}\right)_{j \in \mathbb{Z}}$ can represent the successive resolution levels, where $V_{j}=\left\{f \in L^{2}(\mathbb{R})\right\}$ and $f$ is piecewise constant on the $\left[2^{\jmath} k, 2^{\jmath}(k+1)\right]$. These spaces have following properties:\\

1) Monotonicity: $
\cdots \subset V_{2} \subset V_{1} \subset V_{0} \subset V_{-1} \subset V_{-2} \subset \cdots.
$\\

2) Uniqueness: $
\bigcap_{j \in \mathbb{Z}} V_{j}=\{0\}$.\\

3) Density: $ \overline{\bigcup_{j \in \mathbb{Z}} V_{j}}=L^{2}(\mathbb{R})
$.\\

4) Multi-scale: $f \in V_{j} \leftrightarrow f\left(2^{j} .\right) \in V_{0}$.\\

5) Symmetry: $f \in V_{0} \rightarrow f(\cdot-n) \in V_{0}$ for all $n \in \mathbb{Z}$.\\

With the ``multi-resolution" character, the wavelet transform could do the projection of the signal to space with different wavelet base scales. Therefore, the signal could be decomposed in different resolutions. In the task of image classification, the coarse shape can be represented by the low resolution with a large scale,  and the texture detail can be represented by the high resolution with a small scale. In the task of object detection~\cite{SpatialParamidPooling}, the multi-scale technology may boost performance by generating feature maps with different scales.  For engineering practice, the multi-scale property is beneficial for almost all the tasks of computer vision. 
\subsection{Wavelet Transform On Images}
Eq. (\ref{WaveletBase}) derives the representation formula of the continuous wavelet transform. In reality, a variable $a$ would be discretized that $a=a_{0}^{m}$, for $m \in \mathbb{Z}$ while the variable of $b$ can be described as $b=n b_{0} a_{0}^{m}$. Then, a new discretized wavelet base can be formulated in Eq. (\ref{DiscreteWaveletBase}):
\begin{equation}\label{DiscreteWaveletBase}
\psi_{m, n}(x)=a_{0}^{-m / 2} \psi\left(\frac{x-n b_{0} a_{0}^{m}}{a_{0}^{m}}\right)=a_{0}^{-m / 2} \psi\left(a_{0}^{-m} x-n b_{0}\right).
\end{equation}
It leads to the discrete wavelet transform. The discrete wavelet coefficient $\left\langle f, \psi_{m, n}\right\rangle$ can approximately characterize the function $f$ or reconstruct $f$ in a numerically stable way from the $\left\langle f, \psi_{m, n}\right\rangle$. $f$ can also be written as a superposition of the combination $\psi_{m, n}$. In the framework of discrete wavelet transform, the reconstruction problem is defined in Eq. (\ref{DiscreteWaveletReconstruction}):
\begin{equation}\label{DiscreteWaveletReconstruction}
f=\sum_{m, n}\left\langle f, \psi_{m, n}\right\rangle \widetilde{\psi_{m, n}}.
\end{equation}
For the wavelet base function $\psi$, there exists a scaling function $\varphi$, the 2D representation formula can be defined in Eq. (\ref{2DWaveletScaleBaseFunction}):
\begin{equation}\label{2DWaveletScaleBaseFunction}
\begin{aligned}
\varphi_{j, m, n}(x, y) &=2^{j / 2} \varphi\left(2^{j} x-m, 2^{j} y-n\right), \\
\psi_{j, m, n}^{i}(x, y) &=2^{j / 2} \psi^{i}\left(2^{j} x-m, 2^{j} y-n\right),
\end{aligned}
\end{equation}
where index $i$ identifies the directional wavelets. The discrete wavelet transform of image $f(x,\ y)$ of size $M \times N$ is then defined in Eq. (\ref{DWT_2D}):
\begin{equation}\label{DWT_2D}
\begin{array}{l}
W_{\varphi}\left(j_{0}, m, n\right)=\frac{1}{\sqrt{M N}} \sum_{x=0}^{M-1} \sum_{y=0}^{N-1} f(x, y) \varphi_{j_{0}, m, n}(x, y), \\
W_{\psi}^{i}(j, m, n)=\frac{1}{\sqrt{M N}} \sum_{x=0}^{M-1} \sum_{y=0}^{N-1} f(x, y) \psi_{j, m, n}^{i}(x, y),
\end{array}
\end{equation}
where $j_{0}$ is an arbitrary starting scale, the scale $j$ satisfies $j \geq j_{0}$.
\section{Proposed Method}
\subsection{Haar Wavelet Function}
The Haar wavelet base function defined in Eq. (\ref{HaarFunction}) is found by the French mathematician A. Haar~\cite{HaarFunction}. Haar function is symmetrical, double-orthogonal, and compactly supported with the filter length of $2$ and vanishing movement of $1$. Utilizing the Haar function can help extract the pattern robustly without redundancy due to its orthogonality property as some classical research pointed out~\cite{HaarDetection}.  Its support size is $1$, which is beneficial for the operation of sparse compression. Moreover, since the Haar wavelet is symmetrical, it can be adapted to the reconstruction of the signal.
\begin{equation}\label{HaarFunction}
\psi(x)=\left\{\begin{array}{rl}
1 & 0 \leq x<\frac{1}{2}, \\
-1 & \frac{1}{2} \leq x<1, \\
0 & \text { otherwise.}
\end{array}\right.
\end{equation}
Its scaling function can be formulated as the form of Eq. (\ref{HaarScalingFunction}):
\begin{equation}\label{HaarScalingFunction}
\phi(t)=\left\{\begin{array}{ll}
1 & 0 \leq t<1, \\
0 & \text { otherwise. }
\end{array}\right.
\end{equation}
The filter function $h[n]$ is defined as the form of Eq. (\ref{HaarFilter}):
\begin{equation}\label{HaarFilter}
h[n]=:\left\{\begin{array}{ll}
\frac{1}{\sqrt{2}} & \text { if } \mathrm{n}=0,1, \\
0 & \text { otherwise. }
\end{array}\right.
\end{equation}
The Haar transform can be written as the form of Eq. (\ref{HaarTransform}):
\begin{equation}\label{HaarTransform}
\begin{aligned}
\frac{1}{\sqrt{2}} \psi\left(\left(\frac{t}{2}\right)\right) &=\sum_{n=-\infty}^{\infty}(-1)^{1-n} h[1-n] \phi(t-n) \\
&=\frac{1}{\sqrt{2}}(\phi(t-1)-\phi(t)).
\end{aligned}
\end{equation}

In image science, the utilization of wavelets helps concentrate the signal energy and control the noises of sub-bands. The property of the wavelet is also consistent with the logarithmic characteristics of the human visual system. Among various wavelet bases, the effectiveness of the simple Haar base function has been validated in the computer vision task~\cite{HaarDetection}. The default wavelet regularization method is based on the Haar function, however, its feasibility would be analyzed in a comparison experiment with diverse bases.

%The Haar function can be written as the transform of the round function defined in Eq. (\ref{RoundHaarFunction}). We define the extension of the Haar function to analyze its atomic Lipschitz constraint.
%\begin{equation}\label{RoundHaarFunction}
%\psi(x)=\left\{\begin{array}{cc}
%1-2\nint{x} & 0 \leqslant x<1 \\
%0 & \text { otherwise }
%\end{array}\right.
%\end{equation}
%\newtheorem{definition3}{Definition}[section]\label{AtomicLipschitzConstrainnt}
%\begin{definition}
%Let $\mathcal{A}_{k}^{p}$ be the the set of neural networks with an atomic Lipschitz bound of $k$ in $L_{p}$ space defined in Eq. (\ref{AtomicLipschitzConstraint}).
%\begin{equation}\label{AtomicLipschitzConstraint}
%\mathcal{A}_{k}^{p} \triangleq\left\{l_{W_{n}, b_{n}} \circ \cdots \circ \operatorname{Re} L U \circ l_{W_{1}, b_{1}} \mid \prod\left\|W_{i}\right\|_{p} \leq k, n \geq 2\right\}
%\end{equation}
%For the Haar function, the upper bound of Lipschitz constraint on the wavelet transform is formulated in Eq. (\ref{HaarUpperBound})
%\begin{equation}\label{HaarUpperBound}
%\|[2\ -2]\|_{\infty}\left\|\left[\begin{array}{l}
%2 \\
%2
%\end{array}\right]\right\|_{\infty}=8
%\end{equation}
%If the aggregation coefficient is $0.25$ and the average pooling kernel size is $4$, the Lipschitz constraint would be no larger than $0.5$ which means that it has a fine regularization property.
%\end{definition}

\subsection{Aggregated Wavelet Decomposition}
\begin{figure}[!t]
        \centering
\includegraphics[scale=0.25]{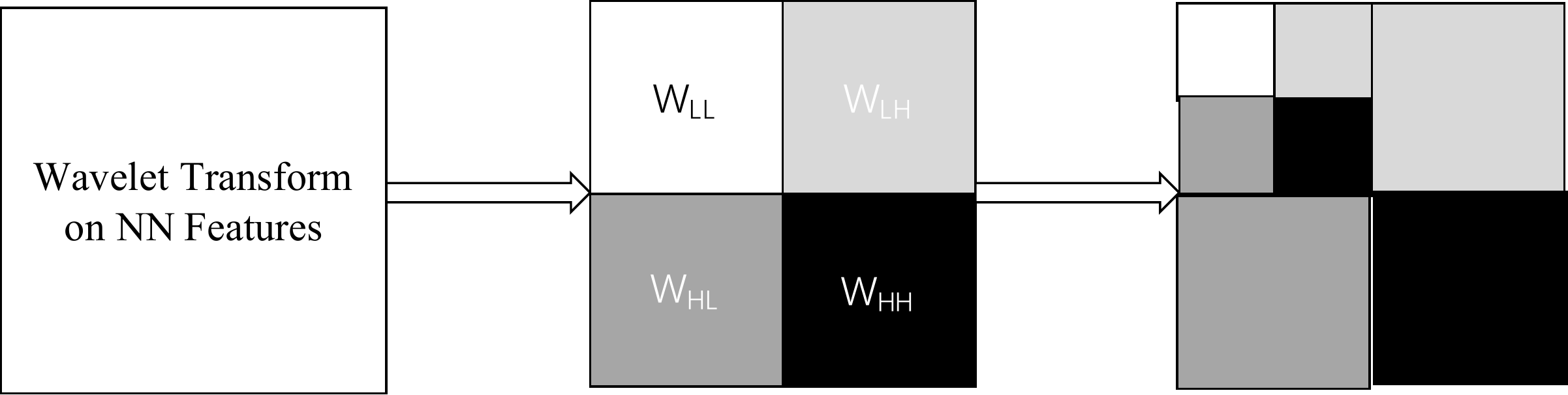}
\caption{The schematic diagram of the wavelet decomposition.}
\label{fig:WaveletDecomposition}
\end{figure}

Two-dimensional discrete wavelet transform (DWT) can be realized with the digital filter and down-sampling module in the reality. By performing a second-order decomposition in the wavelet domain according to the fast wavelet transform (FWT)~\cite{MallatWaveletDecomposition}, the DWT output of $j$-scale coefficient $W_{\varphi}[j, n]$ is $W_{\varphi}[j+1, k]$ with $n=2k$ as illustrated in Eq. (\ref{2nd-order-FWT}). Fig. \ref{fig:WaveletDecomposition} describes the wavelet decomposition process.
\begin{equation}\label{2nd-order-FWT}
\begin{array}{l}
W_{\varphi}[j+1, k]=\left.h_{\varphi}[-n] * W_{\varphi}[j, n]\right|_{n=2 k, k \leq 0}, \\
W_{\psi}[j+1, k]=\left.h_{\psi}[-n] * W_{\psi}[j, n]\right|_{n=2 k, k \leq 0}.
\end{array}
\end{equation}
\begin{figure}[!t]
\centering
\subfigure[]{
\centering
\includegraphics[width=0.15\hsize]{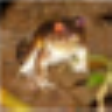}
%\caption{fig1}
}%
\subfigure[]{
\centering
\includegraphics[width=0.75\hsize]{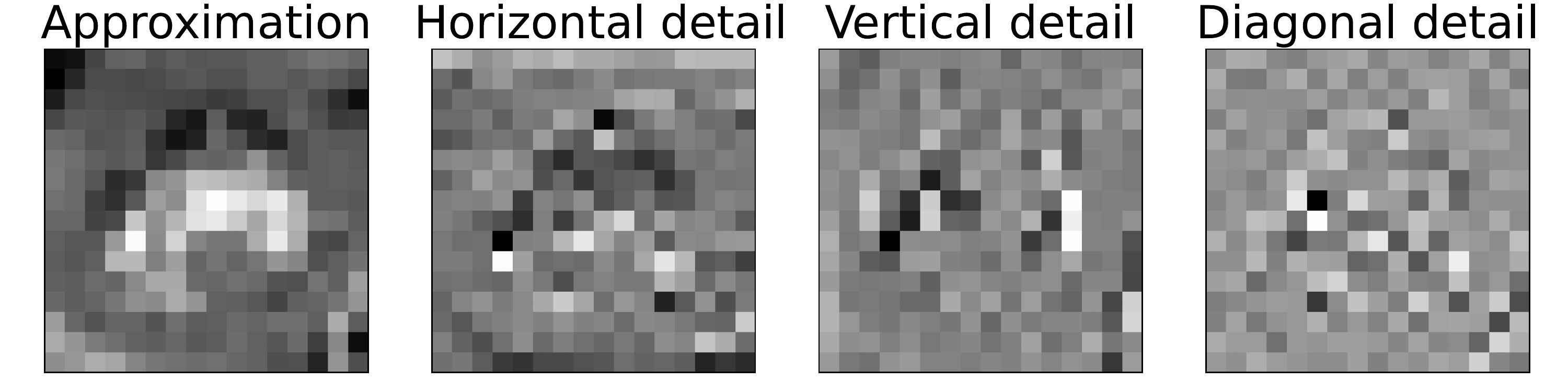}
%\caption{fig2}
}%
\caption{The Haar decomposition on the image sampled from CIFAR-10 dataset.
Fig. (a) denotes the first original CIFAR-10 image. 
Fig. (b) denotes its corresponding approximation, horizontal detail, vertical detail, and diagonal detail.}
\label{fig:cifarHaarDecomposition}
\end{figure}
On the F-Principle in deep learning~\cite{FourierRobustness}, biasing the high-frequency elements towards the low-frequency domain would be better than the filtering operation. Therefore, our wavelet decomposition strategy avoids discarding high-frequency elements. It is assumed that the generated matrix of DWT is $\mathbf{W_{LL}}$, $\mathbf{W_{LH}}$, $\mathbf{W_{HL}}$, and $\mathbf{W_{HH}}$, it corresponds to the approximation, horizontal detail, vertical detail, and diagonal detail of image sampled from the CIFAR-10 dataset as Fig. \ref{fig:cifarHaarDecomposition} demonstrates. The utilization of the multi-resolution property of the wavelet would be intuitively better on a wider frequency band than on a narrower frequency band. To follow the F-Principle and utilize the multi-resolution property efficiently, averaging the elements of the wavelet transform is more appropriate than filtering the high-frequency elements. Thus, the down-sampling operation is defined as the form of Eq. (\ref{WaveletAveragePooling}). We named it \textbf{Wavelet Average Pooling}:
\begin{equation}\label{WaveletAveragePooling}
\mathbf{W_{WaveletAveragePooling}} = (\mathbf{W_{LL}+ W_{LH} +W_{HL} + W_{HH}} )\cdot 0.25.
\end{equation}
\begin{figure*}[!t]
\centering
\subfigure[]{
\centering
\includegraphics[width=0.3\hsize]{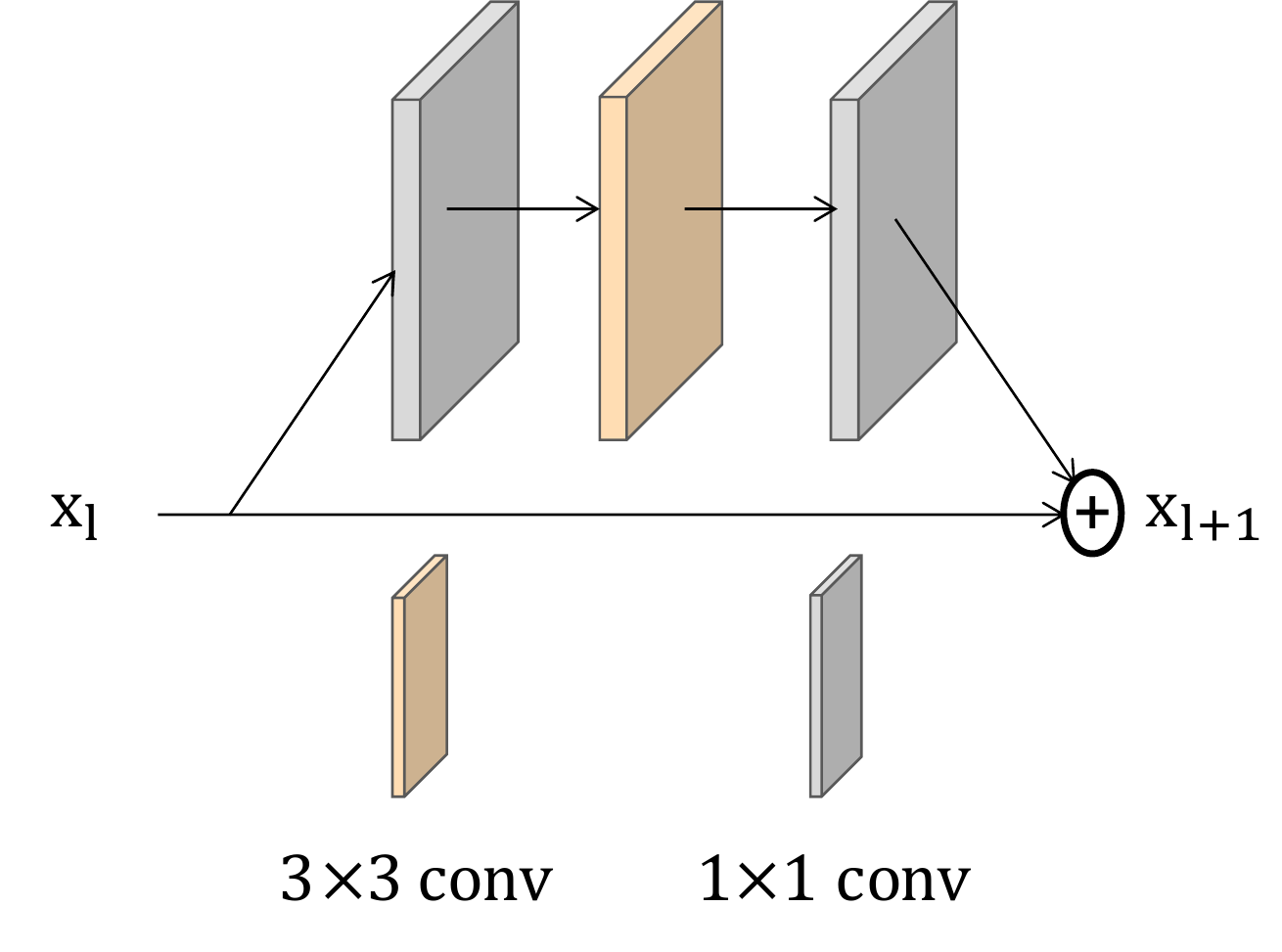}
%\caption{fig1}
}%
\subfigure[]{
\centering
\includegraphics[width=0.45\hsize]{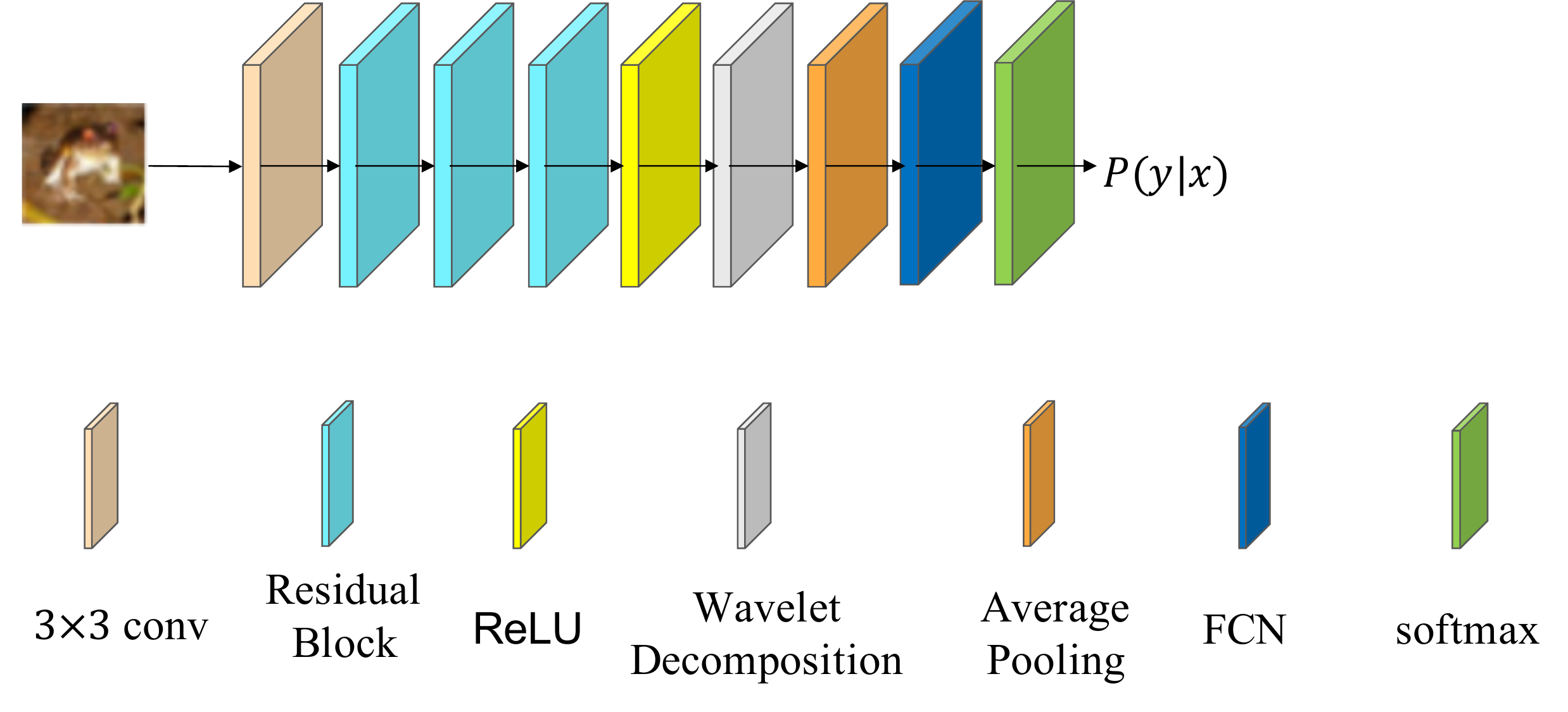}
%\caption{fig2}
}%
\caption{The structure of Wide Residual Neural Network on the Wavelet.
Fig. (a) illustrates the basic residual elements derived from ResNet~\cite{ResNet} and WideResNet~\cite{WideResNet}. 
Fig. (b) illustrates our proposed network structure.}
\label{fig:wideWaveletResNet}
\end{figure*}
\par{\textcolor{black}{The method proposed in the previous research~\cite{WaveletPooling} has been utilized to improve the normal generalization ability without adversarial perturbation. The FWT is utilized in the previous method for forward propagation, and the inverse FWT is operated as a reverse process in backpropagation. As a comparison, our \textbf{Wavelet Average Pooling} would only do the operation of down-sampling to regularize the neural network without the process of backpropagation. It has two advantages compared to the previous method~\cite{WaveletPooling}: 1) It is convenient in the computation; 2) Averaging on the condensed feature map could realize the trade-off between low-frequency elements related to the generalization without the perturbations and high-frequency elements related to adversarial perturbations.}}
\subsection{Robust Deep Wide Neural Network On Wavelet}
\par{\textcolor{black}{The overparameterization of WideResNet~\cite{WideResNet} is beneficial for robustness when the perturbation factor is considered in deep learning. It is analyzed in the theoretical view~\cite{GaussianProcessNN, WideResNetLossSurface, GDSearch, WideNetworkAsLinearModel, ConvergenceOperParameterization, CaoGeneralizationBoundWideNetwork,AllenZhuOverParameterization,ZouOverparameterization} and verified in a detailed empirical work~\cite{BagTricksAT}. The overparameterized deep wide neural networks could achieve a stable global minimum via the perspective of dynamical stability~\cite{WuOverparameterization}, NTK~\cite{AllenZhuOverParameterization}, and modeling of the Gaussian random initialization~\cite{ZouOverparameterization}. The generalization error of the wide residual neural network for the minimum-norm solution is comparable to the Monte Carlo rate due to the phenomenon of overparameterization~\cite{WeinanEOverParameterization}. If the square root of the parameter number of a shallow wide neural network exceeds the size of training data, the stochastic gradient descent with the random initialization can converge at a geometric rate to a nearby global optima~\cite{ModerateOverparameterization}. Moreover, via the tool of isoperimetry, the randomly-distributed points on the surface of a sphere need a smooth curve with the additional parameters for the fitting~\cite{UniversalLawRobustness}. It derives a universal law of robustness that good memorization needs overparameterized networks with a large width and depth. All of the theoretical findings support the motivation to explore the usage of wavelet regularization on the WideResNet model.}}

Although deep wide neural networks are beneficial to adversarial robustness, the explicit regularization methods are still required to achieve suitable perturbation stability~\cite{WuWRNRobustness}. Several regularization tricks in the time domain, e.g., early stopping and weight decay, were applied to the adversarial training and testified to be effective~\cite{BagTricksAT}. Our wavelet regularization method is implemented in the frequency domain.

Based on these motivations, we propose a wide deep residual neural network on the wavelet. We named it \textbf{WideWaveletResNet}. As Fig. \ref{fig:wideWaveletResNet} illustrates, after the propagation of the robustness against adversarial perturbation with three residual blocks and the ReLU module, our proposed wavelet decomposition module regularizes to do a down-sampling operation on the signal and satisfy the small Lipschitz constraint. After the operation of average pooling (Average Pooling) with the kernel size of 4, fully connected layer (FCN), and softmax layer, the inference prediction with confidence can be exported. The WideWaveletResNet with the module of wavelet regularization can promote robustness with the adversarial training method. The experiment result would validate the effectiveness of the wavelet regularization method. The lifting is more intuitive in a wide and deep neural network.

\section{Experiment And Results}
\subsection{Implementation Details}
We train our WideWaveletResNet model and test our proposed method based on the datasets of CIFAR-10 and CIFAR-100~\cite{CIFAR-10}. The adversarial perturbation is generated via the projected gradient descent (PGD)~\cite{PGD} algorithm with a step size of 10. The learning rate schedule is in multi-steps with an initial value of 0.1. The batch size is 128. The attack is in the $l_{\infty}$ norm constraint. The perturbation budget $\varepsilon$ is $0.031$ (approximately equal to $8/255$). The WideWaveletResNet model would be fit by the stochastic gradient descent with the momentum and weight decay of $0.0005$. The trick of ``Early Stopping" mechanism analyzed by the previous work~\cite{RobustOverfitting} is utilized. We follow some useful tricks analyzed in the previous research~\cite{BagTricksAT}. The training time of our \textbf{Adversarial Wavelet Training} method would take around 55h for 110 epochs on a server with a single NVIDIA-RTX 3090 GPU.
\begin{figure}[!t]
\centering
\subfigure[]{
\includegraphics[width=0.5\hsize]{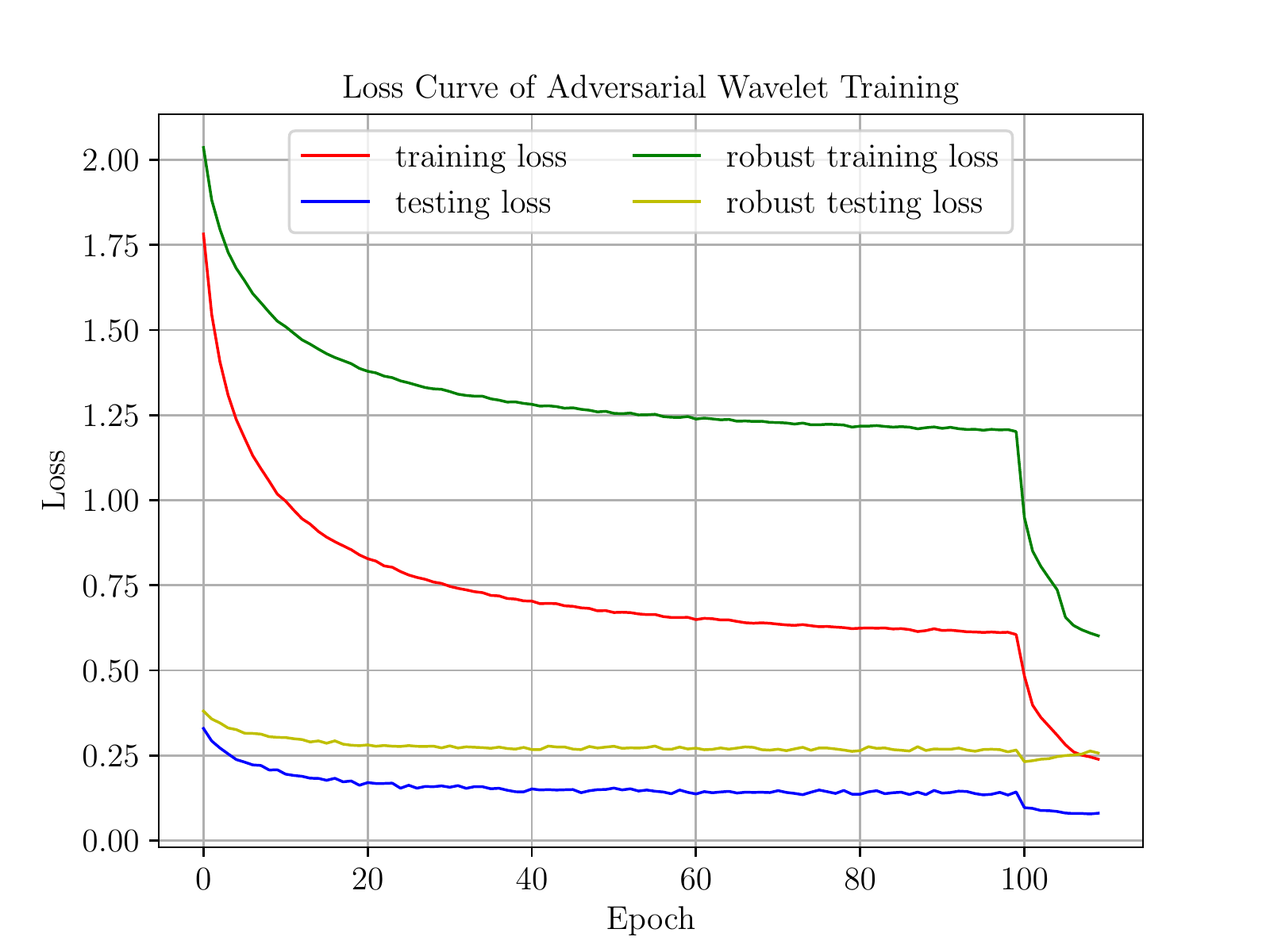}
%\caption{fig1}
}%
\subfigure[]{
\includegraphics[width=0.5\hsize]{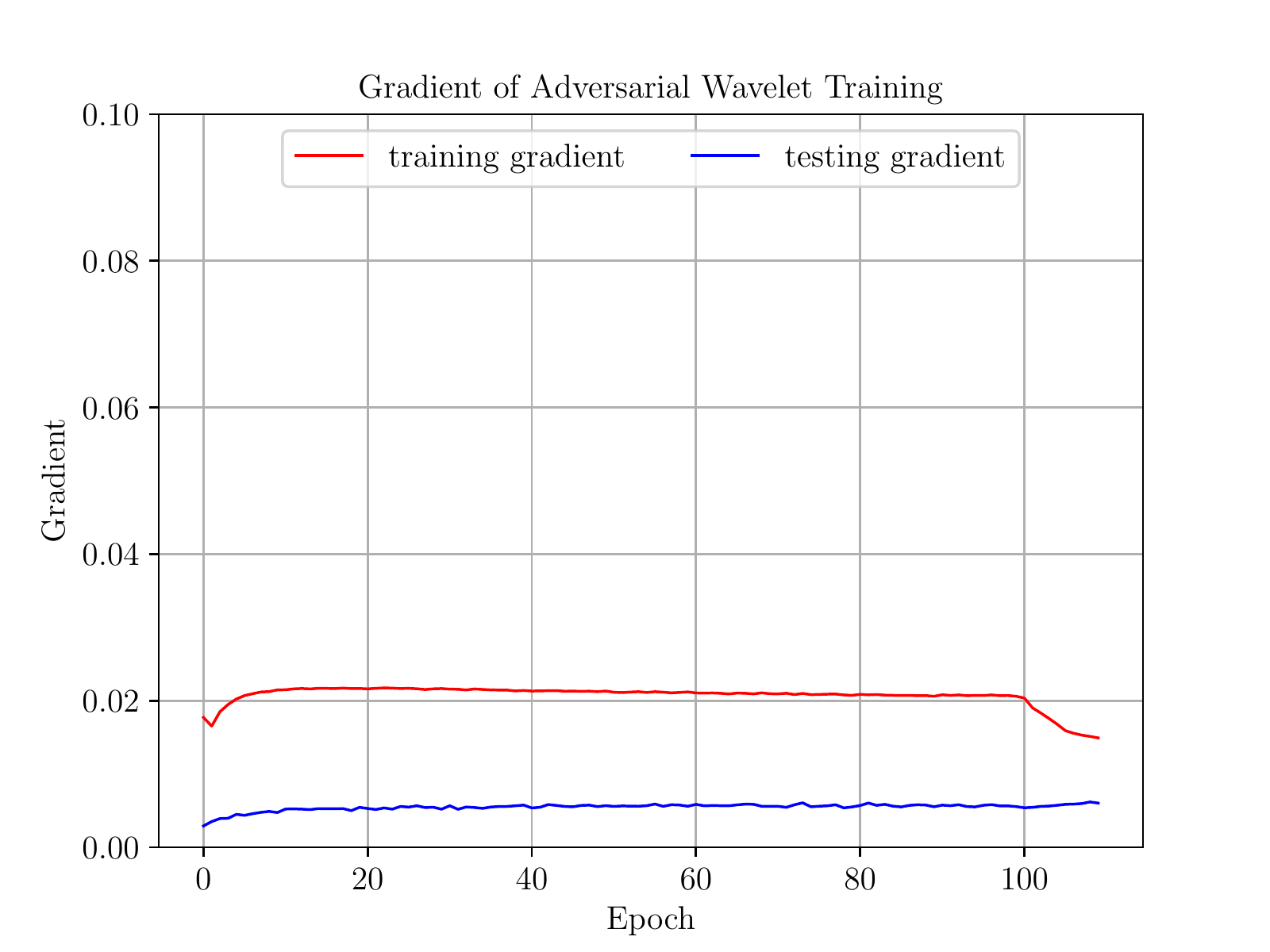}
%\caption{fig2}
}%
\caption{The loss curve and gradient curve of our adversarial training method on the CIFAR-10 dataset.
Fig. (a) illustrates the loss curve.
Fig. (b) illustrates the gradient curve.
}
\label{fig:trainingRecordCIFAR10}
\end{figure}
  \begin{figure}[!t]
\centering
\subfigure[]{
\includegraphics[width=0.5\hsize]{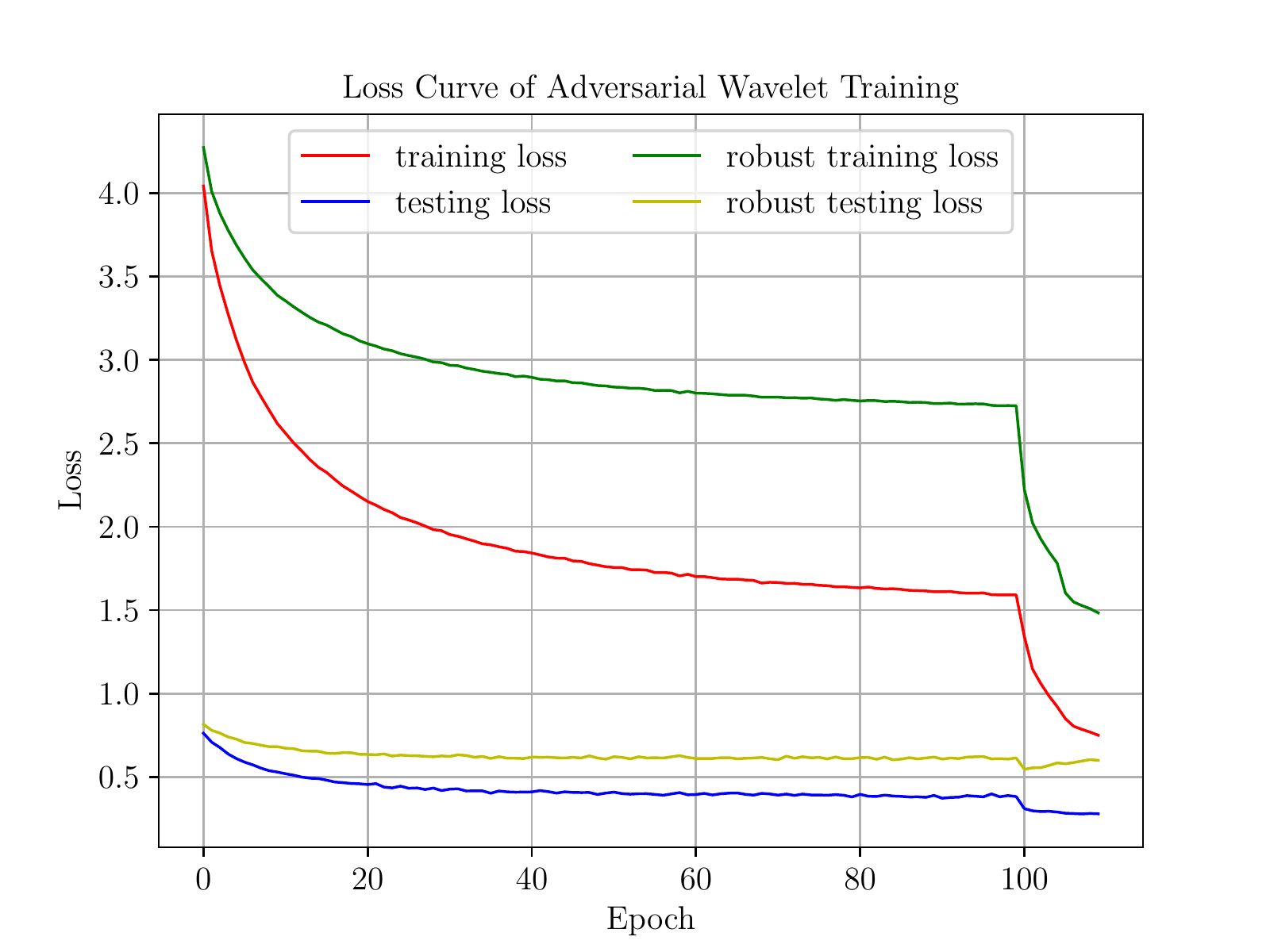}
%\caption{fig1}
}%
\subfigure[]{
\includegraphics[width=0.5\hsize]{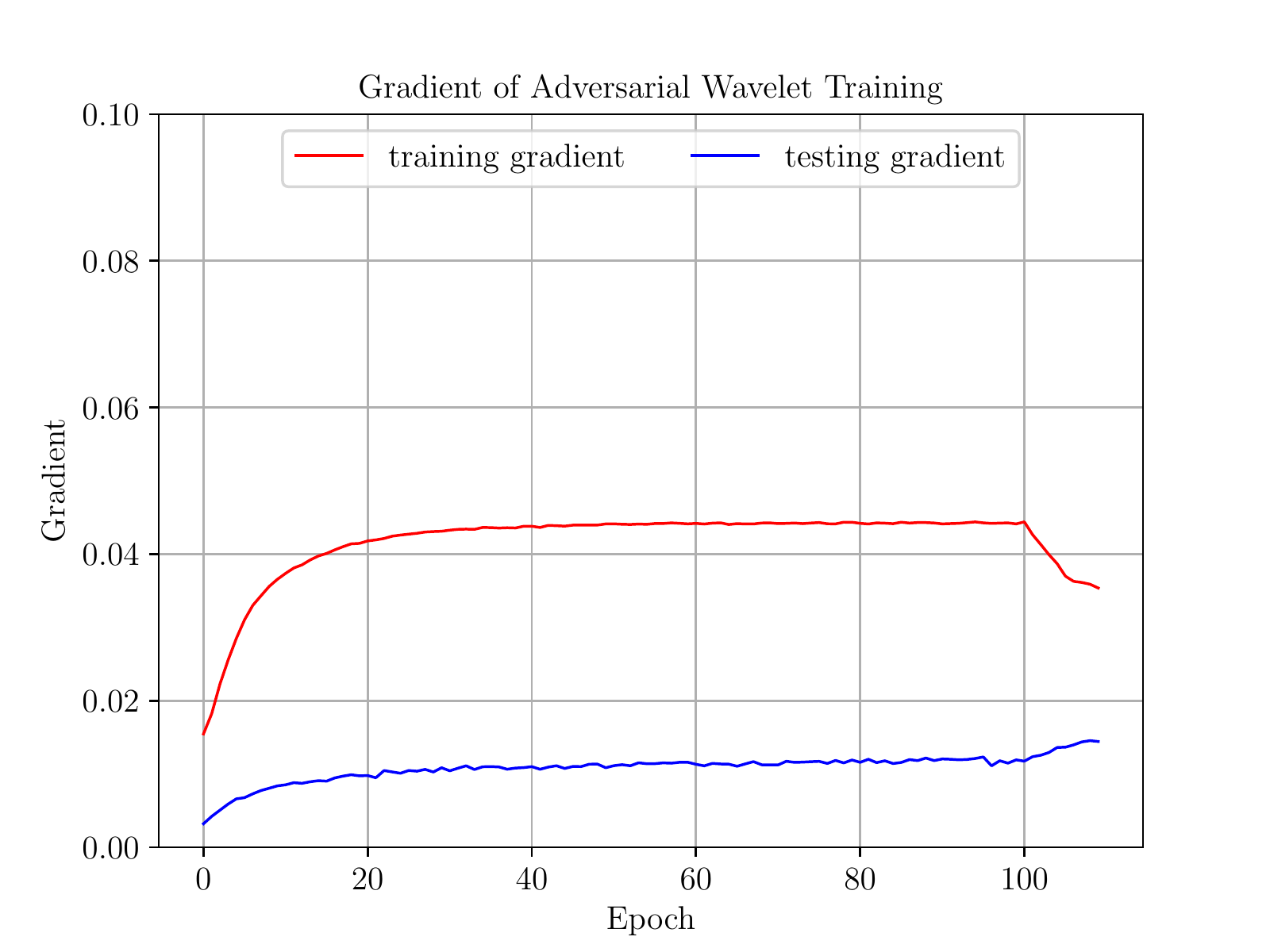}
%\caption{fig2}
}%
\caption{The loss curve and gradient curve of our adversarial training method on the CIFAR-100 dataset.
Fig. (a) illustrates the loss curve.
Fig. (b) illustrates the gradient curve.
}
\label{fig:trainingRecordCIFAR100}
\end{figure} 
   
Fig. \ref{fig:trainingRecordCIFAR10}-(a) and Fig. \ref{fig:trainingRecordCIFAR100}-(a) give the demonstrations of loss curves for our \textbf{Adversarial Wavelet Training} method on the CIFAR-10 and CIFAR-100 datasets. With the wavelet regularization on the wide residual neural network during the adversarial training process, the loss surfaces of our proposed method are smooth.  Empirically, the ``over-fitting" phenomenon can be avoided if the training loss curve is smooth enough. As Fig. \ref{fig:trainingRecordCIFAR10}-(b) and Fig. \ref{fig:trainingRecordCIFAR100}-(b) illustrate, the gradients of our proposed method are not too huge on both CIFAR-10 dataset and CIFAR-100 dataset. The adversarial wavelet training method boosts the robustness via the mechanism of the ``gradient penalty". Our wavelet regularization method restrains the gradient to realize a small Lipschitz constraint ($L$ is not large) on the basis of adversarial training. The experiment result further verifies that the wavelet regularization enhances the adversarial training method.
\subsection{Experiment Of Generalization And Robustness}
\begin{table*}[htbp]
\renewcommand{\arraystretch}{1.5}
% if using array.sty, it might be a good idea to tweak the value of
% \extrarowheight as needed to properly center the text within the cells
\caption{THE RECOGNITION PERFORMANCE (\%) OF DIFFERENT DEFENSE METHODS AGAINST WHITE-BOX ATTACKS ON CIFAR-10 ($\varepsilon=0.031$).}
\label{table:defenseCIFAR10}
\centering
\footnotesize
% Some packages, such as MDW tools, offer better commands for making tables
% than the plain LaTeX2e tabular which is used here.
\begin{tabular}{lccccccr@{.}}
\toprule
     & Adversarial Training?  &   Clen Acc & FGSM Acc &  PGD Acc &  MIM Acc &  C\&W Acc\\  % inserts table
 \midrule
Vanilla WideResNet-34-10~\cite{WideResNet}  &No & 90.04\thinspace$\pm$\thinspace 1.042     & 7.57\thinspace$\pm$\thinspace 2.209 & 0.00\thinspace$\pm$\thinspace 0.000 & 0.00\thinspace$\pm$\thinspace 0.004 & 0.00\thinspace$\pm$\thinspace 0.000\\
Wavelet CNN~\cite{WCNN}  &No &  89.09\thinspace$\pm$\thinspace 0.046 & 44.50\thinspace$\pm$\thinspace 0.240 & 24.35\thinspace$\pm$\thinspace 0.240 & 24.80\thinspace$\pm$\thinspace 0.334 & 24.55\thinspace$\pm$\thinspace 0.337\\
Scattering Network~\cite{DeepHybridNetwork}  &No & 91.33\thinspace$\pm$\thinspace 0.138 & 45.17\thinspace$\pm$\thinspace 0.937 & 23.09\thinspace$\pm$\thinspace 0.568 & 23.81\thinspace$\pm$\thinspace 0.600 & 23.36\thinspace$\pm$\thinspace 0.619\\
Deep Adaptive Wavelet Network~\cite{DeepAdaptiveWaveletNetwork}  &No &  93.08\thinspace$\pm$\thinspace 0.167 & 42.27\thinspace$\pm$\thinspace 0.982 & 9.94\thinspace$\pm$\thinspace 0.712 & 10.78\thinspace$\pm$\thinspace 0.528 & 10.35\thinspace$\pm$\thinspace 0.548\\
Classical Adversarial Training~\cite{PGD} & Yes & 78.24\thinspace$\pm$\thinspace 0.331 & 52.84\thinspace$\pm$\thinspace 0.442& 50.14\thinspace$\pm$\thinspace 0.293 & 49.22\thinspace$\pm$\thinspace 0.440 & 47.89\thinspace$\pm$\thinspace 0.189 \\
Fast Adversarial Training~\cite{FastAdversarialTraining} & Yes & 81.37\thinspace$\pm$\thinspace 0.207 & 53.09\thinspace$\pm$\thinspace 0.199 & 47.12\thinspace$\pm$\thinspace 0.131 & 47.45\thinspace$\pm$\thinspace 0.149 & 46.13\thinspace$\pm$\thinspace 0.123\\
Free Adversarial Training~\cite{FreeAdversarialTraining} & Yes & 82.61\thinspace$\pm$\thinspace 0.266 & 54.97\thinspace$\pm$\thinspace 0.258 & 49.60\thinspace$\pm$\thinspace 0.184 & 49.65\thinspace$\pm$\thinspace 0.151 & 48.65\thinspace$\pm$\thinspace 0.169 & \\
TRADES~\cite{TRADES}  &Yes & 75.38\thinspace$\pm$\thinspace 1.219 & 50.34\thinspace$\pm$\thinspace 1.100 & 46.90\thinspace$\pm$\thinspace 0.790 &  47.39\thinspace$\pm$\thinspace 0.813 & 44.27\thinspace$\pm$\thinspace 0.836 \\
%Self Adaptive Training~\cite{SelfAdaptiveTraining}  &Yes & 0.8235 & 0.5742 & 0.5292 & 0.5421 & 0.5125\\
Friendly Adversarial Training~\cite{FAT} & Yes & 86.47\thinspace$\pm$\thinspace 0.252 & 60.71\thinspace$\pm$\thinspace 0.150 & 51.56\thinspace$\pm$\thinspace 0.458 & 53.24\thinspace$\pm$\thinspace 0.404 & 51.84\thinspace$\pm$\thinspace 0.285\\
%Robust Overfitting~\cite{RobustOverfitting} & Yes & 0.8052 & 0.5421 & 0.5046 & 0.4996 & 0.4805\\
Adversarial Training Trick~\cite{BagTricksAT} & Yes & 87.25\thinspace$\pm$\thinspace 0.334 & 61.45\thinspace$\pm$\thinspace 0.339 & 54.45\thinspace$\pm$\thinspace 0.650 & 54.30\thinspace$\pm$\thinspace 0.451 & 53.88\thinspace$\pm$\thinspace 0.218 \\
Adt-Exp~\cite{AdversarialDistributionalTraining} & Yes & 87.35\thinspace$\pm$\thinspace 0.117 & 60.14\thinspace$\pm$\thinspace 0.531 & 50.98\thinspace$\pm$\thinspace 0.354 & 52.35\thinspace$\pm$\thinspace 0.328 & 52.01\thinspace$\pm$\thinspace 0.122 \\
Adt-Expam~\cite{AdversarialDistributionalTraining} & Yes & 87.66\thinspace$\pm$\thinspace 0.110 & 63.78\thinspace$\pm$\thinspace 0.349 & 52.64\thinspace$\pm$\thinspace 0.240 & 53.75\thinspace$\pm$\thinspace 0.251 & 53.16\thinspace$\pm$\thinspace 0.232\\
Adt-Impam~\cite{AdversarialDistributionalTraining} & Yes & 88.14\thinspace$\pm$\thinspace 0.139 & 67.06\thinspace$\pm$\thinspace 0.823 & 52.10\thinspace$\pm$\thinspace 0.108& 52.64\thinspace$\pm$\thinspace 0.253 & 52.61\thinspace$\pm$\thinspace 0.097 \\
Adversarial Wavelet Training (ours) & Yes & 87.34\thinspace$\pm$\thinspace 0.085 & 61.61\thinspace$\pm$\thinspace 0.361 & 54.76\thinspace$\pm$\thinspace 0.116 & 54.47\thinspace$\pm$\thinspace 0.602 & 54.25\thinspace$\pm$\thinspace 0.074 \\
         \bottomrule
\end{tabular}
\end{table*}
\begin{table*}[htbp]
\renewcommand{\arraystretch}{1.5}
% if using array.sty, it might be a good idea to tweak the value of
% \extrarowheight as needed to properly center the text within the cells
\caption{THE RECOGNITION PERFORMANCE (\%) OF DIFFERENT DEFENSE METHODS AGAINST WHITE-BOX ATTACKS ON CIFAR-100 ($\varepsilon=0.031$).}
\label{table:defenseCIFAR100}
\centering
\footnotesize
% Some packages, such as MDW tools, offer better commands for making tables
% than the plain LaTeX2e tabular which is used here.
\begin{tabular}{lccccccr@{.}}
\toprule
     & Adversarial Training?  &   Clen Acc & FGSM Acc &  PGD Acc &  MIM Acc &  C\&W Acc\\  % inserts table
 \midrule
%Vanilla WideResNet--40-10~\cite{WideResNet}  &No & 0.7851 & 0.1749 & 0.0004 & 0.0082 & 0.0003\\
Vanilla WideResNet-34-10~\cite{WideResNet}  &No & 68.68\thinspace$\pm$\thinspace 1.279 & 4.11\thinspace$\pm$\thinspace 0.545& 0.00\thinspace$\pm$\thinspace 0.000 & 0.00\thinspace$\pm$\thinspace 0.008 & 0.00\thinspace$\pm$\thinspace 0.000\\
Wavelet CNN~\cite{WCNN}  &No & 59.85\thinspace$\pm$\thinspace 0.237 & 18.43\thinspace$\pm$\thinspace 0.230 & 9.71\thinspace$\pm$\thinspace 0.163 & 9.92\thinspace$\pm$\thinspace 0.162 & 9.52\thinspace$\pm$\thinspace 0.183\\
Scattering Network~\cite{DeepHybridNetwork}  &No & 68.98\thinspace$\pm$\thinspace 0.204 & 21.29\thinspace$\pm$\thinspace 0.677 & 9.71\thinspace$\pm$\thinspace 0.304 & 10.12\thinspace$\pm$\thinspace 0.307 & 9.67\thinspace$\pm$\thinspace 0.319\\
Deep Adaptive Wavelet Network~\cite{DeepAdaptiveWaveletNetwork}  &No & 71.52\thinspace$\pm$\thinspace 0.191 & 17.51\thinspace$\pm$\thinspace 0.569 & 4.63\thinspace$\pm$\thinspace 0.160 & 4.86\thinspace$\pm$\thinspace 0.134 & 4.83\thinspace$\pm$\thinspace 0.119\\
%Self Adaptive Training~\cite{SelfAdaptiveTraining}  &Yes & 0.5694 & 0.3470 & 0.3290 & 0.3336 & 0.2938\\
Classical Adversarial Training~\cite{PGD} & Yes & 59.83\thinspace$\pm$\thinspace 0.217 & 31.85\thinspace$\pm$\thinspace 0.502& 29.44\thinspace$\pm$\thinspace 0.202 & 27.84\thinspace$\pm$\thinspace 0.318 & 28.03\thinspace$\pm$\thinspace 0.047 \\
Fast Adversarial Training~\cite{FastAdversarialTraining} & Yes & 62.40\thinspace$\pm$\thinspace 0.075 & 32.01\thinspace$\pm$\thinspace 0.135& 26.23\thinspace$\pm$\thinspace 0.175 & 25.92\thinspace$\pm$\thinspace 0.020 & 26.22\thinspace$\pm$\thinspace 0.085 \\
Free Adversarial Training~\cite{FreeAdversarialTraining} & Yes & 60.65\thinspace$\pm$\thinspace 0.211 & 31.21\thinspace$\pm$\thinspace 0.411& 27.01\thinspace$\pm$\thinspace 0.092 & 26.54\thinspace$\pm$\thinspace 0.330 & 26.47\thinspace$\pm$\thinspace 0.224 \\
Adt-Exp~\cite{AdversarialDistributionalTraining} & Yes & 63.80\thinspace$\pm$\thinspace 0.324 & 34.43\thinspace$\pm$\thinspace 0.430 & 27.97\thinspace$\pm$\thinspace 0.132 & 28.75\thinspace$\pm$\thinspace 0.132 & 28.40\thinspace$\pm$\thinspace 0.175\\
Adt-Expam~\cite{AdversarialDistributionalTraining} & Yes & 63.68\thinspace$\pm$\thinspace 0.212 & 36.33\thinspace$\pm$\thinspace 0.350 & 29.02\thinspace$\pm$\thinspace 0.159 & 29.66\thinspace$\pm$\thinspace 0.160 & 28.99\thinspace$\pm$\thinspace 0.171\\
Adt-Impam~\cite{AdversarialDistributionalTraining} & Yes & 63.71\thinspace$\pm$\thinspace 0.271 & 38.10\thinspace$\pm$\thinspace 0.151 & 28.25\thinspace$\pm$\thinspace 0.335 & 28.59\thinspace$\pm$\thinspace 0.311 & 28.39\thinspace$\pm$\thinspace 0.435 \\
Adversarial Wavelet Training (ours) & Yes & 61.47\thinspace$\pm$\thinspace 0.416 & 34.80\thinspace$\pm$\thinspace 0.394 & 32.19\thinspace$\pm$\thinspace 0.166 & 30.56\thinspace$\pm$\thinspace 0.135 & 30.47\thinspace$\pm$\thinspace 0.142 \\
         \bottomrule
\end{tabular}
\end{table*}
We compare our proposed model with several different models. On CIFAR-10, four models are in the natural training settings: vanilla wide residual network (Vanilla WideResNet)~\cite{WideResNet}, wavelet convolutional neural network (WCNN)~\cite{WCNN}, scattering network~\cite{DeepHybridNetwork}, and deep adaptive wavelet network~\cite{DeepAdaptiveWaveletNetwork}. Other eight methods are trained with the WideResNet models in the adversarial training settings: Classical Adversarial Training~\cite{PGD}, Fast Adversarial Training~\cite{FastAdversarialTraining}, Free Adversarial Training~\cite{FreeAdversarialTraining}, adversarial training with the surrogate loss function (TRADES, regularization coefficient $\beta=6.0$)~\cite{TRADES}, friendly adversarial training (FAT)~\cite{FAT}, adversarial training trick method~\cite{BagTricksAT}, the method that the explicit adversarial distribution is defined by the distribution transformation with activation function (Adt-Exp)~\cite{AdversarialDistributionalTraining}, the method that the explicit adversarial distribution is generated by the neural network (Adt-Expam)~\cite{AdversarialDistributionalTraining}, the method that the implicit adversarial distribution is generated by the neural network (Adt-Impam)~\cite{AdversarialDistributionalTraining}. The models derived from the work of adversarial distributional training~\cite{AdversarialDistributionalTraining} are the WideResNets with a width size of 10 and a depth of 28 layers. Other adversarial training models are WideResNets with a width of 10 and a depth of 34 layers (WideResNet-34-10). Vanilla WideResNet is in the same structure setting as these adversarial training methods. \textcolor{black}{The Fast Gradient Sign Method (FGSM)~\cite{FGSM}, Projected Gradient PGD method~\cite{PGD}, Momentum Iteration Method (MIM)~\cite{MIM}, and C\&W method~\cite{C&W} are utilized to validate the robustness with the perturbation budget $\varepsilon=0.031$. For multiple-step attack methods (PGD, MIM, C\&W), the perturbed number of steps is 20, and the perturbation step size is 2. The decay factor of the MIM attack is $1.0$. The PGD attack and adaptive C\&W attack would be restarted after $20$ steps. All the multiple-step adversarial examples are generated with random initialization. When the perturbation budget $\varepsilon$ is 0, the clean accuracy (Acc) can be seen as the generalization ability under the natural testing setting. }The experiment results on CIFAR-10 is illustrated in Table \ref{table:defenseCIFAR10}. On the dataset of CIFAR-100, similar methods are compared. 
The result on CIFAR-100 is demonstrated in Table \ref{table:defenseCIFAR100}.

As Table \ref{table:defenseCIFAR10} illustrates, most natural training methods have high clean accuracy on the CIFAR-10 dataset. It is accepted that adversarially-trained models degrade a little on the metrics of clean accuracy due to discarding some model weights unrelated to the robustness~\cite{RobustnessOddsAcc}. Our wavelet regularization method of Wavelet Average Pooling helps boost the robustness and enhance adversarial training. The evaluation result on CIFAR-100 can be seen from Table \ref{table:defenseCIFAR100}. On the CIFAR-100 dataset, our model also realizes considerable adversarial robustness. Although our defense model does not perform best on the simpler single-step FGSM attack (compared with the adversarial distributional training method~\cite{AdversarialDistributionalTraining}), the robust accuracy metrics of our proposed method on other multi-step adversarial attacks are better.

\begin{figure}[!t]
\centering
\subfigure[]{
\includegraphics[width=0.15\hsize]{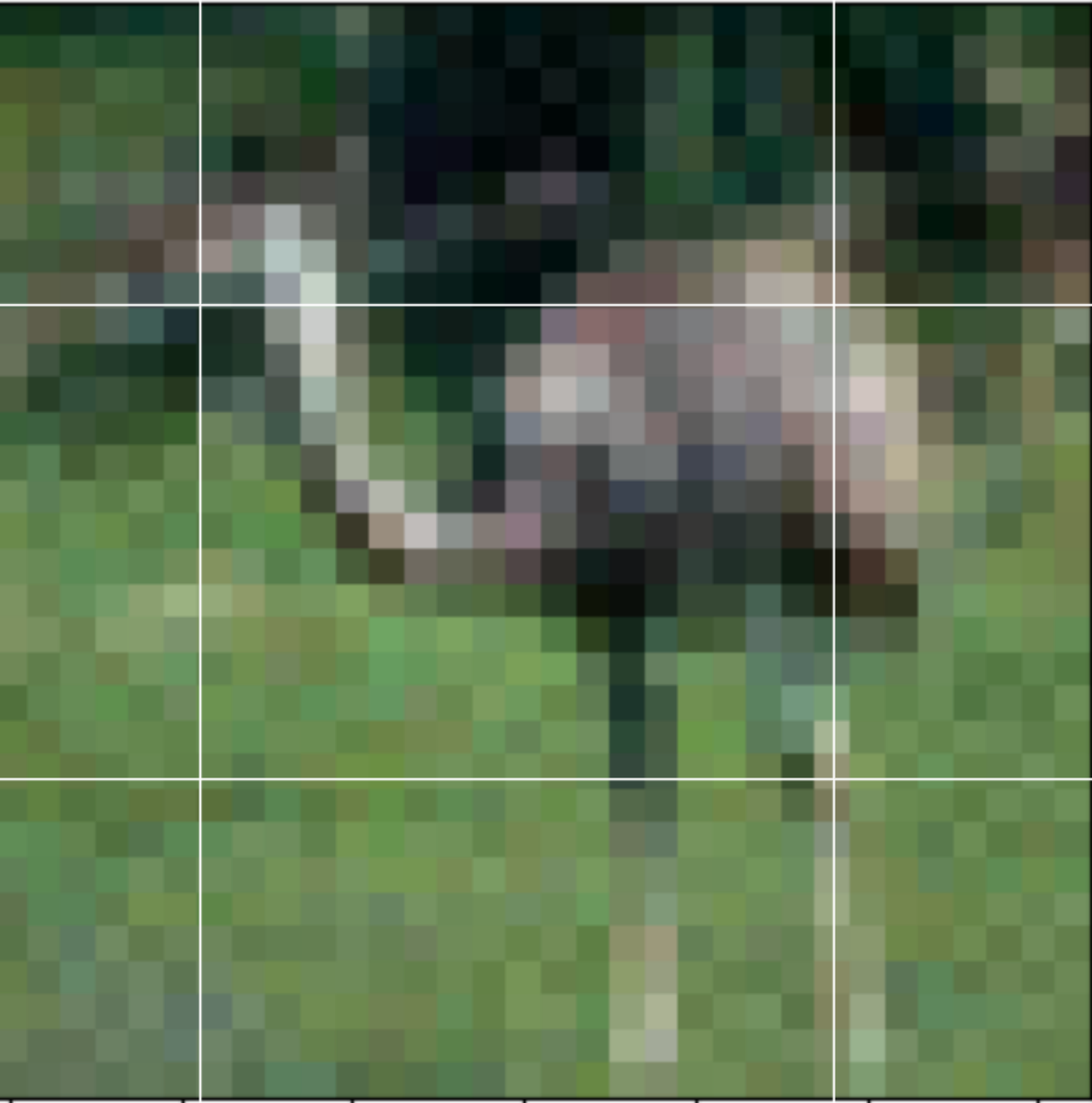}
%\caption{fig1}
}%
\subfigure[]{
\includegraphics[width=0.15\hsize]{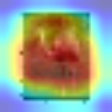}
%\caption{fig2}
}%
\quad 
\subfigure[]{
\includegraphics[width=0.15\hsize]{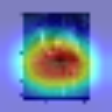}
}%'\quad 
\subfigure[]{
\includegraphics[width=0.15\hsize]{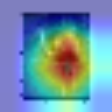}
}%
\caption{Grad-CAM visualization of the heat map of the features on the adversarial sample image from the CIFAR-10 dataset. 
Figure. (a) illustrates an adversarial page of ostrich, which is generated by PGD attack.
Figure. (b) illustrates the extracted feature heat map of vanilla WideResNet model~\cite{WideResNet}.
Figure. (c) illustrates the extracted feature heat map of the model trained by the adversarial training trick method~\cite{BagTricksAT}.
Figure. (d) illustrates the extracted feature heat map of our proposed model. }
\label{fig:gradCAM}
\end{figure}

To prove the interpretability of the learning features of our proposed models, the Grad-CAM~\cite{GradCAMJournal} output on the feature heat map before fully connected layers would be visualized. As Fig. \ref{fig:gradCAM} illustrates, the learnable features of our proposed models on the adversarial image of the ostrich are best. The previous two methods ignore the legs of the ostrich. In contrast, the heatmap generated by our proposed method would focus on the meaningful part of the ostrich. It seems that our model learns interpretable features of ostrich. Nevertheless, the feature heat maps derived from the vanilla WideResNet model~\cite{WideResNet} and the model trained by the adversarial training trick method~\cite{BagTricksAT} are on unrelated regions. 
\textcolor{black}{\subsection{Defense Experiment Against Black-Box Attacks}}
\textcolor{black}{The performance of our proposed method under the black-box attacks is also evaluated. The utilized attack method is the black-box attack method based on the natural evolutionary strategies (NES attack)~\cite{IlyasTransferAttack}. The perturbation budget of NES attack satisfies $\varepsilon=0.05$. The coefficient of gradient estimation satisfies $fd\_eta=2.55$. The query number cannot surpass $10000$. The learning rate is $2.55$. The attack failure rate (AFR) is the metric of robustness to evaluation the ratio of successful defense. The misclassified examples in the natural settings without adversarial perturbations would not be taken into considerations in the black-box attack experiment. The average query number (AQN) can measure the attack efficiency. The experiment is evaluated on the CIFAR-10 dataset. Five list methods in Table \ref{table:blackBoxAttack} are tested.}
\begin{table}[!t]
\renewcommand{\arraystretch}{1.5}
% if using array.sty, it might be a good idea to tweak the value of
% \extrarowheight as needed to properly center the text within the cells
\caption{PERFORMANCE OF DIFFERENT DEFENSE METHODS AGAINST NES ATTACKS ON CIFAR-10 ($\varepsilon=0.05$).}
\label{table:blackBoxAttack}
\centering
\footnotesize
% Some packages, such as MDW tools, offer better commands for making tables
% than the plain LaTeX2e tabular which is used here.
\begin{tabular}{lccr@{.}}
\toprule
     & AFR(\%) &  AQN\\  % inserts table
 \midrule
%Vanilla WideResNet--40-10~\cite{WideResNet}  &No & 0.7851 & 0.1749 & 0.0004 & 0.0082 & 0.0003\\
Classical Adversarial Training~\cite{PGD}  &79.40\thinspace$\pm$\thinspace 3.816 & 676\thinspace$\pm$\thinspace 55\\
Fast Adversarial Training~\cite{FastAdversarialTraining}  &80.69\thinspace$\pm$\thinspace 1.313 & 669\thinspace$\pm$\thinspace 31\\
Free Adversarial Training~\cite{FreeAdversarialTraining}  &80.80\thinspace$\pm$\thinspace 2.697 & 694\thinspace$\pm$\thinspace 18\\
Adversarial Training Trick~\cite{BagTricksAT} & 76.94\thinspace$\pm$\thinspace 2.273 & 579\thinspace$\pm$\thinspace 13\\
Adversarial Wavelet Training (ours) & 82.79\thinspace$\pm$\thinspace 3.089 & 605\thinspace$\pm$\thinspace 19\\
         \bottomrule
\end{tabular}
\end{table}

\textcolor{black}{NES attack~\cite{IlyasTransferAttack} is based on the strategy of gradient estimation in which the adversary cannot obtain the information of model structure. As Table \ref{table:blackBoxAttack} illustrates, both AFR and AQN in the statistical test have a large standard deviation due to the instability of black-box attacks. In such a scenario, our proposed method could realize considerable adversarial robustness among the compared methods under the NES attacks.}
\textcolor{black}{\subsection{Defense Experiment Against Adaptive Attacks}}
\textcolor{black}{An effective defense method should avoid the phenomenon of obfuscated gradient. Therefore, besides C\&W attacks~\cite{C&W}, the evaluation under the attacks of Backward Pass Differentiable Approximation (BPDA)~\cite{BPDA} would help avoid a false sense of security. The learning rate of BPDA attack is $0.5$, and the maximum iteration number is $100$. The robust accuracy under the range of perturbation budgets $[0,\ 0.031,\ 0.062,\ 0.093,\ 0.124,\ 0.155]$ would be evaluated on the CIFAR-10 dataset.}
 \begin{figure}[!t]
        \centering
\includegraphics[scale=0.45]{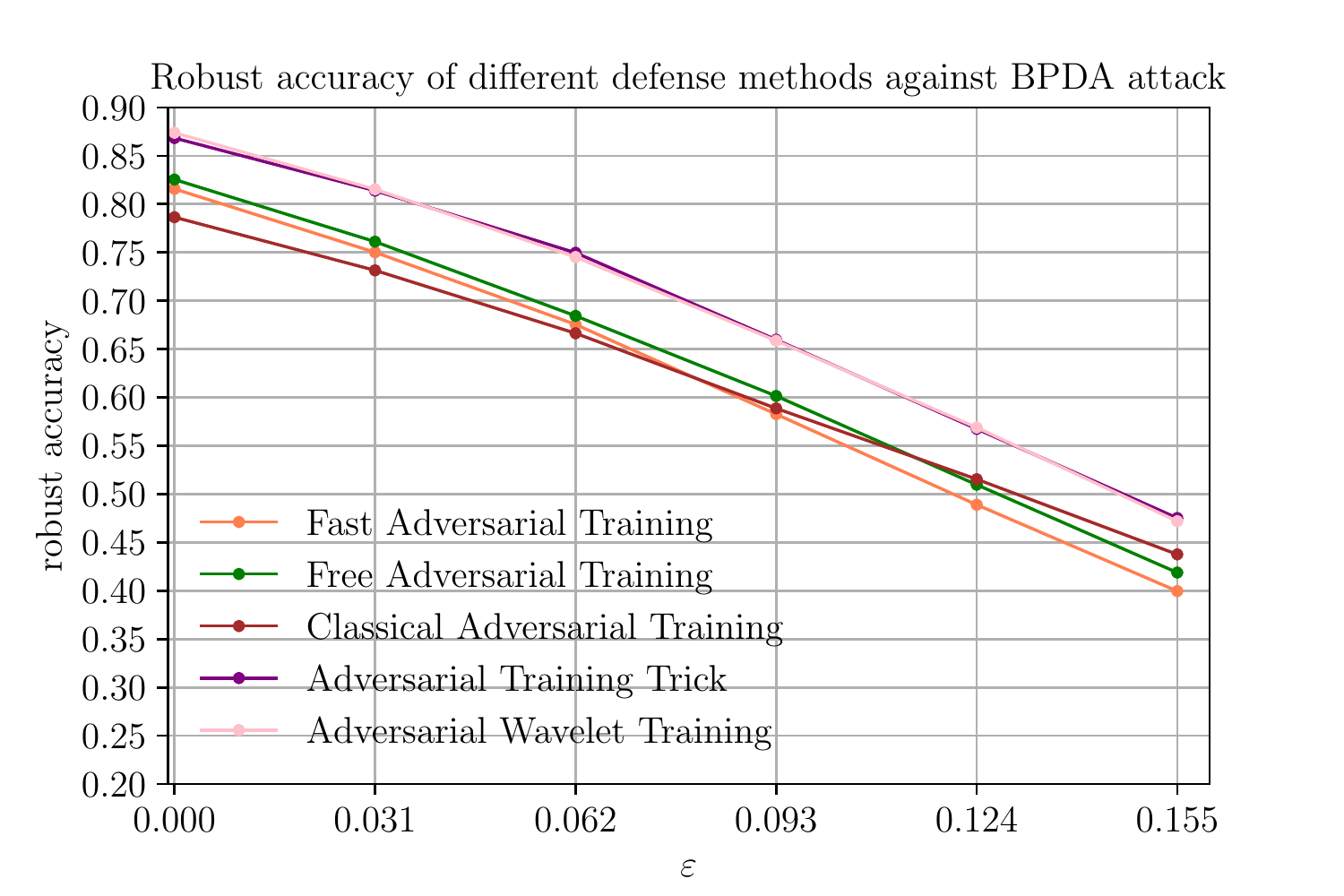}
\caption{The robust accuracy of different defense methods under the BPDA attacks~\cite{BPDA}.}
\label{fig:DPDADefense}
\end{figure}

\textcolor{black}{As can be seen from the Fig. \ref{fig:DPDADefense}, the performance of our method is close to the performance of the adversarial training trick method~\cite{BagTricksAT}. When the perturbation budget increases from $0$ to $0.155$, the robust accuracy of some other defense methods~\cite{PGD, FastAdversarialTraining, FreeAdversarialTraining} drops near 40\%. In the settings with the perturbation budget $\varepsilon=0.155$, the robust accuracy of our method against BPDA attack is 47.19\%. Our method would not be troubled by the phenomenon of obfuscated gradient.}
\textcolor{black}{\subsection{Defense Experiment Against Auto Attacks}}
\textcolor{black}{Due to the sub-optimal step size and problems of the objective function existed in the PGD attacks, the auto attack methods~\cite{AutoAttack} can be seen as a necessary complement of robustness evaluation. Two attack methods described in the seminal work~\cite{AutoAttack} would be evaluated, including the size-free PGD attack method based on the cross-entropy loss (APGD-CE) and size-free PGD attack method based on the targeted difference of logits ratio (DLR) loss (APGD-T).}
\begin{table}[!t]
\renewcommand{\arraystretch}{1.5}
% if using array.sty, it might be a good idea to tweak the value of
% \extrarowheight as needed to properly center the text within the cells
\caption{THE RECOGNITION PERFORMANCE (\%) OF DIFFERENT DEFENSE METHODS AGAINST AUTO ATTACKS ON CIFAR-10 ($\varepsilon=0.031$).}
\label{table:autoAttackCIFAR10}
\centering
\footnotesize
% Some packages, such as MDW tools, offer better commands for making tables
% than the plain LaTeX2e tabular which is used here.
\begin{tabular}{lccr@{.}}
\toprule
     & APGD-CE Acc &  APGD-T Acc\\  % inserts table
 \midrule
%Vanilla WideResNet--40-10~\cite{WideResNet}  &No & 0.7851 & 0.1749 & 0.0004 & 0.0082 & 0.0003\\
Classical Adversarial Training~\cite{PGD}  &49.61\thinspace$\pm$\thinspace 0.323 & 46.04\thinspace$\pm$\thinspace 0.253\\
Fast Adversarial Training~\cite{FastAdversarialTraining}   &46.49\thinspace$\pm$\thinspace 0.141 & 43.53\thinspace$\pm$\thinspace 0.132\\
Free Adversarial Training~\cite{FreeAdversarialTraining}  &48.96\thinspace$\pm$\thinspace 0.164 & 46.41\thinspace$\pm$\thinspace 0.105\\
Adversarial Training Trick~\cite{BagTricksAT} & 53.63\thinspace$\pm$\thinspace 0.724 & 51.55\thinspace$\pm$\thinspace 0.239\\
Adversarial Wavelet Training (ours) & 54.10\thinspace$\pm$\thinspace 0.138 & 51.77\thinspace$\pm$\thinspace 0.131\\
         \bottomrule
\end{tabular}
\end{table}
\begin{table}[!t]
\renewcommand{\arraystretch}{1.5}
% if using array.sty, it might be a good idea to tweak the value of
% \extrarowheight as needed to properly center the text within the cells
\caption{THE RECOGNITION PERFORMANCE (\%) OF DIFFERENT DEFENSE METHODS AGAINST AUTO ATTACKS ON CIFAR-100 ($\varepsilon=0.031$).}
\label{table:autoAttackCIFAR100}
\centering
\footnotesize
% Some packages, such as MDW tools, offer better commands for making tables
% than the plain LaTeX2e tabular which is used here.
\begin{tabular}{lccr@{.}}
\toprule
     & APGD-CE Acc  &  APGD-T Acc\\  % inserts table
 \midrule
%Vanilla WideResNet--40-10~\cite{WideResNet}  &No & 0.7851 & 0.1749 & 0.0004 & 0.0082 & 0.0003\\
Classical Adversarial Training~\cite{PGD}  &28.90\thinspace$\pm$\thinspace 0.200 & 
25.77\thinspace$\pm$\thinspace 0.073\\
Fast Adversarial Training~\cite{FastAdversarialTraining}   &25.54\thinspace$\pm$\thinspace 0.245 & 23.23\thinspace$\pm$\thinspace 0.130\\
Free Adversarial Training~\cite{FreeAdversarialTraining}  &26.53\thinspace$\pm$\thinspace 0.090 & 
24.21\thinspace$\pm$\thinspace 0.128\\
Adversarial Wavelet Training (ours) & 31.56\thinspace$\pm$\thinspace 0.176 & 28.11\thinspace$\pm$\thinspace 0.133\\
         \bottomrule
\end{tabular}
\end{table}

\textcolor{black}{The robustness evaluation result illustrated in Table \ref{table:autoAttackCIFAR10} (CIFAR-10) and Table \ref{table:autoAttackCIFAR100} (CIFAR-100) shows the considerable performance of our methods. The evaluation pipeline of auto attack further validates the feasible robustness of our proposed method.}
\textcolor{black}{\subsection{Ablation Study}}
\begin{figure}[!t]
\centering
\subfigure[]{
\centering
\includegraphics[width=0.45\hsize]{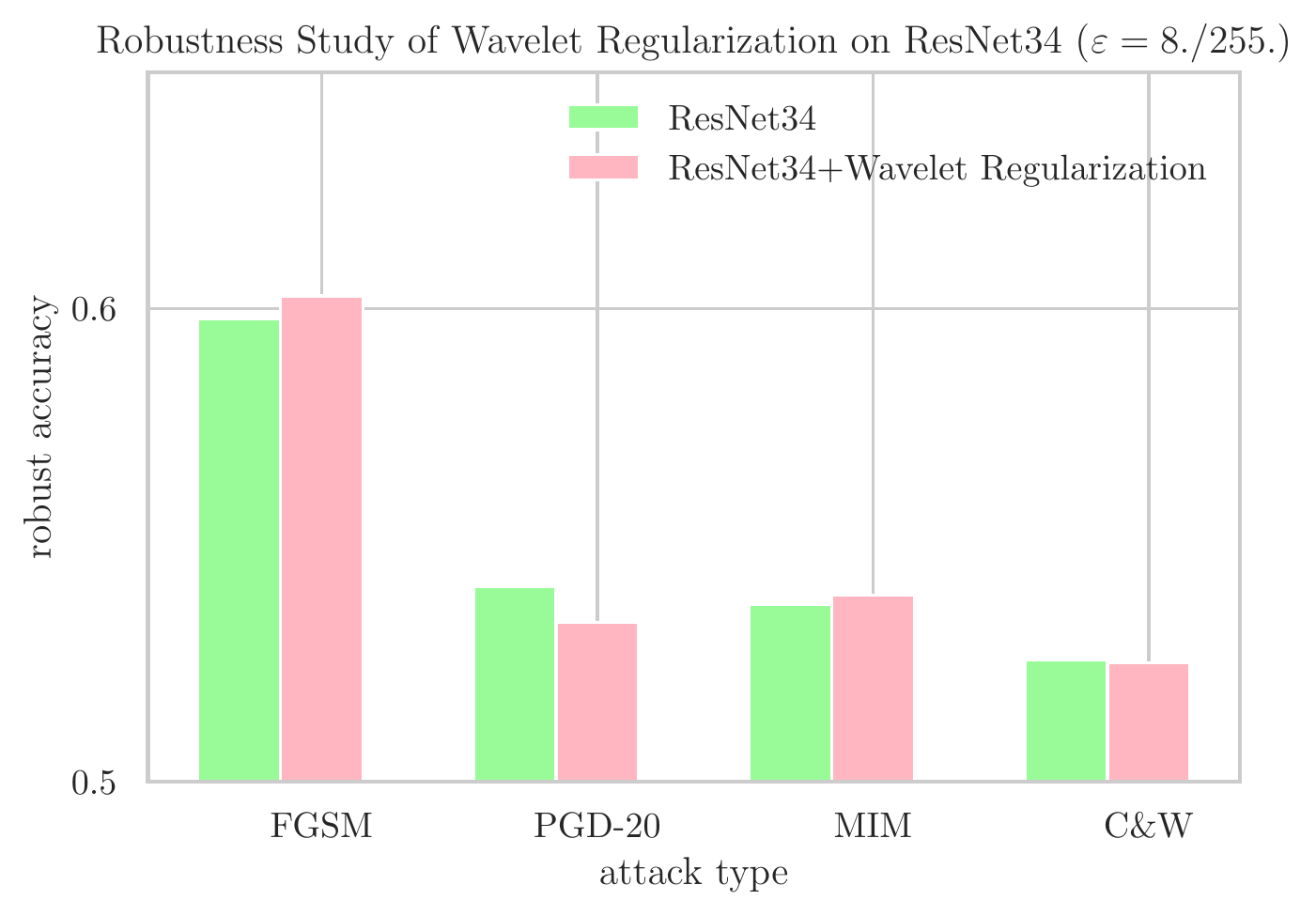}
%\caption{fig1}
}%
\subfigure[]{
\centering
\includegraphics[width=0.48\hsize]{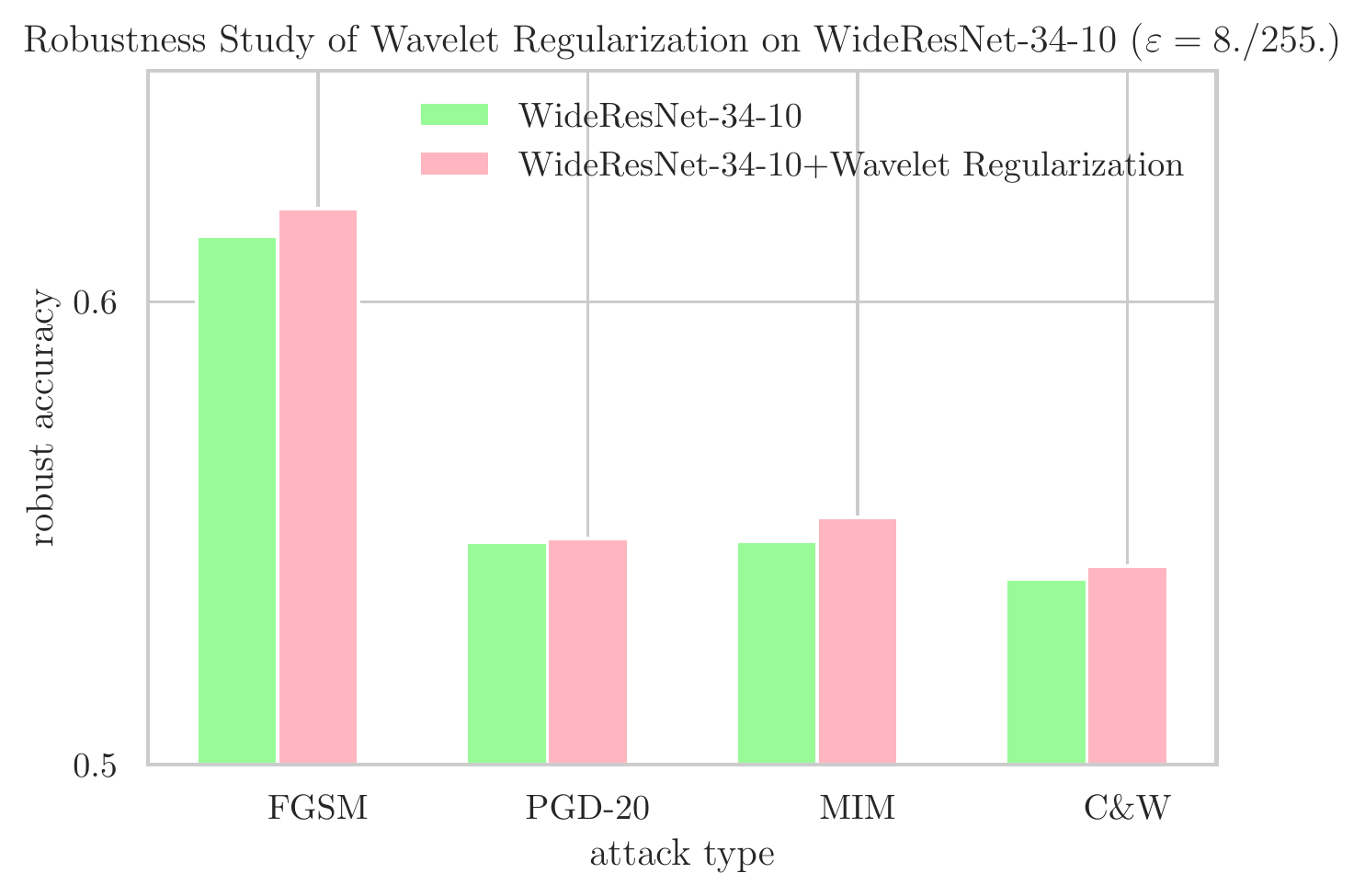}
%\caption{fig2}
}%
\caption{Robustness study of wavelet regularization module ($\varepsilon=0.031$).
Fig. (a) illustrates the effectiveness with/without wavelet regularization module on ResNet~\cite{ResNet}. 
Fig. (b) illustrates the effectiveness with/without wavelet regularization module on WideResNet-34-10~\cite{WideResNet}.}
\label{fig:ablationStudy}
\end{figure}
\textcolor{black}{To further study the applicability of our method, the ablation study is implemented. The intuitive result is shown in Fig. \ref{fig:ablationStudy}. Adding the wavelet regularization module to ResNet would cause the accuracy change with the value of +0.48\%, -0.76\%, +0.2\%, and -0.07\% under the FGSM, PGD, MIM, and C\&W attacks. As a comparison, the combination of WideResNet and wavelet regularization in the adversarial-training settings gains the increase of robust accuracy with the value of +0.6\%, +0.08\%, +0.51\%, and +0.28\% under the FGSM, PGD, MIM, and C\&W attacks. It can be seen that our proposed method is more applicable for the wide and deep neural network model.}
\textcolor{black}{\subsection{Position Of Wavelet Regularization Module}}
 \begin{figure}[!t]
        \centering
\includegraphics[scale=0.45]{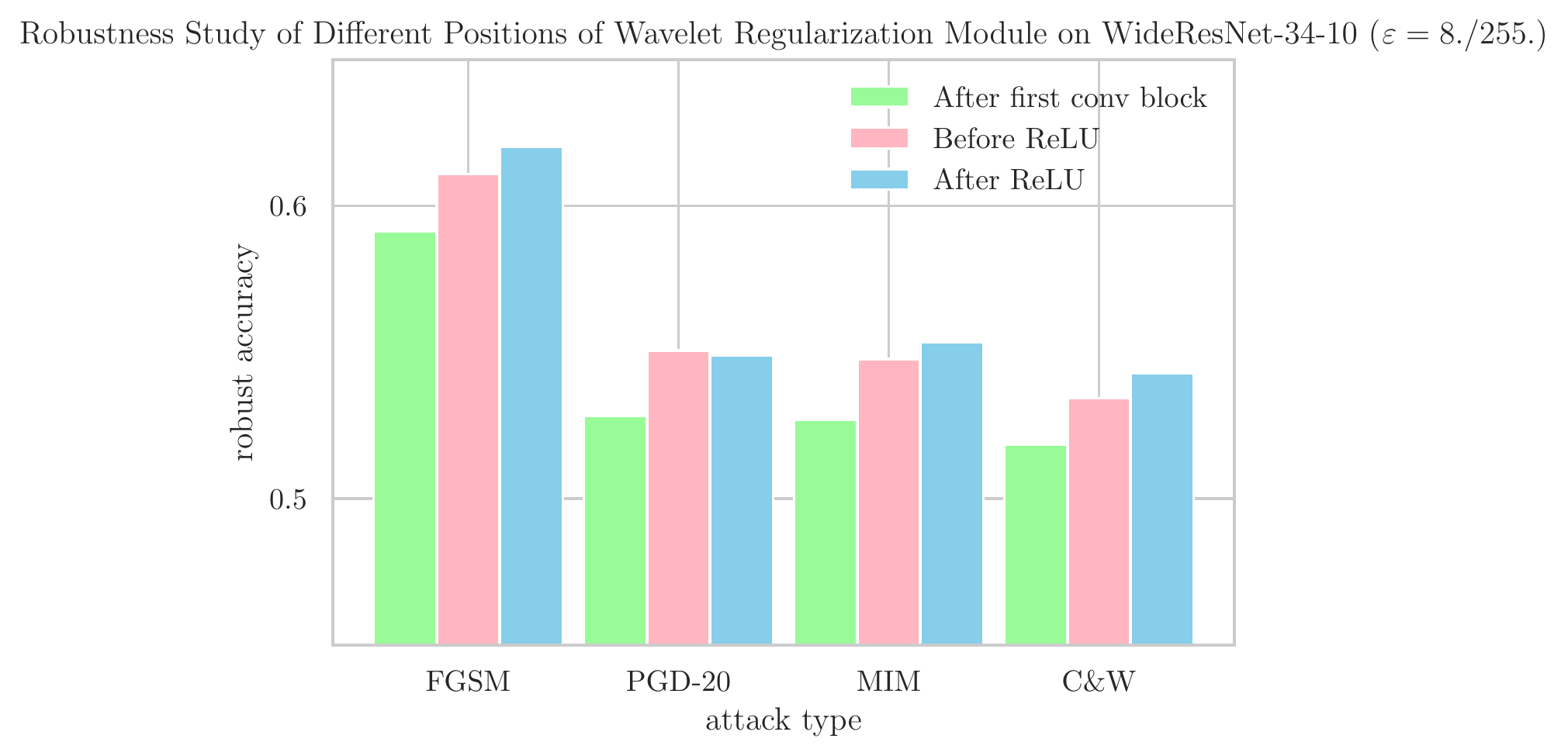}
\caption{The influence of robustness with different deployed positions of wavelet regularization on WideResNet-34-10~\cite{WideResNet} ($\varepsilon=0.031$). The experimental result of the deployment after first $3\times3$ conv block is marked in green color. The experimental result of the deployment before ReLU activation function is marked in pink color. The experimental result of the deployment after ReLU activation function is marked in blue color.}
\label{fig:WaveletRegularizationPosition}
\end{figure}
\textcolor{black}{What is the effect if the wavelet regularization model is inserted in different positions of the neural network model? To answer such a question, an experiment is implemented. The wavelet regularization module would be inserted after first $3\times3$ convolutional block (conv block), before ReLU activation function, after ReLU activation function (default setting). Fig. \ref{fig:WaveletRegularizationPosition} illustrates the quantitative result. The result is: \\FGSM Acc: $62.02\%\ (blue) \textgreater 61.09\%\ (pink) \textgreater 59.13\% (green)$.  \\PGD Acc: $55.06\%\ (pink) \textgreater 54.89\%\ (blue) \textgreater 52.83\% (green)$. \\MIM Acc: $55.34\%\ (blue) \textgreater 54.78\%\ (pink) \textgreater 52.71\% (green)$. \\C\&W Acc: $54.29\%\ (blue) \textgreater 53.44\%\ (pink) \textgreater 51.86\% (green)$.}

\textcolor{black}{In a deep neural network, the shallow convolutional block would focus on the information of edge and texture, while the deep convolutional block would focus on the detailed information of the object part. The insertion of wavelet regularization after the deep convolutional block could realize a positive catalytic effect. The ReLU activation function adds nonlinearity to the deep neural network model. The empirical result shows a tendency that adding the wavelet regularization after the ReLU function looks more feasible.}
\textcolor{black}{\subsection{Influence Of Model Width And Model Depth On Adversarial Robustness}}
\begin{figure*}[!t]
\centering
\subfigure[]{
\centering
\includegraphics[width=0.3\hsize]{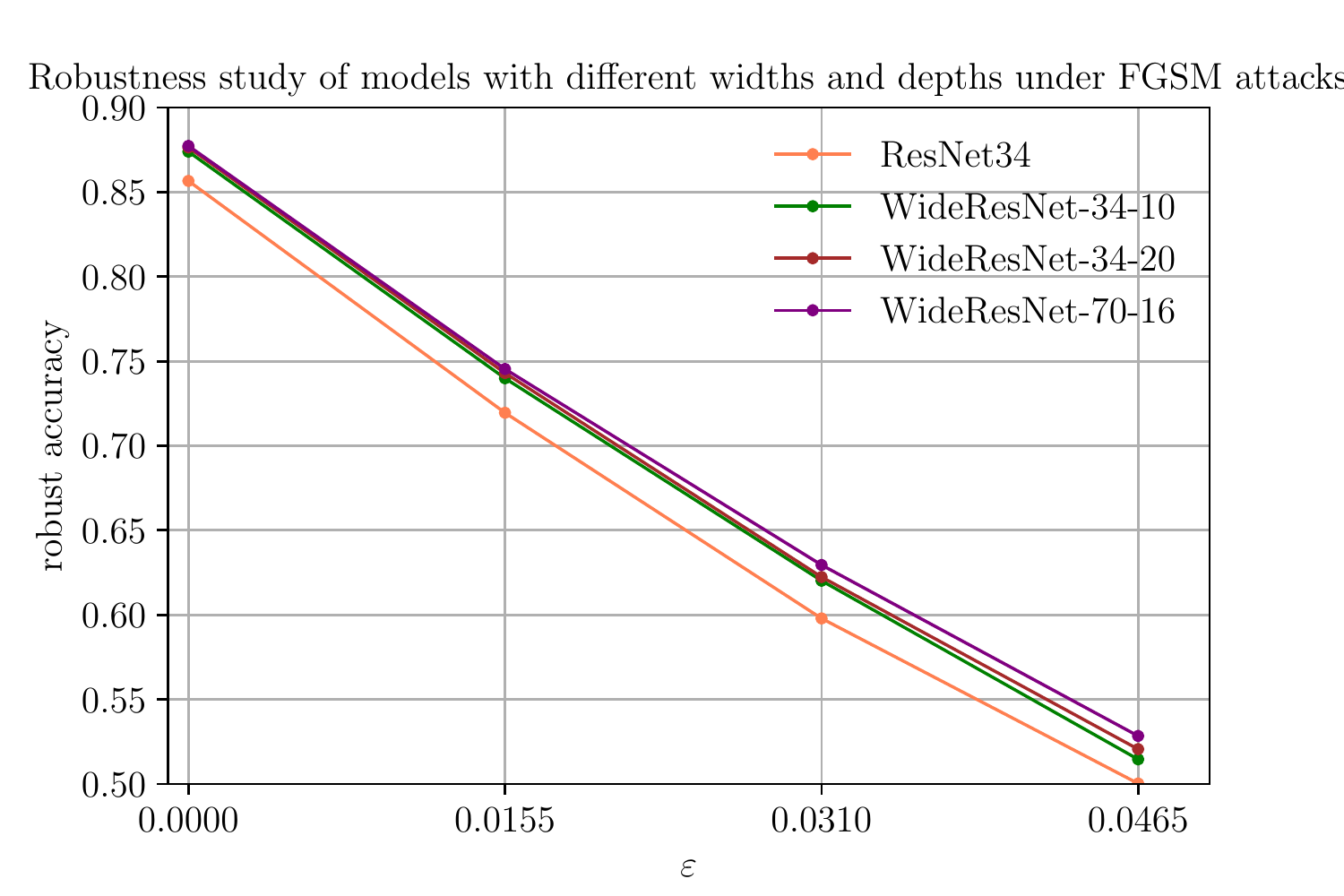}
%\caption{fig1}
}%
\subfigure[]{
\centering
\includegraphics[width=0.3\hsize]{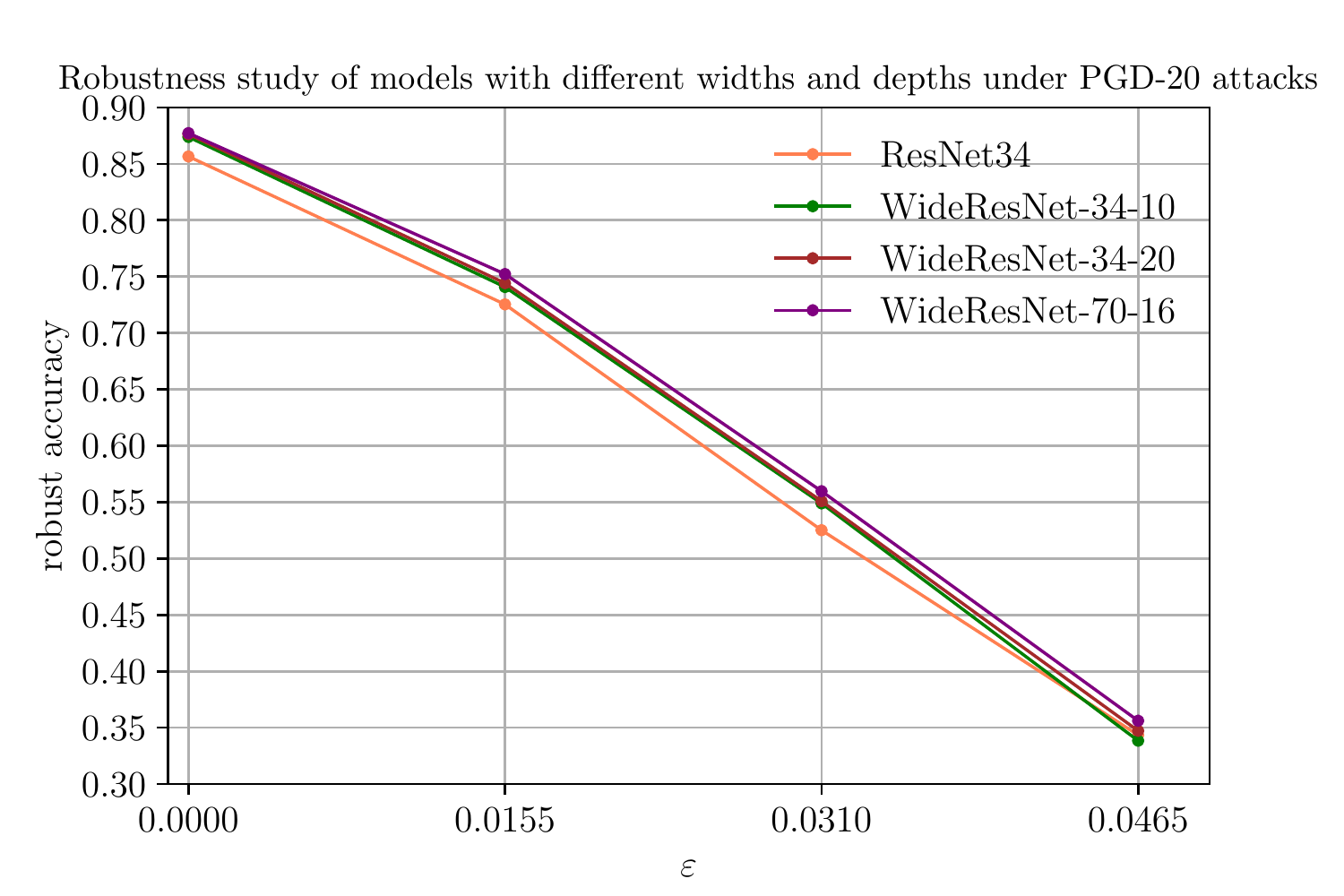}
%\caption{fig2}
}%
\subfigure[]{
\centering
\includegraphics[width=0.3\hsize]{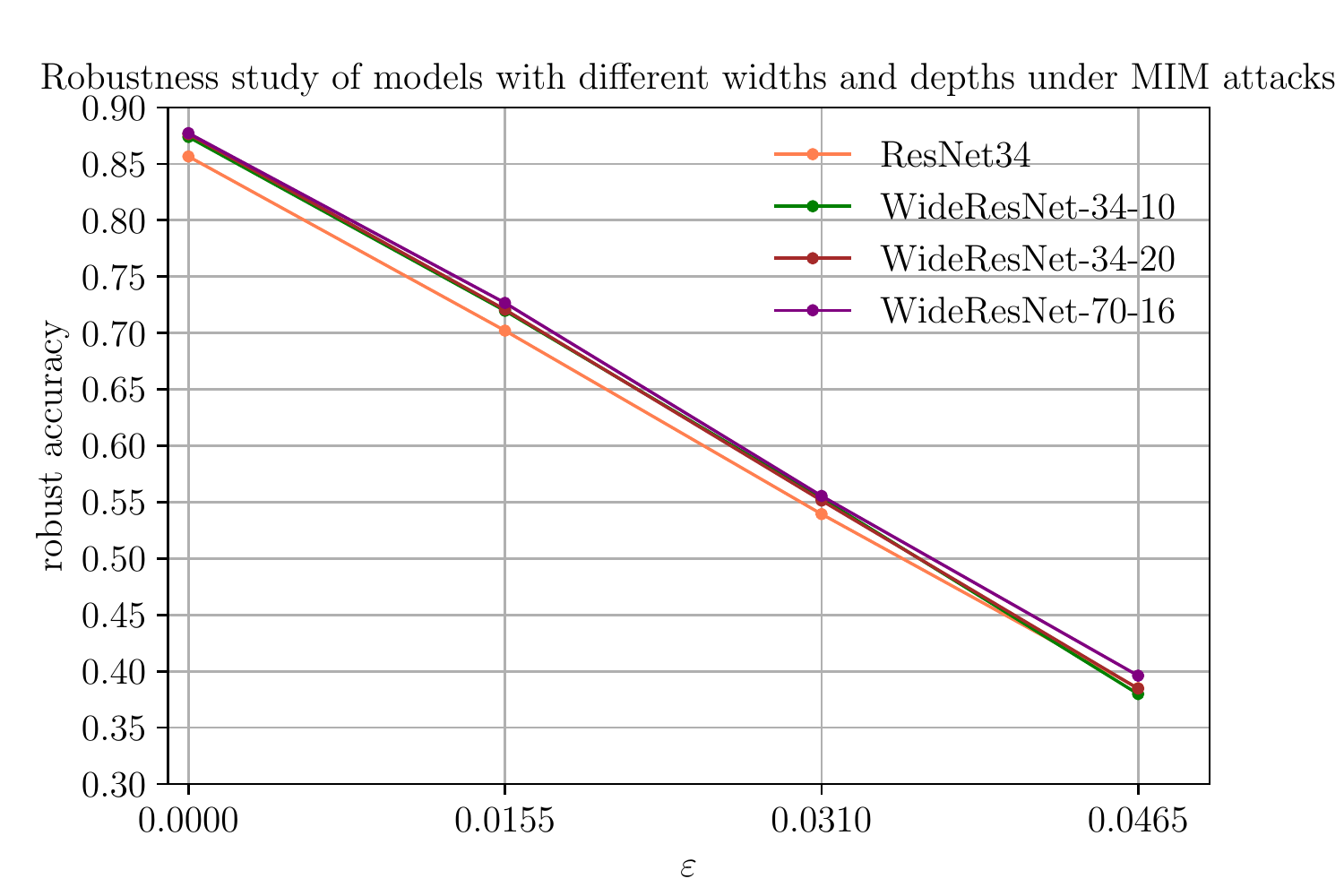}
%\caption{fig2}
}%
\\
\subfigure[]{
\centering
\includegraphics[width=0.3\hsize]{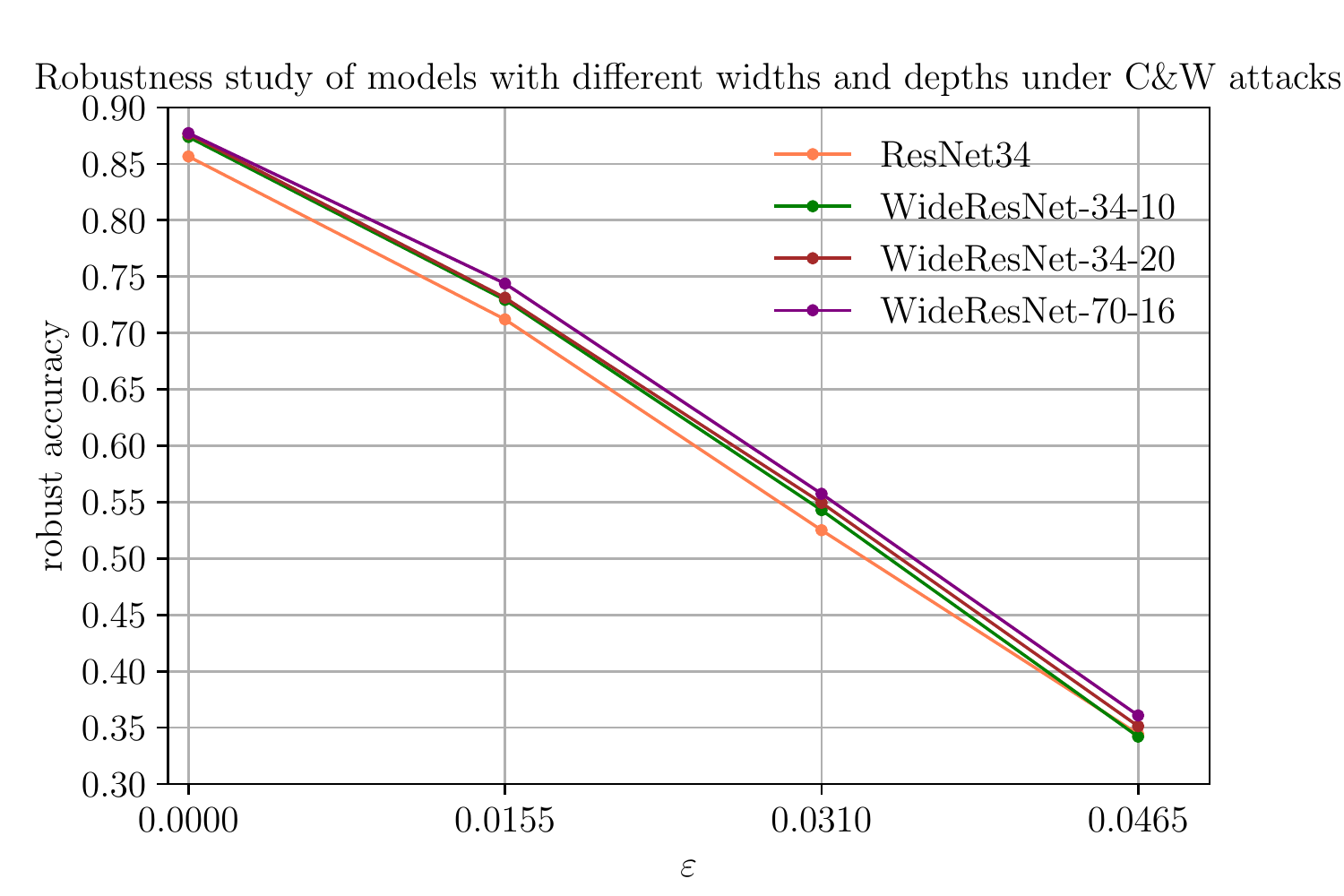}
%\caption{fig2}
}%
\subfigure[]{
\centering
\includegraphics[width=0.3\hsize]{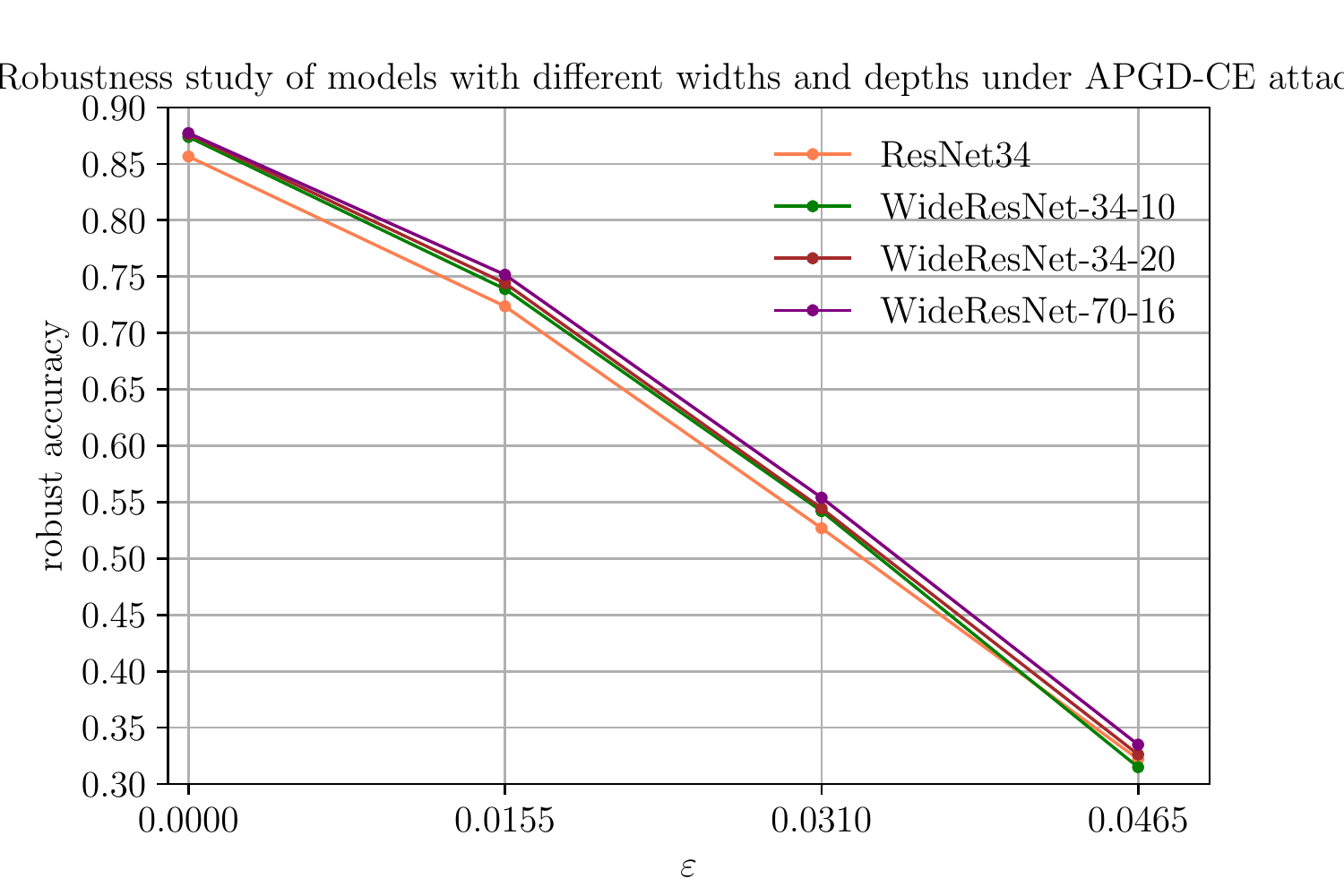}
%\caption{fig2}
}%
\subfigure[]{
\centering
\includegraphics[width=0.3\hsize]{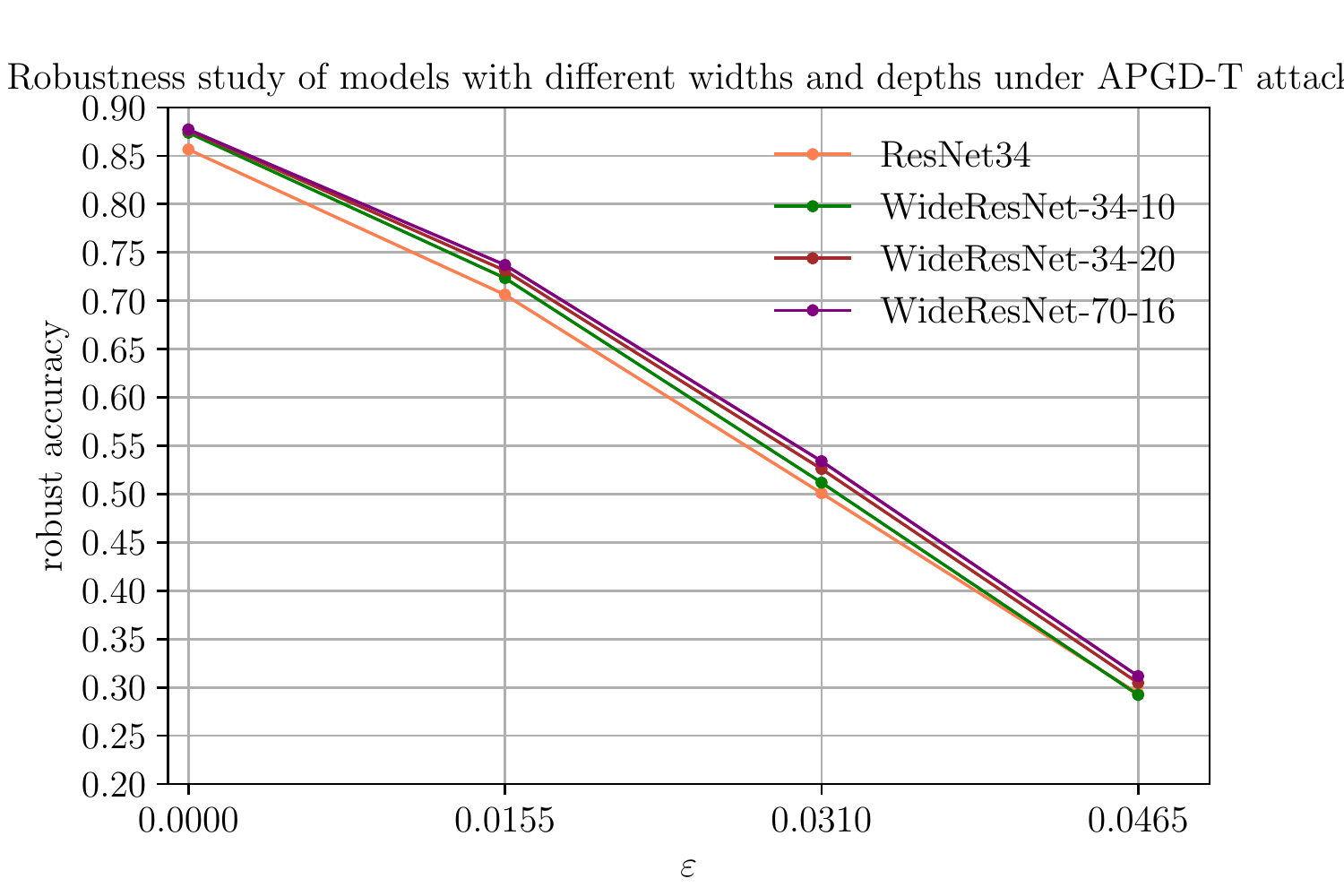}
%\caption{fig2}
}%
\caption{Robustness study of model in different widths and depths with our proposed method. It includes ResNet34 (85M parameters), WideResNet-34-10 (193M parameters), WideResNet-34-20 (772M parameters), and WideResNet-70-16 (1067M parameters). The perturbation budget range is $[0,\ 0.0155,\ 0.031,\ 0.0465]$.
Fig. (a) illustrates the robustness of different models under the FGSM attack~\cite{FGSM}. 
Fig. (b) illustrates the robustness of different models under the PGD attack~\cite{PGD}.
Fig. (c) illustrates the robustness of different models under the MIM attack~\cite{MIM}.
Fig. (d) illustrates the robustness of different models under the C\&W attack~\cite{C&W}.
Fig. (e) illustrates the robustness of different models under the APGD-CE attack~\cite{AutoAttack}.
Fig. (f) illustrates the robustness of different models under the APGD-T attack~\cite{AutoAttack}.}
\label{fig:widthDepth}
\end{figure*}
\textcolor{black}{In this study, there exists a postulation that the deep wide neural network would boost adversarial robustness. To validate such an assumption of overparameterization, an empirical study is made to explore the influence of model width and model depth on adversarial robustness. All listed four models (ResNet34, WideResNet-34-10, WideResNet-34-20, and WideResNet-70-16) are combined with the module of wavelet regularization.}

\textcolor{black}{Fig.\ref{fig:widthDepth} illustrates the quantitative result of the experiment for exploration of model-width and model-depth on adversarial robustness. In the overall dimension, WideResNet-70-16 is most robust under adversarial attacks. It is followed by WideResNet-34-20. WideResNet-34-10 surpasses ResNet34 except some metric under the large perturbation ($\varepsilon=0.0465$). To sum up, the combination of our regularization method and overparameterized model is beneficial to adversarial robustness.}
\subsection{Robustness Evaluation On The Frequency Perspective}
\begin{figure}[!t]
\centering
\subfigure[]{
\includegraphics[width=0.27\hsize]{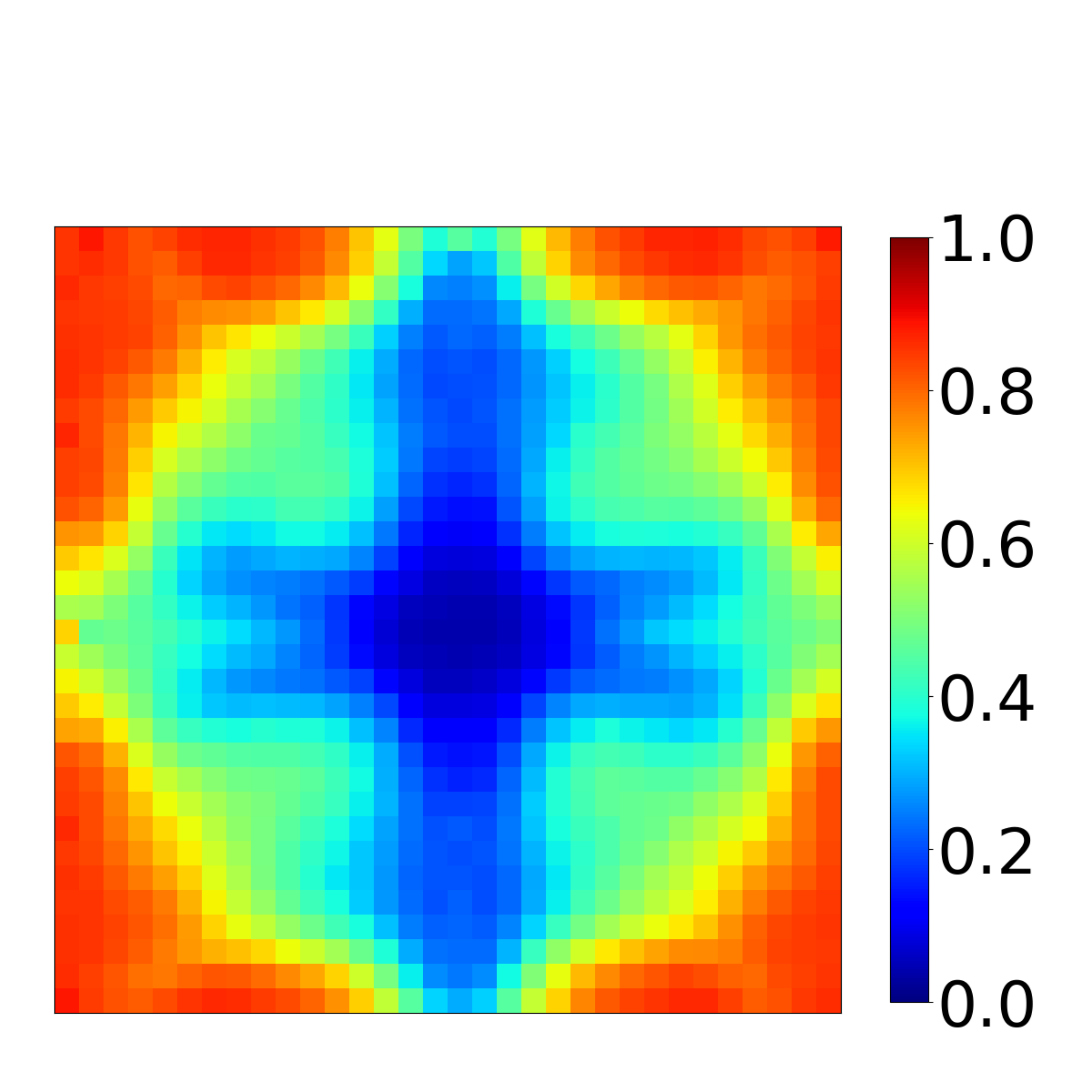}
%\caption{fig1}
}%
\subfigure[]{
\includegraphics[width=0.27\hsize]{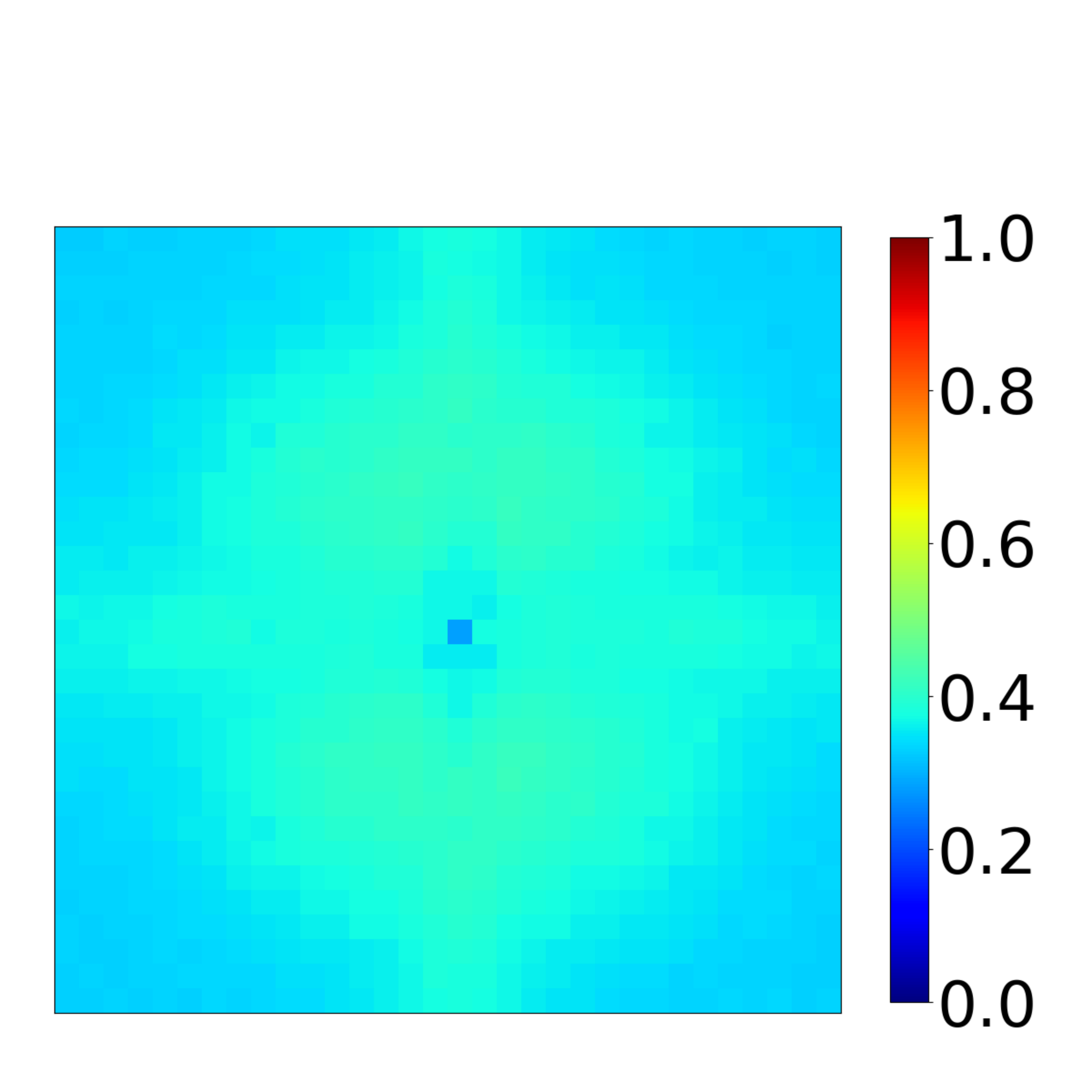}
%\caption{fig2}
}%
\quad 
\subfigure[]{
\includegraphics[width=0.27\hsize]{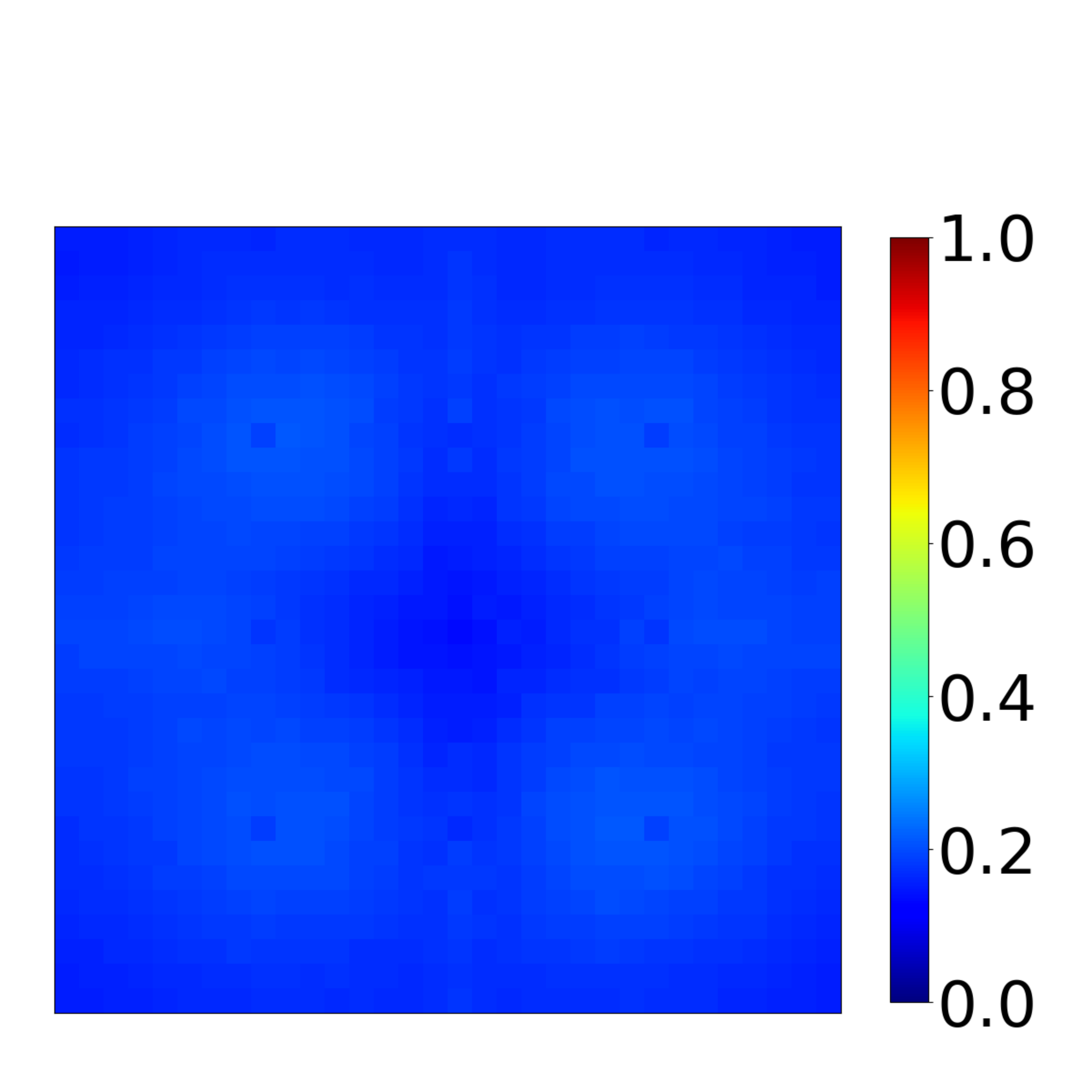}
}%
\caption{Fourier heat map of test error on different defense models ($\varepsilon=0.031$).
Fig. (a) illustrates the heat map of the vanilla WideResNet model~\cite{WideResNet}.
Fig. (b) illustrates the heat map of the adversarial training trick method~\cite{BagTricksAT}.
Fig. (c) illustrates the heat map of our WideWaveletResNet Model.
}
\label{fig:fourierHeatMap}
\end{figure}

In the previous work of the F-Principle of deep learning, Dong et al.~\cite{FourierRobustness} argued that discarding the high-frequency elements absolutely would have a negative impact on the robustness. Adversarial training just biases the perturbations towards the low-frequency domain. The Fourier heat map is proposed in this work~\cite{FourierRobustness} to evaluate the robustness in the frequency domain. In our experiment, the frequency perturbations are added to the CIFAR-10 validation dataset. The robustness of models on the Fourier perspective in different training settings is evaluated intuitively. The Fourier heat map can give a visualization of the robustness of different models in the frequency domain. The model sensitivity to perturbations aligns with different Fourier basis vectors on the CIFAR-10 validation dataset. As Fig. \ref{fig:fourierHeatMap} shows, the WideResNet model~\cite{WideResNet} of natural training is the most fragile which shows higher concentrations in the high-frequency domain and has the largest error rate. In contrast, the model of adversarial training trick method~\cite{BagTricksAT} performs better than the natural-training model. However, our proposed method performs best on the metrics of test error. Although the model trained by the adversarial training trick method~\cite{BagTricksAT} lifts the robustness under the high-frequency perturbations, it sacrifices clean accuracy and robustness under the low-frequency perturbations. Our proposed model has a considerable robustness performance under both low-frequency perturbations and high-frequency perturbations. 
\subsection{Comparison Experiment On Various Wavelet Bases}
Besides the Haar wavelet base, we also test the wavelet regularizations with other base functions: Daubechies-5 wavelet base (Db5), Biorthogonal-3.1 wavelet base (bior3.1), Coiflets 4 wavelet base (Coif4), Discrete Meyer wavelet base (Dmey), Reverse Biorthogonal-2.2 wavelet base (Rbio2.2), and Symlets 4 wavelet base (Sym4). The analysis of these wavelet base functions can be seen in the mathematical books~\cite{DaubechiesBook, MeyerBook, MallatBook}. The attack setting is the same as the main experiment with the perturbation budget $\varepsilon=0.031$ and step sizes $20$ for PGD attack, MIM attack, and C\&W attack. Also the auto attacks including the APGD-CE attack and the APGD-T attack are tested.
\begin{table*}[!t]
\renewcommand{\arraystretch}{1.5}
% if using array.sty, it might be a good idea to tweak the value of
% \extrarowheight as needed to properly center the text within the cells
\caption{THE EVALUATION METRICS (\%) OF WAVELET REGULARIZATION ON DIFFERENT BASE FUNCTIONS UNDER WHITE-BOX ATTACKS AND AUTO-ATTACKS ON CIFAR-10 ($\varepsilon=0.031$).}
\label{table:waveletBase}
\centering
\footnotesize
% Some packages, such as MDW tools, offer better commands for making tables
% than the plain LaTeX2e tabular which is used here.
\begin{tabular}{lcccccccr@{.}}
\toprule
     &   Clean Acc  & FGSM Acc&  PGD Acc&  MIM Acc&  C\&W Acc& APGD-CE Acc& APGD-T Acc\\  % inserts table
 \midrule
Db5 & 86.72\thinspace$\pm$\thinspace 0.103 & 61.19\thinspace$\pm$\thinspace 0.155 & 54.87\thinspace$\pm$\thinspace 0.234 & 54.74\thinspace$\pm$\thinspace 0.217 & 54.05\thinspace$\pm$\thinspace 0.186 & 54.26\thinspace$\pm$\thinspace 0.204 & 51.79\thinspace$\pm$\thinspace 0.217\\
Bior3.1 & 86.56\thinspace$\pm$\thinspace 0.206 & 61.14\thinspace$\pm$\thinspace 0.138 & 54.93\thinspace$\pm$\thinspace 0.096 & 54.66\thinspace$\pm$\thinspace 0.161 & 54.03\thinspace$\pm$\thinspace 0.143& 54.24\thinspace$\pm$\thinspace 0.107 & 51.61\thinspace$\pm$\thinspace 0.237\\
Coif4 & 86.72\thinspace$\pm$\thinspace 0.482 & 61.36\thinspace$\pm$\thinspace 0.141 & 55.25\thinspace$\pm$\thinspace 0.249 & 54.83\thinspace$\pm$\thinspace 0.238 & 54.50\thinspace$\pm$\thinspace 0.083& 54.64\thinspace$\pm$\thinspace 0.195 & 52.23\thinspace$\pm$\thinspace 0.131\\
Dmey & 86.75\thinspace$\pm$\thinspace 0.293 & 61.14\thinspace$\pm$\thinspace 0.250 & 54.82\thinspace$\pm$\thinspace 0.141 & 54.49\thinspace$\pm$\thinspace 0.081 & 54.10\thinspace$\pm$\thinspace 0.221 & 54.17\thinspace$\pm$\thinspace 0.146 & 51.64\thinspace$\pm$\thinspace 0.345\\
Rbio2.2 &  87.07\thinspace$\pm$\thinspace 0.319 & 61.51\thinspace$\pm$\thinspace 0.396 & 54.70\thinspace$\pm$\thinspace 0.331 & 54.53\thinspace$\pm$\thinspace 0.267 & 53.88\thinspace$\pm$\thinspace 0.154 & 54.00\thinspace$\pm$\thinspace 0.301 & 51.44\thinspace$\pm$\thinspace 0.148\\
Sym4 & 86.59\thinspace$\pm$\thinspace 0.111 & 61.18\thinspace$\pm$\thinspace 0.214 & 55.26\thinspace$\pm$\thinspace 0.209 & 54.90\thinspace$\pm$\thinspace 0.137 & 54.25\thinspace$\pm$\thinspace 0.086 & 54.57\thinspace$\pm$\thinspace 0.228 & 52.03\thinspace$\pm$\thinspace 0.164\\
Haar &  87.34\thinspace$\pm$\thinspace 0.085 & 61.61\thinspace$\pm$\thinspace 0.361 & 54.76\thinspace$\pm$\thinspace 0.116 & 54.47\thinspace$\pm$\thinspace 0.602 & 54.25\thinspace$\pm$\thinspace 0.074 & 54.10\thinspace$\pm$\thinspace 0.138 & 51.77\thinspace$\pm$\thinspace 0.131\\
         \bottomrule
\end{tabular}
\end{table*}
\begin{table*}[htbp]
\renewcommand{\arraystretch}{1.5}
% if using array.sty, it might be a good idea to tweak the value of
% \extrarowheight as needed to properly center the text within the cells
\caption{THE EVALUATION METRICS (\%) OF WAVELET REGULARIZATION ON DIFFERENT BASE FUNCTIONS UNDER WHITE-BOX ATTACKS AND AUTO-ATTACKS ON CIFAR-100 ($\varepsilon=0.031$).}
\label{table:waveletBaseCIFAR100}
\centering
\footnotesize
% Some packages, such as MDW tools, offer better commands for making tables
% than the plain LaTeX2e tabular which is used here.
\begin{tabular}{lcccccccr@{.}}
\toprule
     &   Clean Acc& FGSM Acc&  PGD Acc  &  MIM Acc&  C\&W Acc & APGD-CE Acc & APGD-T Acc \\  % inserts table
 \midrule
Db5 & 61.08\thinspace$\pm$\thinspace 0.801 & 
34.80\thinspace$\pm$\thinspace 0.339 & 
32.01\thinspace$\pm$\thinspace 0.261 & 
30.44\thinspace$\pm$\thinspace 0.298 &
30.73\thinspace$\pm$\thinspace 0.196 & 
31.27\thinspace$\pm$\thinspace 0.321 & 
28.21\thinspace$\pm$\thinspace 0.170\\
Bior3.1 & 60.88\thinspace$\pm$\thinspace 0.133 & 
34.68\thinspace$\pm$\thinspace 0.059 & 
32.14\thinspace$\pm$\thinspace 0.189 & 
30.46\thinspace$\pm$\thinspace 0.112 & 
30.55\thinspace$\pm$\thinspace 0.244 & 
31.52\thinspace$\pm$\thinspace 0.166 &
28.19\thinspace$\pm$\thinspace 0.136\\
Coif4 & 60.33\thinspace$\pm$\thinspace 0.200& 34.48\thinspace$\pm$\thinspace 0.203 & 
32.03\thinspace$\pm$\thinspace 0.307 & 
30.53\thinspace$\pm$\thinspace 0.164 & 
30.37\thinspace$\pm$\thinspace 0.316 & 
31.88\thinspace$\pm$\thinspace 0.270 & 
27.89\thinspace$\pm$\thinspace 0.307\\
Dmey & 61.20\thinspace$\pm$\thinspace 0.388 & 
34.60\thinspace$\pm$\thinspace 0.243& 
32.02\thinspace$\pm$\thinspace 0.267 & 30.50\thinspace$\pm$\thinspace 0.304& 30.61\thinspace$\pm$\thinspace 0.126 & 
31.40\thinspace$\pm$\thinspace 0.314 & 
28.23\thinspace$\pm$\thinspace 0.128\\
Rbio2.2 &  61.42\thinspace$\pm$\thinspace 0.480 &
 34.69\thinspace$\pm$\thinspace 0.353 &  32.00\thinspace$\pm$\thinspace 0.208 & 
 30.50\thinspace$\pm$\thinspace 0.126 & 
 30.49\thinspace$\pm$\thinspace 0.296 & 
 31.43\thinspace$\pm$\thinspace 0.237 & 
 28.08\thinspace$\pm$\thinspace 0.252\\
Sym4 & 61.35\thinspace$\pm$\thinspace 0.374 & 34.49\thinspace$\pm$\thinspace 0.381 & 31.95\thinspace$\pm$\thinspace 0.135 & 30.41\thinspace$\pm$\thinspace 0.218 & 30.37\thinspace$\pm$\thinspace 0.310 & 31.21\thinspace$\pm$\thinspace 0.315 & 
27.93\thinspace$\pm$\thinspace 0.304\\
Haar &  61.47\thinspace$\pm$\thinspace 0.416 & 34.80\thinspace$\pm$\thinspace 0.394 & 32.19\thinspace$\pm$\thinspace 0.166 & 30.56\thinspace$\pm$\thinspace 0.135 & 30.47\thinspace$\pm$\thinspace 0.142 &
31.56\thinspace$\pm$\thinspace 0.176 &
28.11\thinspace$\pm$\thinspace 0.133\\
         \bottomrule
\end{tabular}
\end{table*}

As Table \ref{table:waveletBase} and Table \ref{table:waveletBaseCIFAR100} exhibit, the Haar base function is the most suitable. On CIFAR-10, the Sym4 base performs best under the PGD attack and MIM attack, and the Coif4 base performs best under the adaptive C\&W attack and auto attacks. Nevertheless, the Haar base performs best without adversarial perturbation and under the FGSM attack. On CIFAR-100, the Haar base performs best on several metrics (clean Acc, FGSM Acc, PGD Acc, MIM Acc). Summarized from the numerical results on the two datasets, the utilization of the Haar function would be more feasible at the comprehensive level. As we recall from Table \ref{table:waveletBaseProperty}, the Haar function is orthogonal, symmetrical, and compactly supported. It has a short support length, a short filter size, and a short vanishing movement (written as ``Movement" in the table). The goal of wavelet decomposition is to realize the sparsity and reduce the structural risk for robustness improvement. The empirical result validates that the Haar function with a short vanishing movement adapts to such a scenario. 
\begin{table*}[htbp]
\renewcommand{\arraystretch}{1.5}
% if using array.sty, it might be a good idea to tweak the value of
% \extrarowheight as needed to properly center the text within the cells
\caption{THE PROPERTY COMPARISON OF DIFFERENT WAVELET BASES.}
\label{table:waveletBaseProperty}
\centering
\footnotesize
% Some packages, such as MDW tools, offer better commands for making tables
% than the plain LaTeX2e tabular which is used here.
\begin{tabular}{lcccccccr@{.}}
\toprule
     &   Orthogonality  & Double orthogonality &  Compactness  &  Support size &  Filter size & Symmetry & Movement \\  % inserts table
 \midrule
Haar & Yes & Yes & Yes & 1 & 2 & Yes & 1\\
Db5 & Yes & Yes & Yes & 9 & 10 & Approximately yes & 5\\
Bior3.1 & No & Yes & Yes & 3 & 8 & No & 2\\
Coif4 & Yes & Yes & Yes & 23 & 24 & Approximately yes & 8\\
Dmey & No & No & No & N/A & N/A & Yes & N/A\\
Rbio2.2 & No & Yes & Yes & N/A & N/A & Yes & N/A\\
Sym4 & Yes & Yes & Yes & 7 & 8 & Approximately yes & 4\\
         \bottomrule
\end{tabular}
\end{table*}
\subsection{Robustness Evaluation On Two Aggregation Strategies}
\begin{table*}[!t]
% if using array.sty, it might be a good idea to tweak the value of
% \extrarowheight as needed to properly center the text within the cells
\caption{THE PERFORMANCE (\%) OF DIFFERENT WAVELET AGGREGATION STRATEGIES ON CIFAR-10 AND CIFAR-100 ($\varepsilon=0.031$).}
\label{table:waveletAggregation}
\centering
\footnotesize
% Some packages, such as MDW tools, offer better commands for making tables
% than the plain LaTeX2e tabular which is used here.
\begin{tabular}{lccccccccr@{.}}
\hline
     &Method &   Clean Acc & FGSM Acc&  PGD Acc &  MIM Acc&  C\&W Acc& APGD-CE Acc& APGD-T Acc\\  % inserts table
 \hline
\multirow{2}*{CIFAR-10} &LPF & 86.94\thinspace$\pm$\thinspace 0.151 & 61.11\thinspace$\pm$\thinspace 0.201 & 54.94\thinspace$\pm$\thinspace 0.121 & 54.57\thinspace$\pm$\thinspace 0.110 & 53.55\thinspace$\pm$\thinspace 0.177 & 54.14\thinspace$\pm$\thinspace 0.111 & 51.28\thinspace$\pm$\thinspace 0.115\\
&WAP &   87.34\thinspace$\pm$\thinspace 0.085 & 61.61\thinspace$\pm$\thinspace 0.361 & 54.76\thinspace$\pm$\thinspace 0.116 & 54.47\thinspace$\pm$\thinspace 0.602 & 54.25\thinspace$\pm$\thinspace 0.074 & 54.10\thinspace$\pm$\thinspace 0.138 & 51.77\thinspace$\pm$\thinspace 0.131 \\
         \hline
\multirow{2}*{CIFAR-100} &LPF & 61.66\thinspace$\pm$\thinspace 0.489 & 34.36\thinspace$\pm$\thinspace 0.721 & 31.31\thinspace$\pm$\thinspace 0.948 & 30.03\thinspace$\pm$\thinspace 0.794 & 29.77\thinspace$\pm$\thinspace 0.675 &30.69\thinspace$\pm$\thinspace 0.942 & 27.48\thinspace$\pm$\thinspace 0.720\\
&WAP & 61.47\thinspace$\pm$\thinspace 0.416 & 34.80\thinspace$\pm$\thinspace 0.394 & 32.19\thinspace$\pm$\thinspace 0.166 & 30.56\thinspace$\pm$\thinspace 0.135 & 30.47\thinspace$\pm$\thinspace 0.142 &
31.56\thinspace$\pm$\thinspace 0.176 &
28.11\thinspace$\pm$\thinspace 0.133\\
         \hline
\end{tabular}
\end{table*}
We compare our proposed method of Wavelet Average Pooling (WAP) with the strategy of low-pass filter (LPF) which means that only the low-frequency elements with the approximation would be retained. As Table \ref{table:waveletAggregation} illustrates, the aggregation strategy of Wavelet Average Pooling realizes a better trade-off of clean accuracy and adversarial robustness. On CIFAR-10, the method of WAP surpasses the method of LPF on the metrics of clean Acc, FGSM Acc, adaptive C\&W Acc, and APGD-T Acc. Nevertheless, the method of WAP is inferior to the method of LPF on the metrics of PGD Acc, MIM Acc, and APGD-CE Acc. On CIFAR-100, the method of WAP surpass the method of LPF under all types of adversarial perturbations. On CIFAR-100, the clean Acc of LPF method is better than the clean Acc of WAP method. The reason can be attributed to the F-Principle, the low-frequency elements are related to the generalization ability while the high-frequency elements have tight connections to the robustness~\cite{FourierRobustness}. Filtering the high-frequency elements may obtain a higher clean Acc on the CIFAR-100 dataset. However, the trade-off between the low-frequency elements and the high-frequency elements is more applicable to realize adversarial robustness.

\subsection{Discussion}
The proposal of the wavelet regularization of adversarial training aims to improve the adversarial robustness of the neural networks in the frequency domain. We explore the mathematical property of wavelet transform and postulate that its regularization character can help realize a small Lipschitz constraint which is good for robustness. The training record of the experiment shows that our proposed method's gradient value is lower than $0.05$ which means that our wavelet regularization method enhances adversarial training with a low Lipschitz constraint. The empirical study of the defense against white-box attacks, black-box attacks, adaptive attacks, and auto attacks validates the effectiveness of our proposed method. The intuitive visualization experiment shows that our proposed method could realize interpretability in the process of feature extraction. Furthermore, the ablation study and the comparison experiment with the wavelet regularization method on various models such as ResNet34, WideResNet-34-10, WideResNet-34-20, and WideResNet-76-10 demonstrates that our proposed method could boost the adversarial robustness on the basis of overparameterization. The lifting is more intuitive in a wide and deep neural network. 
%On the metrics of clean accuracy, most natural training methods surpass our method and other adversarial training methods. However, the utilization of adversarial training can guarantee robustness to some extent. Among these adversarial training methods, the learning features of our WideWaveletResNet are most interpretable and robust as the visualization experiment demonstrates. In the future, it is meaningful to further study the relationship between generalization without the perturbations (the metrics of clean accuracy in the experiment) and robustness with the perturbations. A more robust model without the sacrifice of generalization ability would be anticipated. Using semi-supervised learning to reinforce the robustness has been studied by Huang et al.~\cite{SelfAdaptiveTraining}. It gives us an inspiration to further study this technology.

Our method is inspired by the F-Principle. In the research of deep learning, the F-Principle expresses an opinion that the low-frequency elements are related to the generalization of natural training while high-frequency elements are related to the robustness under the perturbations. We utilize the Fourier heat map~\cite{FourierRobustness} to validate our robustness in the frequency domain. By biasing the high frequency towards the low frequency, robustness can be guaranteed with a low error rate. In the sub-experiment of different aggregation strategies of wavelets, we find that discarding the high-frequency elements absolutely leads to the degradation of robustness compared to Wavelet Average Pooling. In summary, our adversarial wavelet training method can be regarded as a positive example of the F-Principle.

In the family of wavelet bases, the Haar base is orthogonal, doubly orthogonal, and compactly supported. It has a short filter size, a short support size, and a short vanish movement. This character leads to a robust sparsity property. The utilization of the Haar function cannot guarantee that it is always the most robust under all adversarial attacks. However, it is most suitable to choose the Haar function as a default setting of wavelet regularization at a comprehensive level.

\section{Conclusion}
Adversarial vulnerability can be regarded as a high-frequency phenomenon. Inspired by such a motivation, we propose a wavelet regularization method and integrate it into a wide residual neural network to form a WideWaveletResNet. During the process of adversarial training, our wavelet regularization method based on the Haar function could realize a small Lipschitz constraint for the guarantee of adversarial robustness. Our proposed model achieves considerable robust performance under the diverse types of attacks on the datasets of CIFAR-10 and CIFAR-100. The enhancement of adversarial robustness is intuitive when the wavelet regularization method works on a wide and deep neural network. The visualization experiment result also validates the interpretability of our model's learning features and the robustness of our model from a frequency perspective. A detailed comparison experiment is implemented to show the feasibility of the Haar base function. The absence of the evaluation of the proposed method on the high-resolution database is perhaps a limitation of current research work. In the future, we would extend our wavelet regularization method to high-resolution image datasets to make our methods more applicable. Moreover, the relationship between SRM in traditional statistical learning and overparameterization in deep learning deserves in-depth research. Besides the improvement of adversarial robustness from the perspective of model structure, the utilization of robust non-convex optimization methods and domain adaptation methods~\cite{GANMatchingDA, AdvResTransformUDA} is also feasible to boost adversarial robustness.

\section*{Acknowledgment}
This work was supported by the National Natural Science Foundation of China under Grant No. 61701348. The authors would like to thank TUEV SUED for the kind and generous support. The team of the Sino-German Center of Intelligent Systems in Tongji University gives us support to carry our research.
The authors are grateful to the Ph.D. candidate of Tongji University, Mr. Yujian Mo for his suggestion on our research. The authors would appreciate Dr. Qi Deng and Dr. Yunmei Shi for their suggestions on our manuscript. We sincerely thank Mr. Xi Fang and Mr. Huan Hua for their help in our experimental work.

\bibliographystyle{IEEEtran}%IEEEtran
\bibliography{wavelet_adv_training_ref.bib}
% or


%\documentclass[draftclsnofoot,onecolumn]{IEEEtran}
\begin{IEEEbiography}[{\includegraphics[width=1in,height=1.25in]{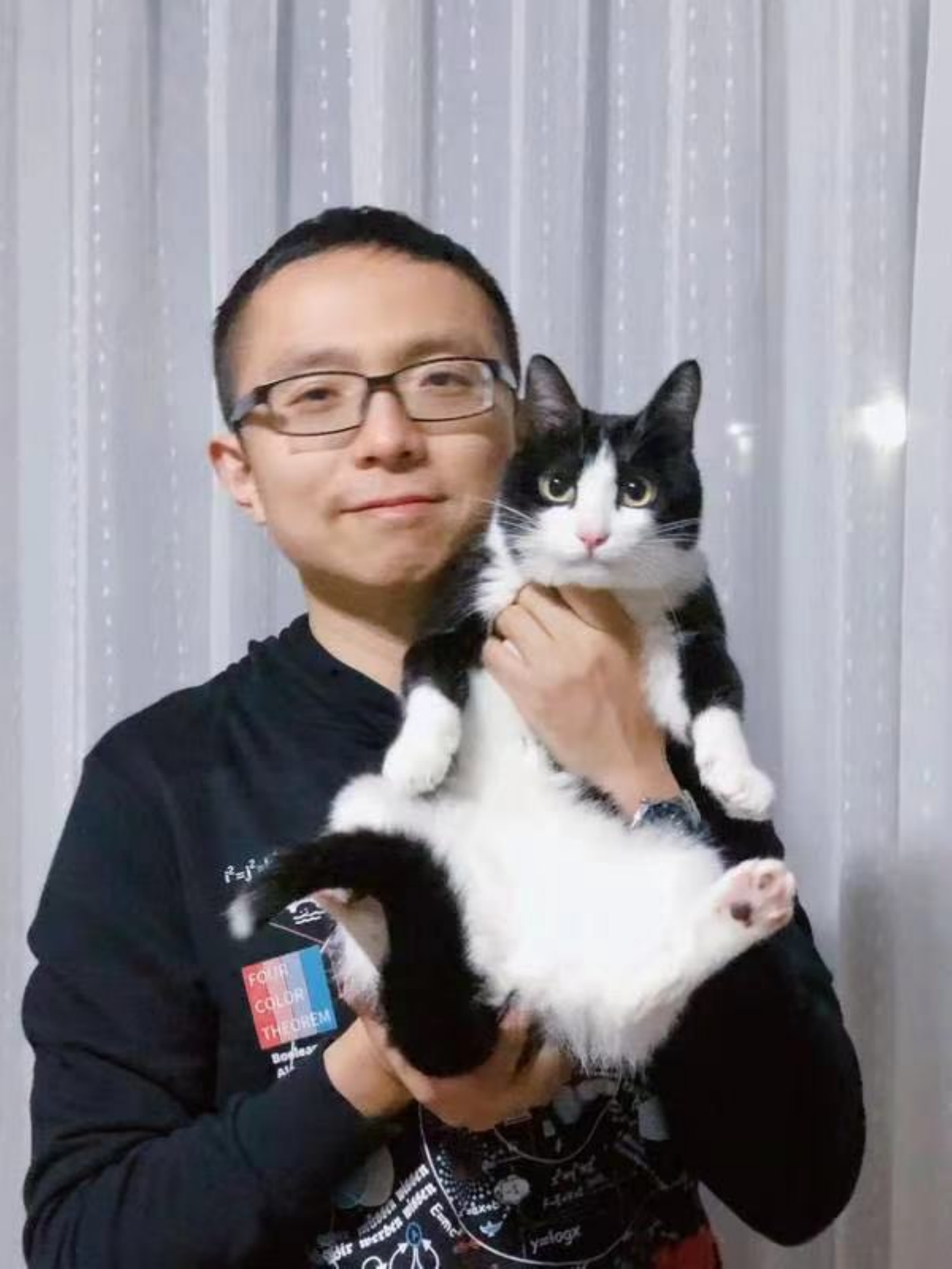}}]{Jun Yan} received a bachelor degree from the Department of Electronic Engineering, Shanghai Jiaotong University in 2009 and a master degree from the Chinese-German High College, Tongji University in 2012. After the eight-year working experience in the industrial field, he is now a Ph.D. candidate in the Department of Information and Communication Engineering, Tongji University, China.
His research interest is the theory of adversarial machine learning and the design of new deep learning models in autonomous vehicles.
\end{IEEEbiography}
\begin{IEEEbiography}[{\includegraphics[width=1in,height=1.25in]{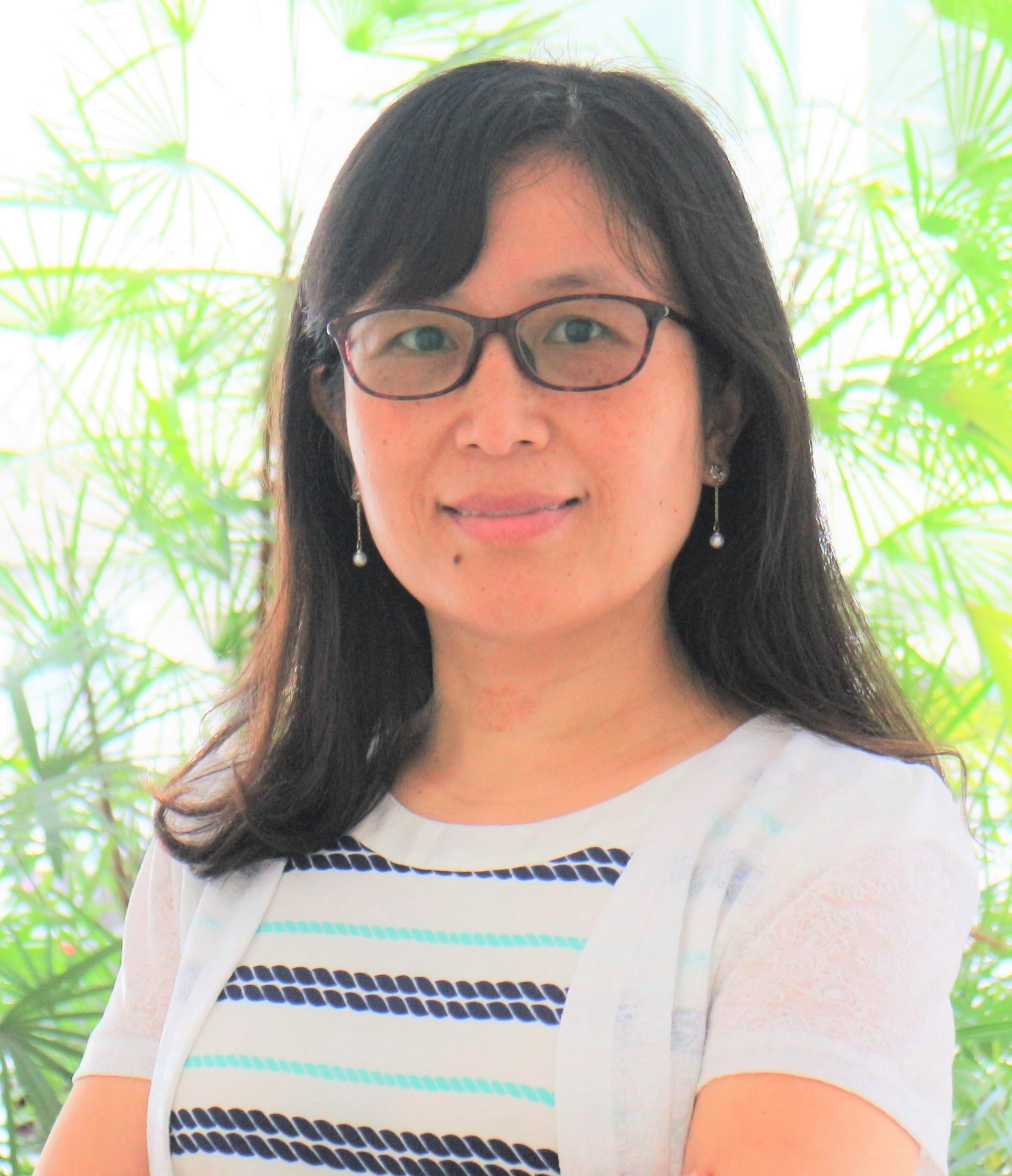}}]{Huilin Yin}
 received the Ph.D. degree in control theory and control engineering from Tongji University, China, in 2006. She is currently the chaired professor of TUEV SUED Chair with the Electronic and Information Engineering College, Tongji University. She received the M.S. double-degree from Tongji University and the Technical University of Munich, Germany. Her research interests include environment perception of intelligent vehicles and safety for autonomous driving.
\end{IEEEbiography}
\begin{IEEEbiography}[{\includegraphics[width=1in,height=1.25in]{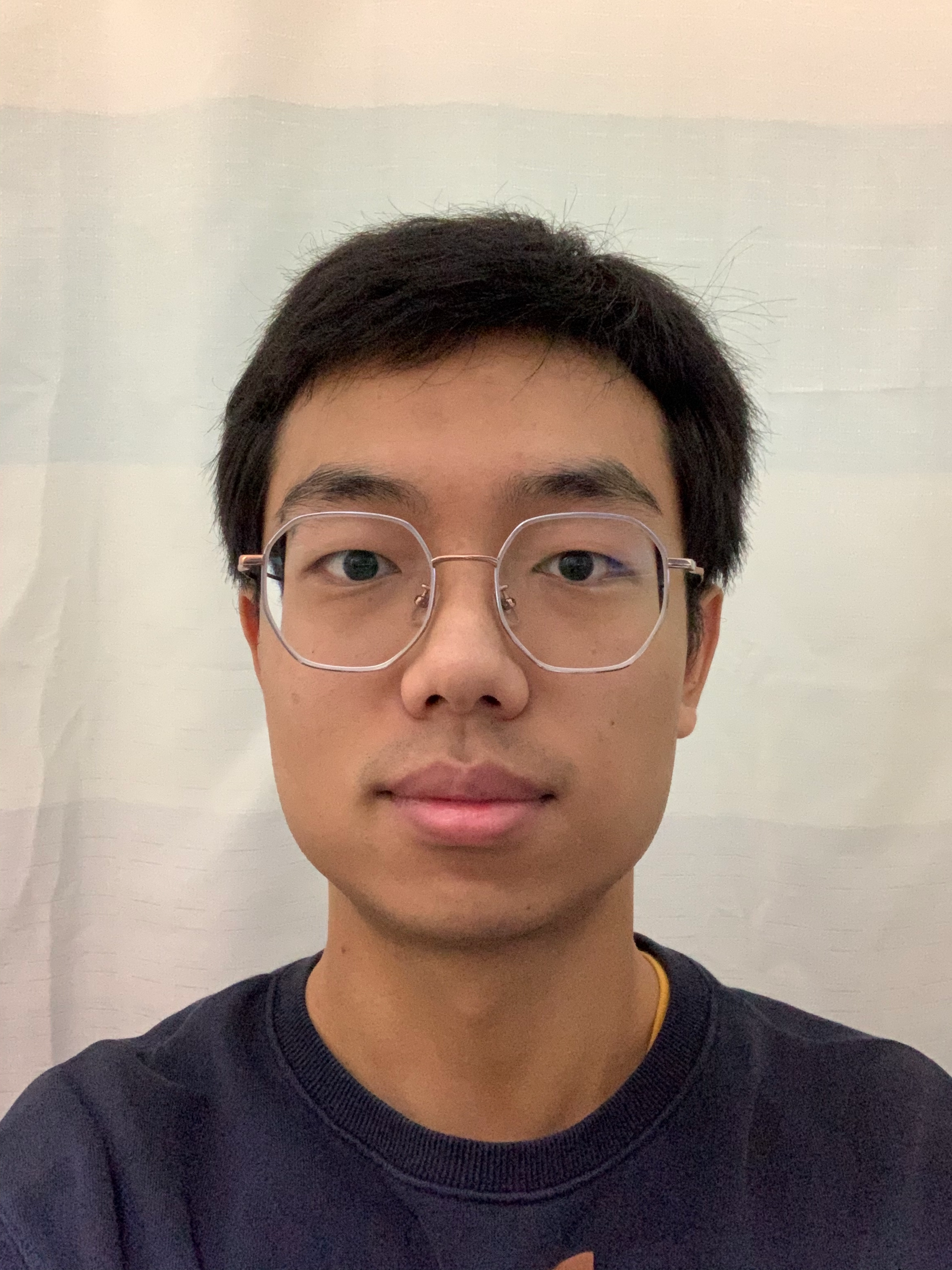}}]{Xiaoyang Deng}
Deng Xiaoyang got his bachelor's degree from the School of Electronics and Information Engineering of Tongji University, China. He is currently studying for a master's degree in the Department of Control Science and Engineering, Tongji University. He focuses on the theory of adversarial machine learning.
\end{IEEEbiography}
\begin{IEEEbiography}[{\includegraphics[width=1in,height=1.25in]{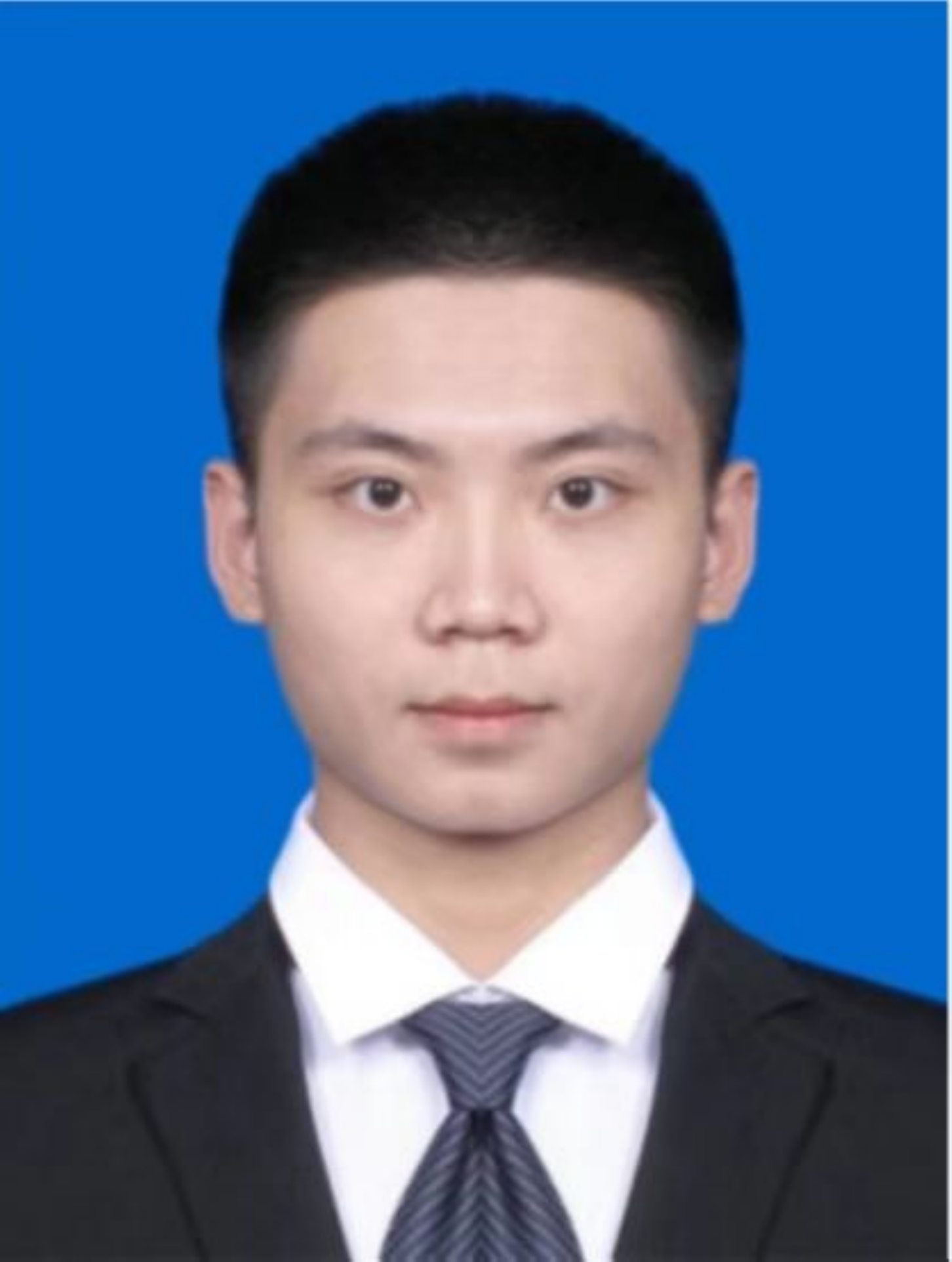}}]{Ziming Zhao}
Ziming Zhao received the bachelor degree from Tongji University, in 2022. From the fall of 2022, he is a master candidate at the School
of Electronics and Information Engineering, Tongji University. He focuses on the deep learning theory.
\end{IEEEbiography}
\begin{IEEEbiography}[{\includegraphics[width=1in,height=1.25in]{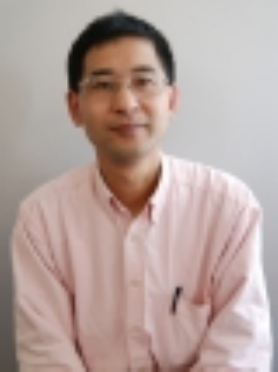}}]{Wanchenge Ge}
is a professor in the Department of Information and Communication Engineering, Tongji University, China. 
His research interest is on wireless communication and artificial intelligence.
\end{IEEEbiography}
\begin{IEEEbiography}[{\includegraphics[width=1in,height=1.25in]{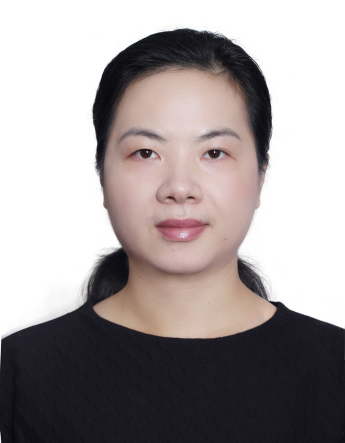}}]{Hao Zhang} received the B.Sc. degree in automatic control from the Wuhan University of Technology, Wuhan, China, in 2001, and the Ph.D. degree in control theory and control engineering from the
Huazhong University of Science and Technology, Wuhan, in 2007. From 2011 to 2013, she was a Postdoctoral Fellow with the City University of Hong Kong, Hong Kong. She is currently a Professor at the School
of Electronics and Information Engineering, Tongji University, Shanghai, China. Her research interests include network-based control systems, multiagent systems, and UAVs. Prof. Zhang is an Associate Editor of IEEE Intelligent Transportation Systems Magazine.
\end{IEEEbiography}
\begin{IEEEbiography}[{\includegraphics[width=1in,height=1.25in]{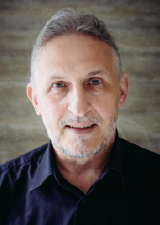}}]{Gerhard Rigoll} received the Dr.-Ing. habil. degree in 1991 from Stuttgart University with a thesis on speech synthesis. In 1993 he was appointed to full professor of computer science at Gerhard-Mercator- University in Duisburg, Germany. In 2002, he joined Technische Universit{\"{a}}t M{\"{u}}nchen (TUM), where he is now heading the institute for Human-Machine Communication. Dr. Rigoll is an IEEE Fellow (for contributions to multimodal human-machine communication) and is the author and co-author of more than 550 papers in the field of pattern recognition.
\end{IEEEbiography}
\clearpage
\appendices

\section{Proof}
\subsection{Proof Of Theorem \ref{LipschitzWaveletHolederSpace}}
\begin{proof}\label{LipschitzWaveletHolederSpaceProof}
Since $\int d x \psi(x)=0$,  it leads to the deduction that $\left\langle\psi^{a, b}, f\right\rangle=\int d x|a|^{-1 / 2} \psi\left(\frac{x-b}{a}\right)[f(x)-f(b)]$. Hence, we have an inference to reach the conclusion of the proof.
$$
\begin{aligned}
\left|\left\langle\psi^{a, b}, f\right\rangle\right| & \leq \int d x|a|^{-1 / 2}\left|\psi\left(\frac{x-b}{a}\right)\right| C|x-b|^{\alpha} \\
& \leq C|a|^{\alpha+1 / 2} \int d x|\psi(x+\delta)||x+\delta|^{\alpha} \\
& \leq C^{\prime}|a|^{\alpha+1 / 2} .
\end{aligned}
$$
\end{proof}
\subsection{Proof Of Theorem \ref{CompactSupportWavelet}}
\begin{proof}\label{CompactSupportWaveletProof}
We have a wavelet base $\psi_{2}$ compactly supported and continuously differentiable with $\int d x \psi_{2}(x)=0$. It is assumed that the normalization of $\psi_{2}$ is $C_{\psi, \psi_{2}}=1$, the integral of the wavelet transform is defined in Eq. (\ref{IntegralWaveletTransform}):
\begin{equation}\label{IntegralWaveletTransform}
f(x)=\int_{-\infty}^{\infty} \frac{d a}{a^{2}} \int_{-\infty}^{\infty} d b\left\langle f, \psi^{a, b}\right\rangle \psi_{2}^{a, b}(x).
\end{equation}
Then the integral would be split with two parts.

1) On the scenario of $|a| \geq 1$, the large scale item of integral is defined in Eq. (\ref{LargeScaleIntegral}). The integral is uniformly bounded for $x$.
\begin{equation}\label{LargeScaleIntegral}
\begin{aligned}
\left|f_{L }(x)\right| & \leq \int_{|a| \geq 1} \frac{d a}{a^{2}} \int_{-\infty}^{\infty} d b\left|\psi_{2}^{a, b}(x)\right|\|f\|_{L^{2}}\|\psi\|_{L^{2}} \\
& \leq C \int_{|a| \geq 1} \frac{d a}{a^{2}} \int_{-\infty}^{\infty} d b|a|^{-1 / 2}\left|\psi_{2}\left(\frac{x-b}{a}\right)\right| \\
& \leq C\left\|\psi_{2}\right\|_{L^{1}} \int_{-\infty} d a|a|^{-3 / 2}=C^{\prime}<\infty.
\end{aligned}
\end{equation}
We have $\left|f_{L }(x+\delta)-f_{L}(x)\right|$ for the perturbation $|\delta| \ll 1$,  the inequation is defined in Eq. (\ref{LargeScaleInequation}):
\begin{equation}\label{LargeScaleInequation}
\begin{aligned}
\left|f_{L}(x+\delta)-f_{L}(x)\right| \leq & \int_{|a| \geq 1} \frac{d a}{|a|^{3}} \int_{-\infty}^{\infty} d b \int_{-\infty}^{\infty} d y|f(y)| \cdot\\
&\left|\psi\left(\frac{y-b}{a}\right)\right|\cdot \\ 
&\left|\psi_{2}\left(\frac{x+\delta-b}{a}\right)-\psi_{2}\left(\frac{x-b}{a}\right)\right|.
\end{aligned}
\end{equation}
There exists the wavelet transform on the signal $\left|\psi_{2}(z+t)-\psi_{2}(z)\right| \leq C|t|$. For the compact support character in the real number space $\psi_{2} \subset[-R, R],\ R<\infty$, Eq. (\ref{LargeScaleInequation2}) can be deduced as follows:
\begin{equation}\label{LargeScaleInequation2}
\begin{aligned}
&\left|f_{L}(x+\delta)-f_{L}(x)\right|  \leq C^{\prime}|\delta| \int_{|a| \geq 1} d a a^{-4} \int_{x-b|\leq \operatorname{l}| R+1 \atop|\boldsymbol{v}-b| \leq|a| R} d y|f(y)| \\
& \leq\left. C^{\prime \prime}|\delta| \int_{|a| \geq 1} d a|a|^{-3} \int_{\mid y-x}\right|_{\leq 2|a| R+1} d y \quad|f(y)| \\
& \leq C^{\prime \prime}|\delta|\|f\|_{L^{2}} \int_{|a| \geq 1} d a|a|^{-3}(4|a| R+2)^{1 / 2} \leq C^{\prime \prime \prime}|\delta|.
\end{aligned}
\end{equation}
This holds for all $|\delta| \ll 1$ with the bound defined in Eq. (\ref{LargeScaleInequation}). Therefore, the function is uniform on the regular property $\left|f_{L}(x+\delta)-f_{L }(x)\right| \leq C|\delta|$.

2) On the scenario of $|a| \leq 1$, the small scale item of integral is satisfied with Eq. (\ref{SmallScaleIntegralInequation}):
\begin{equation}\label{SmallScaleIntegralInequation}
\begin{aligned}
 \left|f_{S }(x)\right| & \leq C \int_{|a| \leq 1} \frac{d a}{a^{2}} \int_{-\infty}^{\infty} d b|a|^{\alpha+1 / 2}|a|^{-1 / 2}\left|\psi_{2}\left(\frac{x-b}{a}\right)\right| \\
& \leq C\left\|\psi_{2}\right\|_{L^{1}} \int_{|a|<1} d a|a|^{-1+\alpha}=C^{\prime}<\infty.
\end{aligned}
\end{equation}
 For the perturbation $|\delta| \ll 1$, the Lipschitz regularity constraint still holds in Eq. (\ref{SmallScaleIntegralInequation2}) when using again the inequation $\left|\psi_{2}(z+t)-\psi_{2}(z)\right| \leq C|t|$. 
\begin{equation}\label{SmallScaleIntegralInequation2}
\begin{array}{l}
\begin{aligned}
&\left|f_{S }(x+\delta)-f_{S }(x)\right| &\\
&\leq \int_{|a| \leq|\delta|} \frac{d a}{a^{2}} \int_{-\infty}^{\infty} d b|a|^{\alpha}\left(\left|\psi_{2}\left(\frac{x-b}{a}\right)\right|+\left|\psi_{2}\left(\frac{x+\delta-b}{a}\right)\right|\right) \\
&+\int_{|\delta| \leq|a| \leq 1} \frac{d a}{a^{2}} \int_{|x-b| \leq|a| R+|\delta|} d b|a|^{\alpha} C\left|\frac{\delta}{a}\right|.
\end{aligned}
\end{array}
\end{equation}
The first part of integral can be written as $C^{\prime}\left[\left\|\psi_{2}\right\|_{L^{2}} \int_{|a| \leq|\delta|} d a|a|^{-1+\alpha}\right]$ while the second part of integral can be written as $C^{\prime}\left[|\delta| \int_{|\delta| \leq|a| \leq 1} d a|a|^{-3+\alpha}(|a| R+|\delta|)\right]$. The property is hold for $\left|f_{S }(x+\delta)-f_{S }(x)\right| \leq C^{\prime \prime}|\delta|^{\alpha}$. Then $f$ is H{\"{o}}lder with the exponent $\alpha$.  The proof has been done.

\end{proof}
\subsection{Proof Of Theorem \ref{LocalRegularityWavletForm}}
\begin{proof}\label{LocalRegularityWavletFormProof}
On the assumption of $x_{0}=0$, since $\int d x \psi(x)=0$, there exists an inequation defined in Eq. (\ref{LocalRegularityWaveletFormEq}) which proves the theorem:
\begin{equation}\label{LocalRegularityWaveletFormEq}
\begin{aligned}
\left|\left\langle f, \psi^{a, b}\right\rangle\right| & \leq \int d x|f(x)-f(0)||a|^{-1 / 2}\left|\psi\left(\frac{x-b}{a}\right)\right| \\
& \leq C \int d x|x|^{\alpha}|a|^{-1 / 2}\left|\psi\left(\frac{x-b}{a}\right)\right| \\
& \leq C|a|^{\alpha+1 / 2} \int d y\left|y+\frac{b}{a}\right|^{\alpha}|\psi(y)| \\
& \leq C^{\prime}|a|^{1 / 2}\left(|a|^{\alpha}+|b|^{\alpha}\right).
\end{aligned}
\end{equation}

\end{proof}
\subsection{Proof Of Theorem \ref{UpperBoundWavelet}}
\begin{proof}\label{UpperBoundWaveletProof}
The proof is very similar with the Proof of Theorem \ref{CompactSupportWavelet}. For simplicity, we can analyze the scenario with $|a| \leq 1$ of $\left|f_{S}\left(x_{0}+\delta\right)-f_{S}\left(x_{0}\right)\right|$ for small $\delta$ on the assumption that $x_{0}=0$ with the translation.
\begin{equation}\label{UpperBoundWaveletInequation}
\begin{aligned}
&\begin{array}{l}
\left|f_{S}(\delta)-f_{S}(0)\right| \\
\quad \leq \int_{|a| \leq|\delta|^{\alpha / \gamma}} \frac{d a}{a^{2}} \int_{-\infty}^{\infty} d b|a|^{\gamma}\left|\psi_{2}\left(\frac{\delta-b}{a}\right)\right| \\
\quad+\int_{|\delta|^{\alpha / \gamma \leq|a| \leq|\delta|}} \frac{d a}{a^{2}} \int_{-\infty}^{\infty} d b\left(|a|^{\alpha}+\frac{|b|^{\alpha}}{|\log | b||}\right)\left|\psi_{2}\left(\frac{\delta-b}{a}\right)\right|
\end{array}\\
&+\int_{|a| \leq|\delta|} \frac{d a}{a^{2}} \int_{-\infty}^{\infty} d b\left(|a|^{\alpha}+\frac{|b|^{\alpha}}{|\log | b||}\right)\left|\psi_{2}\left(-\frac{b}{a}\right)\right|\\
&+\int_{|\delta| \leq|a|} \frac{d a}{a^{2}} \int_{-\infty}^{\infty} d b\left(|a|^{\alpha}+\frac{|b|^{\alpha}}{|\log | b||}\right)\left|\psi_{2}\left(\frac{\delta-b}{a}\right)-\psi_{2}\left(-\frac{b}{a}\right)\right|\\
.
\end{aligned}
\end{equation}
The right hands of Eq. (\ref{UpperBoundWaveletInequation}) can be devided into four parts: $T_{1}$, $T_{2}$ with support of $\psi_{2} \subset[-R, R]$, $T_{3}$, and $T_{4}$.  Four separated inequations can be derived.
\begin{equation}\label{FirstTerm}
\begin{aligned}
T_{1} \leq \int_{|a| \leq|\delta|^{\alpha / \gamma}} d a|a|^{-1+\gamma}\left\|\psi_{2}\right\|_{L^{1}}  \leq C|\delta|^{\alpha}. 
\end{aligned}
\end{equation}
\begin{equation}\label{SecondTerm}
\begin{aligned}
T_{2} &\leq  \int_{|a| \leq|\delta|} d a|a|^{-1+\alpha}\left\|\psi_{2}\right\|_{L^{1}} \\
&+\int_{|\delta|^{\alpha / \gamma} \leq|a| \leq|\delta|} d a|a|^{-1}\left\|\psi_{2}\right\|_{L^{1}} \frac{(|a| R+|\delta|)^{\alpha}}{|\log (|a| R+|\delta|)|} \\
&\leq  C|\delta|^{\alpha}\left[1+\frac{1}{|\log | \delta||} \int_{|\delta| \alpha / \gamma \leq|a| \leq|\delta|} d a|a|^{-1}\right] \\
&\leq  C^{\prime}|\delta|^{\alpha}.
\end{aligned}
\end{equation}
\begin{equation}\label{ThirdTerm}
\begin{aligned}
T_{3} & \leq \int_{|a| \leq \delta} d a|a|^{-1+\alpha}\left\|\psi_{2}\right\|_{L^{1}} \\
&+\int_{|a| \leq|\delta|} d a|a|^{-1}\left\|\psi_{2}\right\|_{L^{2}} \frac{(|a| R)^{\alpha}}{|\log | a|R|} \\
& \leq C|\delta|^{\alpha}.
\end{aligned}
\end{equation}
\begin{equation}\label{FourthTerm}
\begin{aligned}
T_{4} & \leq C|\delta| \int_{|\delta| \leq|a| \leq 1} d a|a|^{-3} \\
&\left[|a|^{\alpha}+\frac{(|a| R+|\delta|)^{\alpha}}{|\log (|a| R+|\delta|)|}\right](|a| R+|\delta|) \\
& \leq C^{\prime}|\delta|\left[1+|\delta|^{-1+\alpha}+|\delta|\left(1+|\delta|^{-2+\alpha}\right)\right] \leq C^{\prime \prime}|\delta|^{\alpha}.
\end{aligned}
\end{equation}

It proves the lower-order local regularity which gives an illustration of the ``mathematical microscope" of the wavelet.

\end{proof}
\subsection{Proof Of Theorem \ref{HoelderContinuityInSmallPerturbation}}
\begin{proof}\label{HoelderContinuityInSmallPerturbationProof}
There are two scenarios: $(\ell-1) 2^{-j} \leq x \leq x+\delta \leq \ell 2^{-j}$, $(\ell-1) 2^{-j} \leq x \leq \ell 2^{-j} \leq x+\delta \leq(\ell+1) 2^{-j}$. The proof on the second scenario can be derived from the similar proof on the first scenario. We then have an inequality defined in Eq. (\ref{LipschitzInequality}):
\begin{equation}\label{LipschitzInequality}
\begin{aligned}
&|f(x)-f(x+\delta)| \leq \left|f(x)-f_{j}(x)\right|+\left|f_{j}(x)-f_{j}\left(\ell 2^{-j}\right)\right| \\
&+\left|f_{j}\left(\ell 2^{-j}\right)-f_{j}(x+\delta)\right|+\left|f_{j}(x+\delta)-f(x+\delta)\right| \\
&\leq  2 C 2^{-\alpha j}+\left|f_{j}(x)-f_{j}\left(\ell 2^{-j}\right)\right|+\left|f_{j}(x+\delta)-f_{j}\left(\ell 2^{-j}\right)\right|.
\end{aligned}
\end{equation}
Different choices of $l$ denote different resolutions. It is assumed that there exists a variable $k \in \mathbb{N}$ so that $x^{\prime}=x-k$ and $\ell^{\prime} 2^{-j}=\ell 2^{-j}-k$ are both in the range of $[0,1]$. If $x^{\prime}$ is dyadic, the binary expansion of Eq. (\ref{LipschitzInequality}) is defined in Eq. (\ref{LipschitzInequality2}).
\begin{equation}\label{LipschitzInequality2}
\begin{aligned}
&\left|f_{j}(x)-f_{j}\left(\ell 2^{-j}\right)\right|  \leq\left\|v_{j}\left(x^{\prime}\right)-v_{j}\left(\ell^{\prime} 2^{-j}\right)\right\| \\
&=\left\|T_{d_{1}\left(x^{\prime}\right)} \cdots T_{d_{j}\left(x^{\prime}\right)}\left[v_{0}\left(\tau^{j} x^{\prime}\right)-v_{0}\left(\tau^{j}\left(\ell^{\prime} 2^{-j}\right)\right)\right]\right\| \\
& \leq C 2^{-\alpha j},
\end{aligned}
\end{equation}
where $\left\|\left.T_{d_{1}} \cdots T_{d_{m}}\right|_{E_{2}}\right\| \leq C 2^{-\alpha j}$ and $v_{0}$ is bounded with the property $v_{0}(u)-v_{0}\left(u^{\prime}\right) \in E_{1}$ for all $u, u^{\prime}$. The similar bound $\left|f_{j}(x+\delta)-f_{j}\left(\ell 2^{-j}\right)\right|$ can also be derived.
The inequation defined in Eq. (\ref{HoelderRepresentation}) can be proven which is H{\"{o}}lder continuous with the exponent $\alpha$.

\end{proof}

\ifCLASSOPTIONcaptionsoff
  \newpage
\fi

\end{document}